\documentclass{article}

\PassOptionsToPackage{numbers, compress}{natbib}
\usepackage[preprint]{neurips_2026}

\usepackage[utf8]{inputenc} % allow utf-8 input
\usepackage[T1]{fontenc}    % use 8-bit T1 fonts
\usepackage{hyperref}       % hyperlinks
\usepackage{url}            % simple URL typesetting
\usepackage{booktabs}       % professional-quality tables
\usepackage{amsfonts}       % blackboard math symbols
\usepackage{nicefrac}       % compact symbols for 1/2, etc.
\usepackage{xcolor}         % colors
\RequirePackage{graphicx}
\usepackage{xspace}
\usepackage{graphicx}
\usepackage{subfigure}
\usepackage{booktabs}
\usepackage[shortlabels]{enumitem}
\usepackage{pifont}
\usepackage[dvipsnames]{xcolor}

\RequirePackage{fancyhdr}
\RequirePackage{xcolor}
\RequirePackage{algorithm}
\RequirePackage{algorithmic}
\RequirePackage{natbib}
\RequirePackage{eso-pic}
\RequirePackage{forloop}
\RequirePackage{url}
\usepackage{hyperref}
\usepackage{booktabs}
\usepackage{amsmath,amssymb,amsthm}
\usepackage{soul}
\usepackage{subcaption}
\usepackage{multirow}
\usepackage{mathtools}
\usepackage{amsthm}
\usepackage[thinc]{esdiff}
\usepackage[table]{xcolor}
\definecolor{red}{rgb}{0.921569, 0.282353, 0.247059}
\definecolor{green}{rgb}{0.501961, 0.792157, 0.239216}
\definecolor{orange}{rgb}{0.937255, 0.529412, 0.2}
\usepackage{longtable}
\usepackage[capitalize,noabbrev]{cleveref}
\usepackage[textsize=tiny]{todonotes}
\setcitestyle{square}

\theoremstyle{plain}
\newtheorem{theorem}{Theorem}[section]
\newtheorem{proposition}[theorem]{Proposition}

\newtheorem{corollary}[theorem]{Corollary}
\theoremstyle{definition}

\theoremstyle{remark}

\title{LiLAW: Lightweight Learnable Adaptive Weighting to Learn Sample Difficulty \& Improve Noisy Training}

\author{%
  Abhishek Moturu \\ %\thanks{Use footnote for providing further information - about author (webpage, alternative address)---\emph{not} for acknowledging funding agencies.} \\
  Department of Computer Science\\
  University of Toronto\\
  The Hospital for Sick Children \\
  UHN KITE Research Institute \\
  T-CAIREM \\
  Vector Institute \\
  \texttt{moturuab@cs.toronto.edu} \\
  \And
  Muhammad Muzammil \\ %\thanks{Use footnote for providing further information - about author (webpage, alternative address)---\emph{not} for acknowledging funding agencies.} \\
  Department of Computer Science\\
  Department of Mathematics \\
  Department of Statistics \\
  University of Toronto\\
  The Hospital for Sick Children \\
  Vector Institute \\
  \texttt{m.muzammil@mail.utoronto.ca} \\
  \And
  Anna Goldenberg $^*$  \\
  Department of Computer Science\\
  Department of Laboratory Medicine and Pathobiology \\
  University of Toronto\\
  The Hospital for Sick Children \\
  T-CAIREM \\
  Vector Institute \\
  \texttt{anna.goldenberg@utoronto.ca} \\
  \AND
    Babak Taati \thanks{equal contribution} \\ %\thanks{Use footnote for providing further information - about author (webpage, alternative address)---\emph{not} for acknowledging funding agencies.} \\
  Department of Computer Science\\
  Institute of Biomedical Engineering \\
  University of Toronto\\
  Rehabilitation Sciences Institute \\
  UHN KITE Research Institute \\
  Vector Institute \\
  \texttt{taati@cs.toronto.edu} \\
}

\begin{document}

\newcommand{\synpain}{Syn\textsc{Pain}\xspace}
\newcommand{\gaitgen}{\textsc{Gait}Gen\xspace}
\maketitle
\setcounter{footnote}{0}

%\printAffiliationsAndNotice{}  % no special notice (required even if empty)
\vspace{-1em}
\begin{abstract}
\vspace{-0.4em}
Training deep neural networks with noise and data heterogeneity is a major challenge. We introduce Lightweight Learnable Adaptive Weighting (LiLAW), a method that dynamically adjusts the loss weight of each training sample based on its evolving difficulty, categorized as easy, moderate, and hard, using only three global learnable scalar parameters. LiLAW learns to adaptively prioritize samples by updating these parameters with a single gradient descent step on a validation mini-batch after each training mini-batch, without requiring a clean, unbiased validation set. Experiments across general and medical imaging datasets, several noise types and levels, loss functions, and architectures with and without pretraining, including linear probing and full fine-tuning, show that LiLAW consistently improves accuracy and AUROC, especially in higher-noise settings, without requiring excessive tuning. We also obtain state-of-the-art results incorporating synthetic and augmented data from \synpain, \gaitgen, ECG5000, and improved fairness on the Adult dataset. LiLAW is lightweight, practical, and computationally efficient, making it an effective, scalable approach to boost generalization and robustness across diverse deep learning training setups, especially in resource-constrained settings. %\footnote{Code in Supplementary Material.}

% TALK ABOUT resource constrained environments
% SAY that we can identify noise well in the training data
\end{abstract}
%\vspace{-1em}
\section{Introduction}
\label{sec:introduction}

The increasing availability of very large labeled datasets has played a major role in advancing machine learning (ML) and computer vision in recent years. However, imaging datasets often contain samples with varying levels of quality, affecting the efficiency and effectiveness of model training. This issue is particularly significant in medical datasets, which typically have smaller sample sizes, exhibit greater heterogeneity, and often require specialized expertise for accurate labeling. Ensuring that models make the best use of a small amount of noisy, heterogeneous data remains a challenge, as not all samples are equally informative or beneficial for model performance. At varying rates at different points during training, samples can benefit the model, hurt it, or not affect it much at all.

Several methods have been proposed to quantify the difficulty or importance of individual samples~\citep{agarwal_estimating_2022, baldock2021deep, dong2023generalized, jiang2020characterizing, kong_adaptive_2021, liu2021probabilistic, maini2022characterizing, paul2021deep, pliushch2022deep, rabanser2022selective, seedat_dissecting_2024, siddiqui2022metadata, toneva_empirical_2019, xu_understanding_2021}. Understanding which samples a model finds difficult to predict is essential for safe model deployment, sample selection for human-in-the-loop auditing, and gaining insights into model behavior~\citep{agarwal_estimating_2022}. Knowing or estimating example difficulty can help separate misclassified, mislabeled, and rare examples~\citep{maini2022characterizing}, prune data~\citep{paul2021deep}, quantify uncertainty~\citep{dong2023generalized}, improve generalization~\citep{xu_understanding_2021}, increase convergence speed~\citep{kong_adaptive_2021}, enhance out-of-distribution detection~\citep{agarwal_estimating_2022}, and provide insights into example memorization and forgetting~\citep{toneva_empirical_2019}. While existing methods provide insights into individual data points and how models learn, they often do not adjust the training process in a dynamic, efficient manner. This may restrict potential improvements in model performance, robustness, and generalization.

\begin{figure}
    \centering
    \includegraphics[scale=0.477]{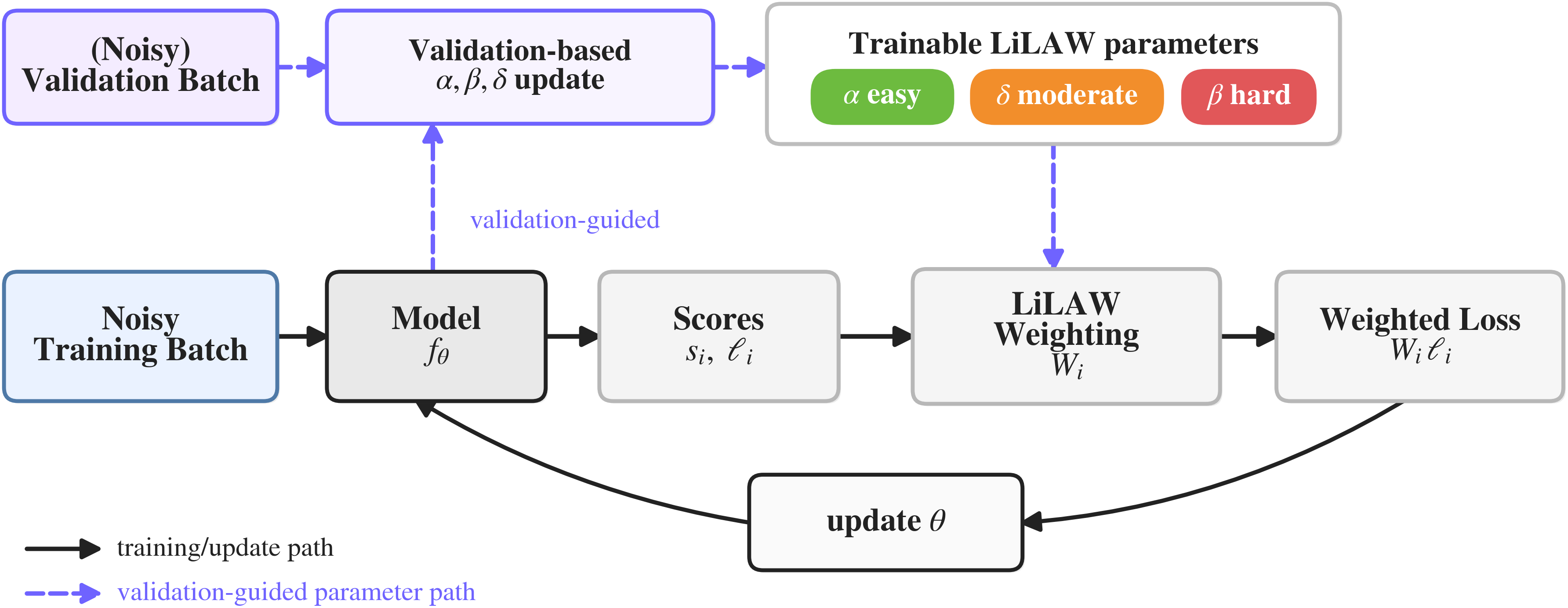}
    \caption{Given noisy training and validation data, LiLAW learns to adaptively weight the loss of each sample based on three trainable parameters, $\alpha, \delta, \beta$, pertaining to learning \textcolor{green}{easy}, \textcolor{orange}{moderate}, and \textcolor{red}{hard} samples, respectively, at different training stages using validation-guided adaptive reweighting.}
    \label{fig:diagram}
\end{figure}

%Focusing training on the right examples at the right time can improve robustness, but many weighting methods are either static, require a clean validation set, or introduce additional models (see Section~\ref{sec:sampleweighting}). We instead design a lightweight approach that uses the model's own confidence and disagreement with the observed label to adjust sample contributions throughout training. 

%We propose \textbf{Li}ghtweight \textbf{L}earnable \textbf{A}daptive \textbf{W}eighting (\textbf{LiLAW}), which parameterizes a smooth per-sample weighting with only three global scalars $(\alpha,\beta,\delta)$, which define easy, hard, and moderate samples in the confidence / disagreement plane . The parameters are updated using validation-guided reweighting, while the classifier is trained with the resulting weighted loss.

Focusing learning on the most informative examples at the right stage of training can effectively improve robustness. We introduce \textbf{Li}ghtweight \textbf{L}earnable \textbf{A}daptive \textbf{W}eighting (\textbf{LiLAW}), which turns the classifier's own predictive behavior into a continuously learned sample-weighting signal. Rather than weighting samples by raw loss alone or relying on fixed notions of difficulty, LiLAW models each example through two quantities: the score assigned to the observed label and the maximum predicted score. These quantities define a confidence-disagreement plane (see \Cref{fig:diagram,fig:graph}) in which easy, moderate, and hard examples occupy distinct geometric regions. LiLAW then learns a smooth, strictly positive, differentiable weighting function over this plane using only three learnable global scalar parameters, $\alpha,\beta,\delta$, which adapt throughout training through validation-guided reweighting while the classifier is optimized with the resulting weighted loss.

LiLAW is deliberately lightweight, introducing no static schedules, hand-designed curricula, dependence on clean validation data, auxiliary weighting network, instance-specific weighting model, separate noise or difficulty detector, extra prediction head, or pruning stage. Its computational overhead consists only of one additional forward-backward pass on a single validation mini-batch after each training mini-batch, while preserving the asymptotic time and memory complexity of standard training. It also does not rely on a clean or unbiased validation set, making the method practical in precisely the regimes where robust reweighting is most needed.

%: we show that it remains robust under noisy validation labels, small validation sets, validation distribution shift, different parameter initializations, and random seeds. This .

Beyond its algorithmic simplicity, LiLAW provides a geometric view of adaptive reweighting. Its learned confidence-disagreement geometry distinguishes easy, moderate, and hard examples without imposing a fixed curriculum, and our theoretical analysis under diagonally dominant label noise shows that LiLAW's weights contain a strict separation signal between clean and noisy samples. 

%This connects the learned weighting dynamics to label-noise robustness while explaining why the method can prioritize useful training signal without collapsing into either memorization of corrupted labels or over-suppression of difficult examples.

Empirically, LiLAW is effective across a wide range of regimes, datasets, and modeling choices. We evaluate it on general imaging benchmarks, ten MedMNISTv2 datasets, and the large-scale Clothing-1M dataset, showing robust gains across diverse noise patterns and data domains. We further demonstrate application across losses, architectures, and training protocols, including linear probing, full fine-tuning, and training from scratch, and that it extends naturally beyond standard multi-class classification to multi-label classification and regression. In application-focused studies, LiLAW also enables effective use of synthetic and augmented data for pain detection on UofR, gait classification on PD-GaM, and heartbeat classification on ECG5000, achieving state-of-the-art results across all three settings. Finally, we analyze when adaptive reweighting can reduce fairness disparities, verify these effects empirically on the Adult dataset, and show that LiLAW's learned weights provide strong diagnostic signals for surfacing mislabeled samples as well as interpretable easy, moderate, and hard examples in potentially noisy training sets.

Taken together, LiLAW offers a combination of simplicity, interpretability, theoretical grounding, and empirical breadth: a three-parameter adaptive weighting mechanism that improves robustness without auxiliary models or clean validation assumptions, generalizes across tasks and training regimes, supports practical data curation, and provides a unified confidence-disagreement perspective on how neural networks should weight examples during learning.

% summary of numerical results
% more about results and impact in a paragraph

\section{Related Work}
\label{sec:relatedwork}

The effectiveness of ML models, especially deep neural networks, is highly dependent on the quality and proper use of training data. Recent literature explored strategies to improve model performance and robustness to address the challenges of noisy, mislabeled, imbalanced, or otherwise hard data.
\vspace{-0.7em}
\subsection{Data-Centric AI}

\citet{pleiss_identifying_2020} proposed using the area under the margin ranking to identify mislabeled data and improve model performance by filtering out mislabeled data, which requires ``dataset cleaning.''
\citet{toneva_empirical_2019} conducted an empirical study on example forgetting during deep neural network training that shed light on data quality, highlighting that some samples are forgotten frequently, some are not forgotten at all, and some could be omitted from training without affecting the model greatly. \citet{paul2021deep} highlight that simple scores such as the Gradient Normed (GraNd) and Error L2-Norm (EL2N) can be used to identify important examples very early in training and to prune significant portions of the training data while maintaining test accuracy. \textit{Our method instead aims to improve test performance without pruning any training data.}

\citet{wu_learning_2023} presented a framework for selecting pivotal samples for meta re-weighting, aiming to optimize performance on a small set of perfect samples, which may not be available in many real-world cases. \citet{mindermann_prioritized_2022} note that many samples may be hard to learn since they are noisy or unlearnable. They propose a method that selects points that are learnable, worth learning, and not yet learned. However, this method requires a separate small model trained on a holdout set. \textit{Our method does not require a clean, unbiased validation set or a separate model.}

\citet{agarwal_estimating_2022} propose estimating example difficulty using variance of gradients to rank data by hardness and identify samples for auditing. \citet{swayamdipta_dataset_2020} present a method to map datasets using training dynamics to identify easy, hard, and ambiguous examples. Area under the margin~\citep{pleiss_identifying_2020} measures the difference between the score of the observed label in logits of the neural network and the maximum logit (discounting the score of the observed label). These methods do not use difficulty information to inform training. \textit{Our method serves both to assess sample difficulty and to use that information to enhance training.}

\citet{jia_learning_2022} use raw training dynamics as input to an LSTM noise detector network which learns to predict mislabels. 
\citet{jiang_delving_2021} use loss curves to identify corrupted labels by training a whole network on the original data and then training a CurveNet on the loss curves of each sample as an additional attribute to identify bias type. \citet{kong_adaptive_2021} propose adaptive (as opposed to fixed) curriculum learning which uses the current model to adjust loss-based difficulty scores while retaining learned knowledge from a pre-trained model using the KL divergence between the outputs of the current model and the pre-trained model and a pacing function to control the learning pace. These methods and others \citep{unicon,dividemix,combatingsurrogate,topological} require additional models for difficulty-learning or data selection. \textit{Our method does not require additional models.}

Accurate uncertainty estimation and model calibration are essential for reliable predictions. Models are well-calibrated if predicted probabilities accurately reflect the chance of an event occurring. \citet{guo_calibration_2017} investigated the calibration of neural networks and found that deeper networks are often poorly calibrated. They propose temperature scaling as an effective calibration method. Another approach to improve calibration is label smoothing \citep{muller_when_2020}. \textit{Our method works synergistically with calibration to improve model performance.}

Several methods were developed to characterize sample hardness and data quality such as Data-IQ (Data-Inherent Qualities)~\cite{seedat_data-iq_2022}, DIPS (Data-centric Insights for Pseudo-labeling with Selection)~\cite{seedat_you_2024}, and H-CAT (Hardness Characterization Analysis Toolkit)~\cite{seedat_dissecting_2024}. We use H-CAT to study the effectiveness of our method across various settings.

\subsection{Sample Weighting}
\label{sec:sampleweighting}
Sample weighting aims to improve model training by assigning weights to samples based on their learning difficulty. Traditional methods often rely on static weighting schemes, which fail to adapt to a model's evolving learning dynamics. Several works introduced a theoretical framework connecting generalization error to difficulty \citep{zhou_which_2023, zhou_understanding_2023, zhu_exploring_2022}. \citet{zhou_which_2023} propose adaptive weighting strategies that consider easy, moderate, and hard samples and use hyperparameter tuning to find the most effective weighting solution, which stays fixed, from easy-first, hard-first, medium-first, and two-ends-first or a pre-defined schedule (easy-first, then hard-first) during training. \textit{Our method adaptively prioritizes different samples as training progresses.}

\citet{xu_understanding_2021} mention that using counterfactual modeling to jointly train a weighting model with the classifier eventually causes weights to converge to the same constant. Others state that increasing weights on hard samples may improve both convergence and performance, but assume that training noise is absent~\citep{xu_understanding_2021, zhou_understanding_2023}. \textit{Our method assumes that training label noise may exist and does not require extensive hyperparameter tuning to find the best learning strategy.}

Meta-learning aims to improve the learning process itself. It is often referred to as ``learning to learn'' in literature. Recent work has shown promise in using meta-learning to learn weights for samples. %\citet{wu_revisiting_2024} propose a theoretical framework for sample weighting based on the effective number theory for imbalanced learning using meta-learning, which usually requires an unbiased meta-set.
To help select samples for the meta-set, \citet{jain_improving_2024} employ a bi-level objective aimed at identifying the most challenging samples in the training data to use as a validation set and subsequently train the classifier to reduce errors on those specific samples, which is a way of learning a Learned Reweighting (LRW) classifier without a fixed, unbiased, and clean validation set~\citep{ren_learning_2018}. However, this requires three networks: one to learn the task, one that identifies challenging samples to validate, and one that learns sample weights, vastly increasing computational complexity. \textit{Our method only requires three additional trainable global parameters.}

%Methods like probabilistic margin~\citep{liu2021probabilistic} and area under the margin~\citep{pleiss_identifying_2020} measure the difference between the score of the observed label in logits of the neural network and the maximum logit (discounting the score of the observed label). The range of possible negative values makes it difficult to use these methods in loss-based weighting schemes. \textit{In contrast, our method guarantees reasonable positive weights.}

%This assumes the existence of a pre-trained network for the task at hand. \textit{Our method does not require additional models and only needs one additional forward and backward pass to do a single gradient descent step on a validation mini-batch after every training mini-batch.}

\section{Method}
\label{sec:method}

Consider the following supervised multi-class classification problem setup used in prior work~\citep{northcutt_confident_2021}.
Let $\mathcal{D}_{t} = \{(x_i, \widetilde{y_i})\}^N_{i=1}$ represent the training set and $\mathcal{D}_{v} = \{(x_j, \widetilde{y_j})\}^{N+M}_{j=N+1}$ represent the validation set. Note that $(x_i, \widetilde{y_i})$ represents the pairs of inputs and observed (potentially noisy) targets. Specifically, $x_i \in \mathcal{X}$, where $\mathcal{X}$ is the input space (e.g.: images) and $\widetilde{y_i} \in \mathcal{Y} = \{0,...,c-1\},$ is the output space with $c \in \mathbb{N}$ such that $c \ge 2$ is the total number of classes. Note that $\widetilde{y_i}$ is a single integer value, i.e. $x_i$ belongs to a single class. Let $y_i \in \mathcal{Y}$ be the true target. Let $f_{\theta}: \mathcal{X} \rightarrow \mathbb{R}^c$ be the neural network model, $\theta$ be its parameters, and $s_i = \mathrm{softmax}(f_{\theta}(x_i))$ be the softmax of its logits. Therefore, $s_i \in \mathbb{R}_{>0}^c$ such that $\sum_{j=0}^{c-1} s_i[j] = 1$. Note that $s_i[j]$, where $j \in \{0,...,c-1\}$, refers to the softmaxed logit for class $j$ after passing input $x_i$ through the model.

Our method's motivation stems from the properties of two key values, $s_i[\widetilde{y_i}]$ and $\max(s_i)$, which, as shown below, indicate whether the model prediction agrees with observed label and whether it does so with low or high confidence:
\vspace{-0.5em}
\begin{itemize}
    \item $s_i[\widetilde{y_i}] = \max(s_i)$ and $\max(s_i)$ is high (\textit{easy})
    \item $s_i[\widetilde{y_i}] = \max(s_i)$ and $\max(s_i)$ is low (\textit{moderate})
    \item $s_i[\widetilde{y_i}] < \max(s_i)$ and $\max(s_i)$ is low (\textit{moderate})
    \item $s_i[\widetilde{y_i}] < \max(s_i)$ and $\max(s_i)$ is high (\textit{hard})
\end{itemize}

See \ref{supp:example} for a motivating example. Note: In the noisy setting, i.e. when the observed target is not the same as the true target (i.e. $\widetilde{y_i} \neq y_i$), we rely on prediction confidence to inform sample difficulty. 

\begin{figure}
    \centering
    \includegraphics[scale=0.27]{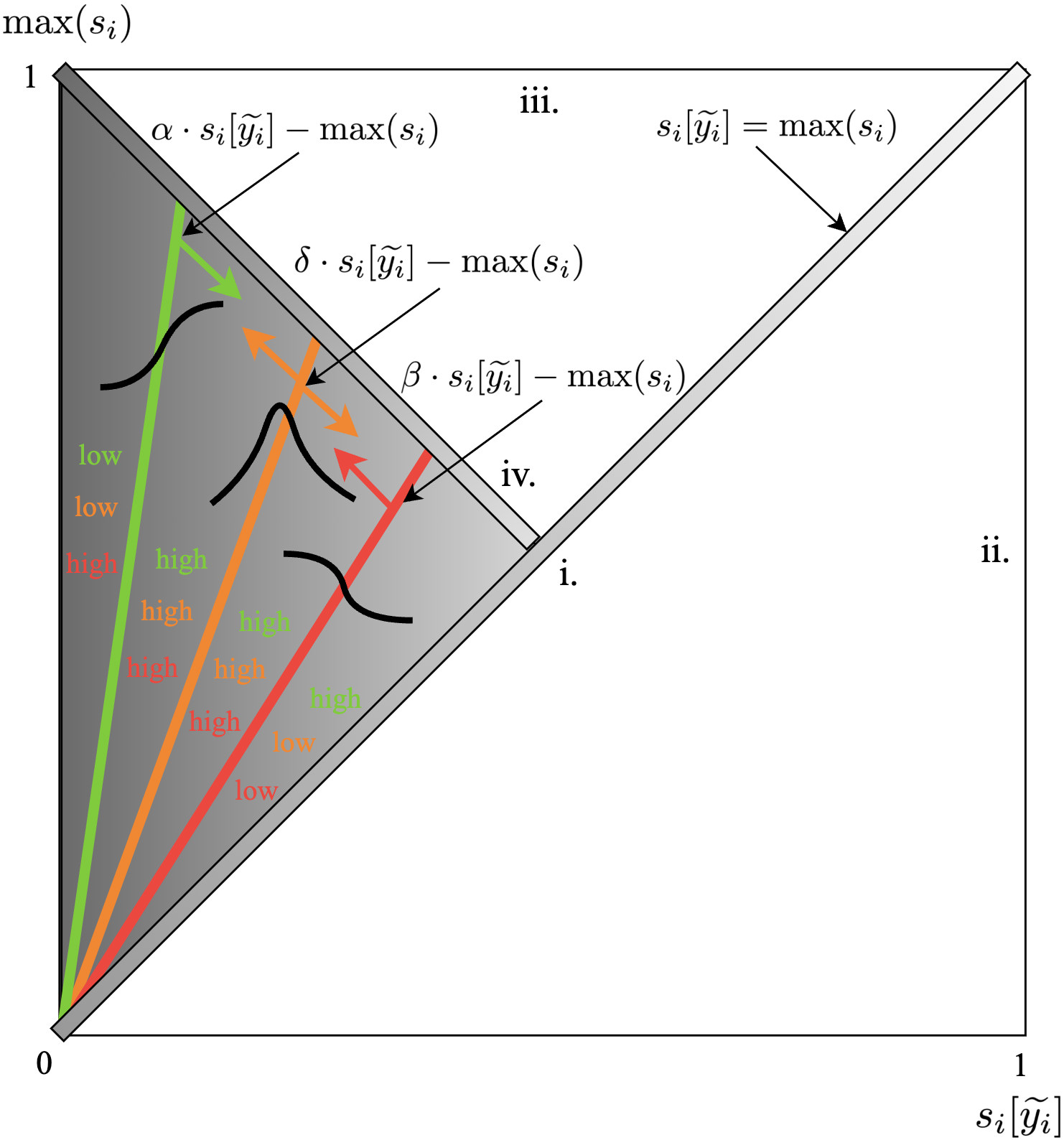}
    \caption{A graphical representation of the LiLAW weighting method. Darker areas correspond to high loss (due to disagreement and/or low confidence) and lighter areas correspond to low loss (due to agreement and/or confidence). The three weight functions in Equations (2), (3), and (4) correspond to the green, red, and orange lines, respectively, with the colored arrows representing the descent direction for each weight function. Note: Equations (2) and (3) use a sigmoid and Equation (4) uses a radial basis function. We also signify whether each weight is high or low in a given region.}
    \label{fig:graph}
\end{figure}

\noindent We have the following relations between $s_i[\widetilde{y_i}]$ and $\max(s_i)$ which define the area in Figure~\ref{fig:graph}:
\vspace{-0.5em}
\begin{enumerate}[label=\roman*.]
    \itemsep-0.2em
    \item $0 \leq s_i[\widetilde{y_i}] \leq \max(s_i)$,
    \item $0 \leq s_i[\widetilde{y_i}] \leq 1$,
    \item $0 < \max(s_i) \leq 1$,
    \item Since $\sum_{j=0}^{c-1} s_i[j] = 1$, we know $s_i[\widetilde{y_i}] + \max(s_i) \leq 1$ when $s_i[\widetilde{y_i}] \neq \max(s_i)$.
\end{enumerate}

The above constraints form a region, shown in gray in Figure \ref{fig:graph}, which contains the values that $s_i[\widetilde{y_i}]$ and $\max(s_i)$ can attain. The darker shading corresponds to high loss areas and the lighter shading corresponds to lower loss. The diagonal band going from the bottom left to the top right corresponds to: $s_i[\widetilde{y_i}] = \max(s_i)$. Closer to the bottom left, we make low confidence predictions and closer to the top right, we make more confident predictions. The remaining shaded areas correspond to: $s_i[\widetilde{y_i}] < \max(s_i)$. Near the top left of the region, we have more confident predictions and near the bottom left, we have low confidence predictions. Closer to the center of the region, we have somewhat low confidence predictions. Our method attempts to push low confidence predictions to being confident and incorrect predictions to being correct (with respect to the observed label). %As a result, it is able to deal with label noise.

In~\ref{supp:geometric_interpretation}, we give a geometric view of LiLAW as two smooth half-space detectors and one localized band-pass detector in the confidence/disagreement plane, explaining why the three terms capture instance-dependent and class-dependent difficulty.

The above properties provide granularity to help us identify predictions as being incorrect or correct (with respect to the observed label) and low confidence or high confidence. However, they are not sufficient to learn to adaptively weight the losses of certain samples during training. Therefore, we use parameters to control the relationship between $s_i[\widetilde{y_i}]$ and $\max(s_i)$ during training. We define three scalar parameters, $\alpha, \beta, \delta$, to define the contribution of different data points to the loss function.

We define our unweighted loss function (in our case, $\mathcal{L}$ is cross-entropy loss or focal loss~\citep{focalloss}): 
\begin{align}
    \mathcal{L} = \ell(f_{\theta}(x_i), \widetilde{y_i})
\end{align}

We define the weights pertaining to $\alpha,\beta,\delta$ as follows:
\begin{align}
    \mathcal{W}_\alpha (s_i, \widetilde{y_i}) = \sigma(\alpha \cdot s_i[\widetilde{y_i}] - \max(s_i))
\end{align}
\begin{align}
    \mathcal{W}_\beta (s_i, \widetilde{y_i}) = \sigma(-(\beta \cdot s_i[\widetilde{y_i}] - \max(s_i)))
\end{align}
\begin{align}
    \mathcal{W}_\delta (s_i, \widetilde{y_i}) = \exp\biggl(-\frac{(\delta \cdot s_i[\widetilde{y_i}] - \max(s_i))^2}{2}\biggr)
\end{align}

\noindent We define the weight for each sample as follows:
\begin{align}
    \mathcal{W}(s_i, \widetilde{y_i}) = \mathcal{W}_\alpha (s_i, \widetilde{y_i}) + \mathcal{W}_\beta (s_i, \widetilde{y_i}) + \mathcal{W}_\delta (s_i, \widetilde{y_i})
\end{align}

\noindent and define our LiLAW (weighted) loss function as follows:
\begin{align}
\mathcal{L}_{W} &= \mathcal{W}(s_i, \widetilde{y_i}) \cdot \mathcal{L}
\end{align}

We define a sample's weight as the sum of $\mathcal{W}_\alpha$, $\mathcal{W}_\beta$, and $\mathcal{W}_\delta$ so that the final contribution of a sample is not governed by one parameter, but by the combined effect of all three. Easy, hard, and moderate samples primarily activate $\mathcal{W}_\alpha$, $\mathcal{W}_\beta$, and $\mathcal{W}_\delta$, respectively, while their overlap provides smooth transitions near boundaries and keeps LiLAW differentiable with respect to $\alpha,\beta,\delta$. See \ref{supp:samples} for a clear set of sample images from CIFAR-100-M with the highest $\mathcal{W}_\alpha$, $\mathcal{W}_\beta$, and $\mathcal{W}_\delta$ weights that can be used to very effectively surface easy, hard, and moderate samples, respectively.

Thus, \(W_\alpha\) is high in low-disagreement regions where \(\max(s_i) - s_i[\tilde y_i] \lesssim (\alpha-1)s_i[\tilde y_i]\) (easy), \(W_\beta\) is high in high-disagreement regions where \(\max(s_i) - s_i[\tilde y_i] \gtrsim (\beta-1)s_i[\tilde y_i]\) (hard), and \(W_\delta\) is high near the transition band where \(\max(s_i) - s_i[\tilde y_i] \approx (\delta-1)s_i[\tilde y_i]\) (moderate). We enforce $\alpha \geq \delta \geq \beta \geq 1$ to weight low confidence samples sufficiently and encourage moderate samples to occupy a transition region between easy and hard samples.

\begin{algorithm}[ht]
\caption{Training with LiLAW in black and \textcolor{gray}{gray}, \textcolor{gray}{compared to standard training in gray}}
\begin{algorithmic}[1]
        \STATE \textbf{Inputs:} \textcolor{gray}{training set $(\mathcal{D}_{t})$}, \textcolor{gray}{validation set $(\mathcal{D}_{v})$}, \textcolor{gray}{classifier model $f_\theta$}, LiLAW parameters $\alpha, \beta, \delta$, optimizer for \textcolor{gray}{$\theta$}, $\alpha, \beta, \delta$,  learning rates $\alpha_{lr}, \beta_{lr}, \delta_{lr},$ weight decay values $\alpha_{wd}, \beta_{wd}, \delta_{wd}$
        \STATE \textbf{Output:} \textcolor{gray}{classifier model after training} using LiLAW
            \FOR{\textcolor{gray}{epoch $\gets 1$ \textbf{to} $n$}}
                \FOR{\textcolor{gray}{batch $(x_t, \widetilde{y}_t) \gets $ dataloader($\mathcal{D}_{t}$)}}
                    \STATE set requires\_grad to False on $\alpha, \beta, \delta,$ True on $\theta$
                    \STATE \textcolor{gray}{$s = \mathrm{softmax}(f_\theta(x_t))$}
                    \STATE \textcolor{gray}{calculate train loss $\mathcal{L}_{W} =$} $\mathcal{W}(s, \widetilde{y}_t)$\textcolor{gray}{$\cdot \ell(f_{\theta}(x_t), \widetilde{y}_t)$}
                    \STATE \textcolor{gray}{$\mathcal{L}_{W}$.backward()}
                    \STATE \textcolor{gray}{optimizer.step() to update $\theta$}
                    \STATE \textcolor{gray}{optimizer.zero\_grad()}
                    \STATE choose a random batch $(x_v, \widetilde{y}_v)$ $\gets$ dataloader($\mathcal{D}_{v}$)
                    \STATE set requires\_grad to True on $\alpha, \beta, \delta$, False on $\theta$
                    \STATE $s = \mathrm{softmax}(f_\theta(x_v))$
                    \STATE calculate val loss $\mathcal{L}_{W} = \mathcal{W}(s, \widetilde{y}_v)$$\cdot \ell(f_{\theta}(x_v), \widetilde{y}_v)$
                    \STATE $\mathcal{L}_{W}$.backward()
                    \STATE $\alpha = \alpha - \alpha_{lr} * (\nabla_{\alpha}\mathcal{L}_{W} + \alpha_{wd} * \alpha)$
                    \STATE $\beta = \beta - \beta_{lr} * (\nabla_{\beta}\mathcal{L}_{W} + \beta_{wd} * \beta)$
                    \STATE $\delta = \delta - \delta_{lr} * (\nabla_{\delta}\mathcal{L}_{W} + \delta_{wd} * \delta)$
                    \STATE project $(\alpha, \beta, \delta)$ to satisfy $\alpha \geq \delta \geq \beta \geq 1$
                    \STATE $\nabla_{\alpha}\mathcal{L}_{W} = \nabla_{\beta}\mathcal{L}_{W} = \nabla_{\delta}\mathcal{L}_{W} = 0$
                \ENDFOR
            \ENDFOR
            \STATE \textbf{return} $f_\theta(x)$
\end{algorithmic}
\label{algo}
\end{algorithm}

During training (see Algorithm~\ref{algo}, which only adds/modifies around 10 extra lines to standard training), we aim to find the model parameters $\theta^*$ that best minimize LiLAW loss on the training set using the parameters $\alpha^*, \beta^*, \delta^*$ that best minimize LiLAW loss on the validation set to get the objective:
\begin{align}
    \displaystyle\theta^* = \arg\min_{\theta} \sum_{(x_i, \widetilde{y_i}) \in \mathcal{D}_{t}}\mathcal{L}_{W_{\alpha^*, \beta^*, \delta^*}} \text{ such that }
    \alpha^*, \beta^*, \delta^* = \displaystyle\arg\min_{\alpha, \beta, \delta} \sum_{(x_i, \widetilde{y_i}) \in \mathcal{D}_{v}}\mathcal{L}_{W_{\theta^*}}
\end{align}
Note that $\mathcal{L}_{W_{\alpha^*, \beta^*, \delta^*}}$ is the loss computed using parameters $\alpha^*, \beta^*, \delta^*$ and $\mathcal{L}_{W_{\theta^*}}$ is the loss computed using the network parameters $\theta^*$. This objective is similar to the formulation in~\citet{jain_improving_2024}; however, we do not require an instance-wise weighting function, but a more general weighting function based on $\alpha, \beta, \delta$. Unlike bilevel meta-reweighting methods such as MOLERE~\citep{jain_improving_2024}, LRW~\citep{ren_learning_2018}, and MAML~\citep{finn_model-agnostic_2017}, LiLAW does not try to improve unweighted validation loss after a weighted training update, but instead, after each weighted training update, freezes the model parameters and directly updates only $\alpha,\beta,\delta$ using a weighted validation loss.

Prior meta-reweighting work~\citep{ren_learning_2018} assumed a small clean validation set to steer training towards the clean-label objective. Our setting is different and more realistic: we explicitly allow the validation set to contain label noise, because a clean set is often unavailable in practice. Using a noisy training set and a noisy validation set, as in LiLAW, is well‑motivated, as demonstrated by the theoretical and empirical results of~\citet{chen_robustness}. Specifically, they show that during training, maximizing accuracy over a sufficiently large noisy dataset leads to an approximately optimal classifier. They show that using a noisy validation set is reasonable because the accuracy on held‑out noisy samples is an unbiased estimator of the accuracy under the noisy data distribution.

We prove a corollary to Theorems~1, 2 in~\citet{gui2021towards} (Theorems~\ref{thma1},~\ref{thma2} in~\ref{proof:diag} for completeness), to show that diagonally dominant noise is sufficient for LiLAW.

\renewcommand{\thetheorem}{1}
\begin{corollary}
\label{corr}
Let $c\ge 2$. Assume label noise with transition matrix
$T\in[0,1]^{c\times c}$ where $T_{ij} := \mathbb{P}(\tilde y=j \mid y=i)$.
Assume the diagonally dominant condition from Theorems~\ref{thma1},~\ref{thma2} in \citet{gui2021towards}:
\begin{equation}
\label{eq:dd}
T_{ii} > \max\Bigl\{\max_{j\neq i} T_{ij},\ \max_{j\neq i} T_{ji}\Bigr\}\quad \forall i\in\{0,\dots,c-1\}.
\end{equation}
Let $(x,\tilde y)$ be an input and observed label pair, let $f^*(x)\in\{0,\dots,c-1\}$ be the true label function (referred to as the target concept in~\citet{gui2021towards}), let $g^*(x)$ be the deep neural network minimizing the expected loss, and let $s=\mathrm{softmax}(g^*(x))$.
For LiLAW parameters $(\alpha,\beta,\delta)$, we define the per-sample LiLAW weight as before with $\mathcal{W}_\alpha, \mathcal{W}_\beta, \mathcal{W}_\delta$. Fix any observed label $\tilde{y}$ and consider two points $(x_c,\tilde y)$ and $(x_n,\tilde y)$
with the same observed label, where $f^*(x_c)=\tilde y$ (clean label) and
$f^*(x_n)=i\neq \tilde y$ (noisy label).
Then, under diagonally dominant noise \eqref{eq:dd}, the following hold:
\begin{enumerate}[label=(\roman*),leftmargin=*]
\item \textbf{(Loss separation)} unweighted clean label loss is less than unweighted noisy label loss:
\begin{equation}
\ell(g^*(x_c),\tilde y) < \ell(g^*(x_n),\tilde y)
\end{equation}
\item \textbf{(LiLAW geometry separation)} disagreement is 0 for clean labels and $> 0$ for noisy labels:
\begin{equation}
m(x,\tilde y) = \max(s) - s[\tilde y] \implies m(x_c,\tilde y)=0 \wedge m(x_n,\tilde y)=T_{ii}-T_{i\tilde y}>0
\end{equation}
\item \textbf{(LiLAW weights separation)} if $\alpha\ge 1$ and $\beta\ge 1$, then:
\begin{equation}
\mathcal{W}_{\alpha}(s(x_c),\tilde y) \;>\; \mathcal{W}_{\alpha}(s(x_n),\tilde y)
\quad\text{and}\quad
\mathcal{W}_{\beta}(s(x_c),\tilde y) \;<\; \mathcal{W}_{\beta}(s(x_n),\tilde y).
\end{equation}
\end{enumerate}
Thus, diagonal dominance is sufficient to guarantee that \(s[\tilde y]\), \(\max(s)\), and the disagreement gap $m(x,\tilde y)$ contain a strict clean vs. noisy separation signal for samples sharing the same observed label. Moreover, when \(\alpha,\beta \geq 1\), the monotone LiLAW components \(W_\alpha\) and \(W_\beta\) preserve this ordering. The localized component \(W_\delta\) instead acts as a smooth transition-band weight.
\end{corollary}
\textit{Proof sketch.}
    Under diagonally dominant noise, the classifier assigns a higher score to the observed label when that label is actually correct than when it is corrupted. So, clean samples sharing the same observed label have lower cross-entropy loss and zero disagreement, while noisy samples have higher loss and positive disagreement. Since $\mathcal{W}_\alpha$ and $\mathcal{W}_\beta$ are monotone functions of these confidence / disagreement quantities, LiLAW preserves this ordering, yielding a strict clean vs. noisy separation signal for samples with the same observed label. Full proof is in~\ref{proof:diag}.

\iffalse
We generally see that:
\vspace{-0.5em}
\begin{enumerate}[leftmargin=*,labelindent=0pt,label=\textbf{--}]
    \itemsep-0.2em 
    \item $\mathcal{W}_\alpha$ is high when $\alpha$ is large or when $s_i[\widetilde{y_i}] = \max(s_i)$ and $\max(s_i)$ is high \\(agrees confidently = \textit{easy})
    \item $\mathcal{W}_\beta$ is high when $\beta$ is small or when $s_i[\widetilde{y_i}] < \max(s_i)$ and $\max(s_i)$ is high \\(disagrees confidently = \textit{hard})
    \item $\mathcal{W}_\delta$ is high when $\delta$ is large or when $\max(s_i)$ is low \\(agrees or disagrees with low confidence = \textit{moderate})
\end{enumerate}
\fi

%We also extend the results from~\citet{gui2021towards}'s work to prove that diagonally dominant noise is sufficient for LiLAW to separate clean and noisy labels in~\ref{proof:diag}.

\section{Experiments}
\label{sec:experiments}

We conducted extensive experiments with and without LiLAW to comprehensively assess its ability to boost test performance. We used H-CAT~\citep{seedat_dissecting_2024}, an API interface for several hardness and data characterization techniques. We emphasize that our goal is not to achieve state-of-the-art results, but to demonstrate the effectiveness of LiLAW in various settings. % To this end, we make the following considerations detailed below.

We show how the weight functions (2), (3), and (4) change with respect to $\alpha, \beta, \delta$ in~\ref{supp:derivative}. We also show that training with and without LiLAW share similar time complexity in~\ref{supp:timecomplexity} and space complexity in~\ref{supp:spacecomplexity} with minimal overhead for one additional backward pass on the validation mini-batch after every training mini-batch. We also extend LiLAW to be used for multi-label classification in~\ref{sec:multilabel} and regression in~\ref{sec:regression}. We compare the three formulations in~\ref{comparison}.

\subsection{Datasets}
\label{sec:datasets}

We use the CIFAR-100 dataset~\citep{cifar100} (and CIFAR-10~\citep{cifar10}, FashionMNIST~\citep{fashionmnist}, MNIST~\cite{deng2012mnist} in~\ref{supp:moredatasets}), which are all for research use, but we do not use the full training sets. Instead, we reserve 15\% of the training set as the validation set to evaluate and adjust our model's performance, as required by our method. We call our modified datasets, CIFAR-100-M, CIFAR-10-M, FashionMNIST-M, and MNIST-M, respectively. We also use Clothing-1M in~\ref{supp:clothing1m_tradeoff}, which is a real-world noisy dataset and for research use. To demonstrate generalizability and applicability to the medical domain, we evaluate LiLAW on ten 2D datasets with different medical imaging modalities focusing on multi-class classification tasks from MedMNISTv2~\citep{medmnistv1,medmnistv2,doerrich2024rethinking}, which are all CC BY 4.0, at varying noise levels in Section~\ref{sec:medical_imaging}. We discuss more datasets in~\ref{sec:synthetic},~\ref{fairness}.

%We also demonstrate strong results on a non-imaging dataset, ECG5000, which contains time-series data for heartbeat classification with 5 highly imbalanced classes.

\subsection{Models}
\label{sec:models}
We use models with a varying number of parameters to demonstrate the utility of LiLAW: ViT-Base-16-224~\citep{vit} for general imaging (CIFAR-100-M, CIFAR-10-M, FashionMNIST-M, MNIST-M) and ResNet-18~\citep{resnet} for medical imaging (ten 2D MedMNISTv2 datasets), since it achieves the highest average accuracy of all the benchmark models in~\citet{medmnistv2}. Further implementation details are in~\ref{imp} and models for synthetic / augmented data are discussed in~\ref{sec:synthetic},~\ref{fairness}.

\subsection{Evaluation}
\label{sec:evaluation}

We study the LiLAW extensively with various levels of injected noise with symmetric and asymmetric noise (see Table \ref{table:noisetypes}) and with ten MedMNISTv2 datasets (see Table \ref{table:medmnist} and Figure \ref{fig:med1}). In the Appendix, we show that LiLAW is not highly sensitive to $\alpha, \beta, \delta$ initialization (see~\ref{supp:abd_init}, Table \ref{tab:abd_sensitivity}), evaluate with and without calibration (see~\ref{supp:calibration}, Table \ref{table:calibration}), with and without a clean validation set (see~\ref{supp:cleanvalset}, Table \ref{table:cleanval}), and with various random seeds (see~\ref{supp:randseed}, Table \ref{table:seeds}).  We show that LiLAW is effective with different loss functions (see~\ref{supp:loss}, Table \ref{table:losses}), validation set sizes (see~\ref{supp:valsize}, Table \ref{table:valsize}), validation set noise (see~\ref{supp:valnoise}, Table \ref{tab:valnoise}), and with and without the same validation and training distribution (see~\ref{supp:valdist}, Table \ref{table:valdist}). We report results on four general imaging datasets (see~\ref{supp:moredatasets}, Table \ref{table:datasets}) and report the AUROC results (see~\ref{supp:medmnist}, Table~\ref{table:medmnist2}) for MedMNISTv2. We also provide results on using LiLAW with synthetic and augmented datasets (see~\ref{sec:pain}, Table~\ref{pain} for pain detection,~\ref{sec:gait}, Table~\ref{tab:gait} for gait classification, and~\ref{sec:heart}, Table~\ref{tab:ecg} for heartbeat classification), and provide fairness results on the Adult dataset along with a proof of why LiLAW helps in~\ref{fairness}.

%To comprehensively understand LiLAW's impact, we evaluated gains in test accuracy (top-1 and top-5) and AUROC with and without LiLAW across different settings. For the selected MedMNISTv2 datasets, we report the test accuracy (top-1 only, since some have $<5$ classes) and AUROC.

\subsection{Performance Comparison: General Imaging}
\label{sec:general_imaging}
%In all of the following tables, the results show test performance without LiLAW along with the difference in performance, i.e. the improvement (\textcolor{Green}{$\uparrow$}) or deterioration (\textcolor{BrickRed}{$\downarrow$}) with LiLAW, respectively. 

\iffalse
\begin{figure}[htb!]
    \centering
    \includegraphics[scale=0.34]{ciftop1.png}
    \caption{Top-1 accuracy with LiLAW with linear probing on the last two layers, without LiLAW with linear probing on the last two layers, and without LiLAW with fine-tuning on the full network using different levels of symmetric noise on CIFAR-100-M.} % with a random seed of 42.}
    \label{fig:noiselevel}
\end{figure}
\fi

In Tables~\ref{table:noisetypes}, \ref{table:noiselevels}-\ref{table:cleanval}, and \ref{table:losses}-\ref{table:datasets}, each metric reports results for linear probing (in our case, fine-tuning only the last two layers) without LiLAW and the improvement (\textcolor{Green}{$\uparrow$}) / deterioration (\textcolor{BrickRed}{$\downarrow$}) with LiLAW. We discuss full fine-tuning without LiLAW in~\ref{supp:randseed}, Table~\ref{table:seeds}. For all the other datasets and settings, we report training from scratch or full fine-tuning without LiLAW and \textcolor{Green}{$\uparrow$}/\textcolor{BrickRed}{$\downarrow$} with LiLAW.

In Table~\ref{table:noisetypes}, LiLAW shows performance improvement for 50\% symmetric and asymmetric noise on CIFAR-100-M. In~\ref{supp:noiselevel}, Table~\ref{table:noiselevels}, we report results across noise levels from 0\% to 90\%. And in Table \ref{table:ablation}, we ablate the LiLAW parameters ($\alpha,\beta,\delta$) under varying noise levels to show that using all three parameters yields the highest performance gains, with higher gains at higher noise levels.

\begin{table}[ht!]
\centering
\vspace{-0.9em}
\caption{CIFAR-100-M on different noise types at 50\% noise level.}
\footnotesize{
\begin{tabular}{lccc}
\hline
\textbf{Noise Type} & \textbf{Top-1 Acc. (\%)} & \textbf{Top-5 Acc. (\%)} & \textbf{AUROC} \\
\hline
Symmetric  & 75.20 \textcolor{Green}{$\uparrow$} 1.35 & 92.61 \textcolor{Green}{$\uparrow$} 0.67 & 0.9811 \textcolor{Green}{$\uparrow$} 0.0013 \\
Asymmetric & 74.69 \textcolor{Green}{$\uparrow$} 1.44 & 92.26 \textcolor{Green}{$\uparrow$} 0.82 & 0.9792 \textcolor{Green}{$\uparrow$} 0.0017 \\
\hline
\label{table:noisetypes}
\end{tabular}}
\end{table}

\begin{table}[ht!]
\centering 
\vspace{-2.5em} 
\caption{Results of ablating parameters with different levels of symmetric noise on CIFAR-100-M. \label{table:ablation}} 
\footnotesize{\begin{tabular}{lccccc} \hline \textbf{Parameters Used} & \textbf{Noise Level (\%)} & \textbf{Top-1 Acc. (\%)} & \textbf{Top-5 Acc. (\%)} & \textbf{AUROC} \\ \hline 
\multirow{2}{*}{$\alpha, \beta, \delta$} 
& 0  & 80.85 \textcolor{BrickRed}{$\downarrow$} 0.01 & 96.07 \textcolor{Green}{$\uparrow$} 0.16 & 0.9888 \textcolor{BrickRed}{$\downarrow$} 0.0005 \\
& 50 & 75.20 \textcolor{Green}{$\uparrow$} 1.35 & 92.61 \textcolor{Green}{$\uparrow$} 0.67 & 0.9811 \textcolor{Green}{$\uparrow$} 0.0013 \\ \hline

\multirow{2}{*}{$\alpha, \beta$} 
& 0  & 80.85 \textcolor{Green}{$\uparrow$} 0.07 & 96.07 \textcolor{Green}{$\uparrow$} 0.06 & 0.9888 \textcolor{BrickRed}{$\downarrow$} 0.0002 \\
& 50 & 75.20 \textcolor{Green}{$\uparrow$} 0.35 & 92.61 \textcolor{Green}{$\uparrow$} 0.07 & 0.9811 \textcolor{Green}{$\uparrow$} 0.0001 \\ \hline

\multirow{2}{*}{$\alpha, \delta$} 
& 0  & 80.85 \textcolor{Green}{$\uparrow$} 0.01 & 96.07 \textcolor{BrickRed}{$\downarrow$} 0.01 & 0.9888 \textcolor{BrickRed}{$\downarrow$} 0.0008 \\
& 50 & 75.20 \textcolor{BrickRed}{$\downarrow$} 0.69 & 92.61 \textcolor{BrickRed}{$\downarrow$} 0.31 & 0.9811 \textcolor{BrickRed}{$\downarrow$} 0.0012 \\ \hline

\multirow{2}{*}{$\beta, \delta$} 
& 0  & 80.85 \textcolor{BrickRed}{$\downarrow$} 1.59 & 96.07 \textcolor{BrickRed}{$\downarrow$} 0.29 & 0.9888 \textcolor{BrickRed}{$\downarrow$} 0.0023 \\
& 50 & 75.20 \textcolor{BrickRed}{$\downarrow$} 0.14 & 92.61 \textcolor{Green}{$\uparrow$} 0.35 & 0.9811 \textcolor{Green}{$\uparrow$} 0.0005 \\ \hline

\multirow{2}{*}{$\alpha$} 
& 0  & 80.85 \textcolor{BrickRed}{$\downarrow$} 1.10 & 96.07 \textcolor{Green}{$\uparrow$} 0.05 & 0.9888 \textcolor{BrickRed}{$\downarrow$} 0.0001 \\
& 50 & 75.20 \textcolor{BrickRed}{$\downarrow$} 0.69 & 92.61 \textcolor{BrickRed}{$\downarrow$} 0.30 & 0.9811 \textcolor{BrickRed}{$\downarrow$} 0.0007 \\ \hline

\multirow{2}{*}{$\beta$} 
& 0  & 80.85 \textcolor{BrickRed}{$\downarrow$} 1.23 & 96.07 \textcolor{BrickRed}{$\downarrow$} 0.21 & 0.9888 \textcolor{BrickRed}{$\downarrow$} 0.0011 \\
& 50 & 75.20 \textcolor{BrickRed}{$\downarrow$} 0.69 & 92.61 \textcolor{BrickRed}{$\downarrow$} 0.31 & 0.9811 \textcolor{BrickRed}{$\downarrow$} 0.0007 \\ \hline

\multirow{2}{*}{$\delta$} 
& 0  & 80.85 \textcolor{BrickRed}{$\downarrow$} 1.90 & 96.07 \textcolor{BrickRed}{$\downarrow$} 0.26 & 0.9888 \textcolor{BrickRed}{$\downarrow$} 0.0026 \\
& 50 & 75.20 \textcolor{BrickRed}{$\downarrow$} 0.68 & 92.61 \textcolor{BrickRed}{$\downarrow$} 0.30 & 0.9811 \textcolor{BrickRed}{$\downarrow$} 0.0007 \\ \hline \end{tabular}} \end{table}
\vspace{-0.5em}
\subsection{Performance Comparison: Medical Imaging}
\label{sec:medical_imaging}
In Table~\ref{table:medmnist} and Figure~\ref{fig:med1}, we see the effect of LiLAW across ten 2D datasets from MedMNISTv2 with varying levels of symmetric label noise (see~\ref{supp:medmnist}, Table \ref{table:medmnist2}, Figure \ref{fig:med2} for AUROC metrics). We see an increase in accuracy and/or AUROC in nearly all cases with LiLAW. Given the noisiness of medical data, LiLAW's ability to enhance performance under such conditions is highly valuable, especially since we are full fine-tuning with ImageNet-1K pretraining.

\begin{table*}[ht!]
\centering
\caption{Accuracy on ten 2D datasets from MedMNISTv2 with different levels of symmetric noise with full fine-tuning without LiLAW and with LiLAW. \label{table:medmnist}}
\resizebox{\textwidth}{!}{
{
\begin{tabular}{l|cccccc}
\hline
 & & & \textbf{Acc. (\%)} & & & \\
\hline
\textbf{Dataset} & \textbf{0\% Noise} & \textbf{10\% Noise} & \textbf{20\% Noise} & \textbf{30\% Noise} & \textbf{40\% Noise} & \textbf{50\% Noise} \\
\hline
PathMNIST & 95.49 \textcolor{Green}{$\uparrow$} 0.08 & 95.38 \textcolor{BrickRed}{$\downarrow$} 0.08 & 94.70 \textcolor{Green}{$\uparrow$} 0.20 & 94.19 \textcolor{BrickRed}{$\downarrow$} 0.59 & 94.01 \textcolor{Green}{$\uparrow$} 0.18 & 92.87 \textcolor{Green}{$\uparrow$} 0.69 \\
\hline
DermaMNIST & 79.88 \textcolor{Green}{$\uparrow$} 0.88 & 76.12 \textcolor{Green}{$\uparrow$} 1.80 & 74.38 \textcolor{Green}{$\uparrow$} 2.13 & 73.92 \textcolor{Green}{$\uparrow$} 0.27 & 71.89 \textcolor{Green}{$\uparrow$} 1.46 & 68.03 \textcolor{Green}{$\uparrow$} 4.52 \\
\hline
OCTMNIST & 75.73 \textcolor{Green}{$\uparrow$} 0.65 & 72.84 \textcolor{Green}{$\uparrow$} 4.74 & 71.42 \textcolor{Green}{$\uparrow$} 1.97 & 67.11 \textcolor{Green}{$\uparrow$} 3.44 & 68.19 \textcolor{Green}{$\uparrow$} 4.94 & 60.79 \textcolor{Green}{$\uparrow$} 7.07 \\
\hline
PneumoniaMNIST & 89.04 \textcolor{Green}{$\uparrow$} 2.61 & 85.40 \textcolor{Green}{$\uparrow$} 1.10 & 86.65 \textcolor{BrickRed}{$\downarrow$} 0.41 & 85.40 \textcolor{Green}{$\uparrow$} 4.18 & 81.50 \textcolor{Green}{$\uparrow$} 2.90 & 84.24 \textcolor{Green}{$\uparrow$} 2.14 \\
\hline
BreastMNIST & 85.83 \textcolor{Green}{$\uparrow$} 1.39 & 83.87 \textcolor{BrickRed}{$\downarrow$} 1.17 & 79.13 \textcolor{Green}{$\uparrow$} 2.96 & 78.74 \textcolor{Green}{$\uparrow$} 0.39 & 74.78 \textcolor{Green}{$\uparrow$} 6.75 & 70.10 \textcolor{Green}{$\uparrow$} 4.34 \\
\hline
BloodMNIST & 98.50 \textcolor{Green}{$\uparrow$} 0.16 & 97.24 \textcolor{Green}{$\uparrow$} 0.18 & 97.36 \textcolor{Green}{$\uparrow$} 0.21 & 96.93 \textcolor{BrickRed}{$\downarrow$} 0.28 & 95.72 \textcolor{Green}{$\uparrow$} 0.71 & 94.72 \textcolor{Green}{$\uparrow$} 0.42 \\
\hline
TissueMNIST & 68.74 \textcolor{BrickRed}{$\downarrow$} 1.33 & 65.02 \textcolor{Green}{$\uparrow$} 0.85 & 61.33 \textcolor{Green}{$\uparrow$} 1.17 & \hspace{0.5em}45.08 \textcolor{Green}{$\uparrow$} 19.16 & \hspace{0.5em}38.50 \textcolor{Green}{$\uparrow$} 13.22 & \hspace{0.5em}31.82 \textcolor{Green}{$\uparrow$} 20.41 \\
\hline
OrganAMNIST & 94.87 \textcolor{Green}{$\uparrow$} 0.18 & 94.19 \textcolor{Green}{$\uparrow$} 0.22 & 94.02 \textcolor{Green}{$\uparrow$} 0.18 & 90.76 \textcolor{Green}{$\uparrow$} 2.73 & 92.50 \textcolor{Green}{$\uparrow$} 0.52 & 89.45 \textcolor{Green}{$\uparrow$} 1.63 \\
\hline
OrganCMNIST & 90.66 \textcolor{Green}{$\uparrow$} 0.93 & 88.85 \textcolor{Green}{$\uparrow$} 1.01 & 85.37 \textcolor{Green}{$\uparrow$} 0.30 & 84.45 \textcolor{Green}{$\uparrow$} 2.73 & 84.19 \textcolor{Green}{$\uparrow$} 0.87 & 82.95 \textcolor{Green}{$\uparrow$} 1.24 \\
\hline
OrganSMNIST & 80.54 \textcolor{Green}{$\uparrow$} 0.49 & 78.28 \textcolor{Green}{$\uparrow$} 0.66 & 75.34 \textcolor{Green}{$\uparrow$} 2.16 & 72.54 \textcolor{Green}{$\uparrow$} 3.35 & 72.95 \textcolor{Green}{$\uparrow$} 1.42 & 70.82 \textcolor{Green}{$\uparrow$} 2.65 \\
\hline
\end{tabular}}}
\end{table*}

%Table~\ref{tab:ecg5000_results} highlights the effectiveness of LiLAW on ECG5000. Compared to standard training, LiLAW yields a substantial improvement in accuracy and AUROC, demonstrating its ability to enhance predictive performance and discrimination in inherently noisy domains. Namely, LiLAW can dynamically reweight samples based on difficulty to effectively mitigate the impact of inherent noise in physiological signals. This suggests that LiLAW provides a generalizable framework to improve robustness across modalities where label quality and data heterogeneity are common challenges.

\subsection{Performance Comparison: Identifying Mislabels}
\label{sec:mislabels}
In~\ref{supp:comp}, Table \ref{table:comp}, $\mathcal{W}_\alpha$ achieves the best AUROC at both early-stage and late-stage training (0.9838 at epoch 3, 0.9782 at epoch 10), and the best AUPRC at both stages (0.9810 at epoch 3, 0.9755 at epoch 10), comfortably outperforming several established methods for difficulty estimation, including Data-IQ~\citep{seedat_data-iq_2022}, DataMaps~\citep{swayamdipta_dataset_2020}, CNLCU-S~\citep{xia_sample_2022}, AUM~\citep{pleiss_identifying_2020}, EL2N~\citep{paul2021deep}, GraNd~\citep{paul2021deep}, and Forgetting~\citep{toneva_empirical_2019}. $\mathcal{W}_\beta$ attains strong AUROC but weaker AUPRC since it ranks many mislabeled points highly but also assigns high scores to hard yet correctly labeled samples, while $\mathcal{W}_\delta$ improves from early to late training, consistent with identifying samples that become easier during training.

\subsection{Performance Comparison: Noisy-Label Learning}
\label{sec:noisy_label_learning}
Using the Clothing-1M dataset \citep{xiao2015learning}, a real-world noisy dataset, we evaluate LiLAW using ResNet-50 with ImageNet-1K pretraining. The same training method used in the 16 baseline methods in~\ref{supp:clothing1m_tradeoff}, Table~\ref{tab:clothing1m_complexity} was used with LiLAW. We also report the cost-performance trade-off, showing a +2.23 point gain over cross-entropy on Clothing-1M while retaining the asymptotic time and space complexity as standard training. Existing noisy-label learning methods usually require substantially heavier training pipelines, pruning (worsening fairness), more data, or more models and are not directly comparable to LiLAW. LiLAW achieves a competitive 71.44\% accuracy, while maintaining a lightweight design, suggesting its practicality and scalability for real-world noisy-label settings.

\iffalse
\begin{table}[ht!]
\centering
\begin{footnotesize}
\begin{tabular}{l|c}
\toprule
\textbf{Method} & \textbf{Accuracy (\%)} \\
\midrule
Cross-entropy & 69.21 \\
\midrule
Backward \citep{patrini2016making} & 69.13 \\
GCE \citep{zhang2018generalized} & 69.75 \\
Forward \citep{patrini2016making} & 69.84 \\
Co-teaching \citep{han2018co} & 70.15 \\
SEAL \citep{chen2021beyond} & 70.63 \\
SL \citep{wang2019symmetric} & 71.02 \\
Weakly Supervised \citep{zhang2019metacleaner} & 71.36 \\
LRT \citep{zheng2020error} & 71.74 \\
Joint-Optim \citep{tanaka2018joint} & 72.16 \\
MixNN \citep{lu2024mitigating} & 72.39 \\
MetaCleaner \citep{zhang2019metacleaner} & 72.50 \\ 
\midrule
LiLAW (ours) & 71.44 \\
\bottomrule
\end{tabular}
\caption{Noisy-label learning methods on Clothing-1M.}
\label{tab:noisy_label_results}
\end{footnotesize}
\vspace{-2.1em}
\end{table}
\fi

\section{Limitations} \label{sec:limitations} 

LiLAW can help in settings where labels are noisy, scarce, heterogeneous, or partly synthetic, which is common in medical ML and other resource-constrained domains. However, there are limitations. First, LiLAW is deliberately lightweight. This is a strength, but means that it may sometimes not capture arbitrary weighting rules not fully reflected in confidence and disagreement. So, although LiLAW can help mitigate bias empirically, it should not be viewed as sufficient justification for autonomous deployment in high-stakes settings. Second, our clean vs. noisy separation relies on diagonally dominant noise, which may not cover every real-world noise setting. Clothing-1M is \textbf{not} diagonally dominant in at least 2-3 classes~\cite{hedderich2021analysing}. Even so, our gains remain competitive.

\section{Conclusion}
\label{sec:conclusion}
In this work, we introduced LiLAW, a simple, lightweight, adaptive weighting method that improves training under noisy and heterogeneous conditions. By learning only three parameters that evolve with sample difficulty, LiLAW can surface difficult samples and be easily integrated into existing pipelines with minimal overhead, which is especially useful for data- and resource-constrained settings. We highlight practical utility using medical, time-series, and synthetic data. It can also help with fairness and imbalance and open up opportunities for active learning (select difficult samples), continual learning (stabilize drift), and semi-supervised learning (reweight pseudo-labels). 

\newpage

%\section*{Impact Statement}
%This paper presents work whose goal is to advance the field of machine learning. There are many potential societal consequences of our work, none of which we feel must be specifically highlighted here.

%\section*{Acknowledgements}
%We acknowledge the support of the Vector Institute for Artificial Intelligence, Ontario Graduate Scholarships, the Canadian Institutes of Health Research, the Natural Sciences and Engineering Research Council of Canada, and the University of Toronto.

%\section*{Ethics Statement}
%All datasets used in this work are publicly available and widely used for research purposes. This work does not present any known ethical concerns.

%\section*{Reproducibility Statement}

%The details provided about Datasets (see Section~\ref{sec:datasets}) and Models \& Implementation (see Section~\ref{sec:models}) along with the code provided in the Supplementary Material make this work reproducible.

\bibliography{neurips_refs}
\bibliographystyle{plainnat}

\newpage
%%%%%%%%%%%%%%%%%%%%%%%%%%%%%%%%%%%%%%%%%%%%%%%%%%%%%%%%%%%%

\appendix
\label{supp}

\section{Technical appendices and supplementary material}

\setcounter{section}{1}

\subsection{Motivating example}
\label{supp:example}

\begin{table}[ht] \centering 
\caption{Comparison of cross-entropy (CE) and LiLAW-weighted loss under two $\alpha,\beta,\delta$ settings.}
\label{tab:lilaw_toy}
\setlength{\tabcolsep}{3pt}
\renewcommand{\arraystretch}{1.15}
\resizebox{\textwidth}{!}{
\begin{tabular}{@{}lccccrrrrrrrrrr@{}}
\toprule
\multirow{2}{*}{Case ($\tilde y=0$)} 
& \multirow{2}{*}{Prediction} 
& \multirow{2}{*}{$s_{\tilde y}$} 
& \multirow{2}{*}{$s_{\max}$} 
& \multirow{2}{*}{CE} 
& \multicolumn{5}{c}{$\alpha=12,\ \delta=2,\ \beta=1$} 
& \multicolumn{5}{c}{$\alpha=8,\ \delta=6,\ \beta=5$} \\
\cmidrule(lr){6-10} \cmidrule(lr){11-15}
& & & & 
& $\mathcal{W}_\alpha$ & $\mathcal{W}_\beta$ & $\mathcal{W}_\delta$ & $W$ & $\mathcal{L}_W$
& $\mathcal{W}_\alpha$ & $\mathcal{W}_\beta$ & $\mathcal{W}_\delta$ & $W$ & $\mathcal{L}_W$ \\
\midrule
Correct \& Confident   
& $[0.95,0.05]$ & 0.95 & 0.95 & \textbf{0.051} 
& 1.000 & 0.500 & 0.637 & 2.137 & \textbf{0.110}
& 0.999 & 0.022 & 0.000 & 1.021 & \textbf{0.052} \\

Correct \& Low confidence 
& $[0.60,0.40]$ & 0.60 & 0.60 & \textbf{0.511} 
& 0.999 & 0.500 & 0.835 & 2.334 & \textbf{1.192}
& 0.985 & 0.083 & 0.011 & 1.080 & \textbf{0.551} \\

Incorrect \& Low confidence 
& $[0.40,0.60]$ & 0.40 & 0.60 & \textbf{0.916} 
& 0.985 & 0.550 & 0.980 & 2.515 & \textbf{2.305}
& 0.931 & 0.198 & 0.198 & 1.327 & \textbf{1.216} \\

Incorrect \& Confident 
& $[0.05,0.95]$ & 0.05 & 0.95 & \textbf{2.996} 
& 0.413 & 0.711 & 0.697 & 1.821 & \textbf{5.456}
& 0.366 & 0.668 & 0.810 & 1.844 & \textbf{5.523} \\
\bottomrule
\end{tabular}}
\end{table}
As shown in Table~\ref{tab:lilaw_toy}, CE assigns the same loss to a sample whenever the model produces the same prediction, whereas LiLAW can change the sample's contribution through the learned parameters $(\alpha,\delta,\beta)$. Under the initial setting $(\alpha,\beta,\delta)=(12,1,2)$, LiLAW amplifies the contribution of low confidence and ambiguous examples: the correct and low confidence and incorrect and low confidence cases receive much larger weighted losses than their CE values alone. After the transition to $(\alpha,\beta,\delta)=(8,5,6)$, while preserving $\alpha \geq \delta \geq \beta$, the weighted losses of the correct/confident, correct/low confidence, and incorrect/low confidence examples decrease substantially, from $0.110$ to $0.052$, $1.192$ to $0.551$, and $2.305$ to $1.216$, respectively. In contrast, the confidently incorrect case remains heavily penalized, and its weighted loss slightly increases from $5.456$ to $5.523$. This illustrates the main adaptive behavior of LiLAW: rather than using a fixed curriculum or relying only on the magnitude of CE, LiLAW can first allocate more training signal to low-confidence boundary examples and later reduce their relative influence while preserving strong gradients on high-confidence disagreements with the observed label. This encourages the model to resolve ambiguous samples early and to continue correcting confident mistakes later, without using a hard sample-selection or pruning rule, while still keeping the weighting smooth, differentiable, and sample-dependent.

\subsection{Diagonally Dominant Noise}
\label{proof:diag}

Learning a reliable model under label noise with no assumptions about the noise transition matrix, which describes how labels are corrupted, is usually unrealistic. That said, \textit{our method does not require estimating this matrix or using it directly}. It only assumes a mild property of the noise transition matrix. A common assumption is that the noise is diagonally dominant~\citep{han2018co, yu2019does}, meaning that for each class, a sample is more likely to keep its true label than to be flipped to any one other class. This is different from assuming that the noise rate is below 50\% for every class. Even if more than half of the labels are wrong in a given class, the correct label can still be the most likely outcome when mistakes are spread across many different wrong classes. 

Under diagonally dominant noise, even when the learned model is only close to the optimal noisy-data classifier, \citet{gui2021towards} prove that examples whose observed label is actually correct tend to align better with the input patterns, so a reasonably trained model will usually assign them lower loss than mislabeled examples that share the same observed label. In practice, pruning can find a mostly clean subset and loss-based reweighting can downweight mostly clean samples and upweight noisier samples to potentially correct them. This is why both pruning-based and reweighting-based approaches often perform well under diagonally dominant noise.

\subsubsection{Proof that diagonally dominant noise is sufficient for LiLAW}

% Sigmoid:
\newcommand{\sigmoid}{\sigma}
\newcommand{\CE}{\ell_{\mathrm{CE}}}

We prove Corollary~\ref{corr} in Section~\ref{sec:method}, to Theorems~\ref{thma1}, \ref{thma2} from~\citet{gui2021towards}, with a similar setup, to show that diagonally dominant noise is sufficient for LiLAW. We rewrite all three below.

\renewcommand{\thetheorem}{\arabic{theorem}}
\begin{theorem}
\label{thma1}
Let $g^*$ denote the deep neural network minimizing the cross-entropy loss in Eq.~(1),
$(\mathbf{x}_1,\tilde{y})$ and $(\mathbf{x}_2,\tilde{y})$ are any two examples
with the same observed label $\tilde{y}$ in $\widetilde{D}$ satisfying that
$f^*(\mathbf{x}_1)=\tilde{y}$ and $f^*(\mathbf{x}_2)\neq \tilde{y}$.
If $T$ satisfies the diagonally-dominant condition:
\[
T_{ii} >
\max\left\{
\max_{j\neq i} T_{ij},\,
\max_{j\neq i} T_{ji}
\right\},
\quad \forall i,
\]
then
$
\ell_{\mathrm{CE}}\bigl(g^*(\mathbf{x}_1),\tilde{y}\bigr)
<
\ell_{\mathrm{CE}}\bigl(g^*(\mathbf{x}_2),\tilde{y}\bigr).
$
\end{theorem}

\begin{theorem}
\label{thma2}
Suppose $g$ is $\epsilon$-close to $g^*$, i.e.,
$\|g-g^*\|_{\infty}=\epsilon$. For two examples
$(\mathbf{x}_1,\tilde{y})$ and $(\mathbf{x}_2,\tilde{y})$, assume
$f^*(\mathbf{x}_1)=\tilde{y}$ and $f^*(\mathbf{x}_2)\neq \tilde{y}$.
If $T$ satisfies the diagonally-dominant condition:
\[
T_{ii} >
\max\left\{
\max_{j\neq i} T_{ij},\,
\max_{j\neq i} T_{ji}
\right\},
\quad \forall i,
\]
and $
\epsilon
<
\frac{1}{2}\cdot
\left(
T_{\tilde{y}\tilde{y}}
-
T_{f^*(\mathbf{x}_2)\tilde{y}}
\right),
$
then $\ell_{\mathrm{CE}}\bigl(g(\mathbf{x}_1),\tilde{y}\bigr)
<
\ell_{\mathrm{CE}}\bigl(g(\mathbf{x}_2),\tilde{y}\bigr).$
\end{theorem}

\renewcommand{\thetheorem}{1}

\begin{corollary}
Let $c\ge 2$. Assume label noise with transition matrix
$T\in[0,1]^{c\times c}$ where $T_{ij} := \mathbb{P}(\tilde y=j \mid y=i)$.
Assume the diagonally dominant condition from Theorems~\ref{thma1}, \ref{thma2} in \citet{gui2021towards}:
\begin{equation}\tag{8}
T_{ii} > \max\Bigl\{\max_{j\neq i} T_{ij},\ \max_{j\neq i} T_{ji}\Bigr\},\quad \forall i\in\{0,\dots,c-1\}.
\end{equation}
Let $(x,\tilde y)$ be an input and observed label pair, let $f^*(x)\in\{0,\dots,c-1\}$ be the true label function (referred to as the target concept in~\citet{gui2021towards}), let $g^*(x)$ be the deep neural network minimizing the expected loss, and let $s=\mathrm{softmax}(g^*(x))$.
For LiLAW parameters $(\alpha,\beta,\delta)$, we define the per-sample LiLAW weight as before with $\mathcal{W}_\alpha, \mathcal{W}_\beta, \mathcal{W}_\delta$. Fix any observed label $\tilde{y}$ and consider two points $(x_c,\tilde y)$ and $(x_n,\tilde y)$
with the same observed label, where $f^*(x_c)=\tilde y$ (clean label) and
$f^*(x_n)=i\neq \tilde y$ (noisy label).
Then, under diagonally dominant noise \eqref{eq:dd}, the following hold:
\begin{enumerate}[label=(\roman*),leftmargin=*]
\item \textbf{(Loss separation)} unweighted clean label loss is less than unweighted noisy label loss:
\begin{equation}\tag{9}
\ell(g^*(x_c),\tilde y) < \ell(g^*(x_n),\tilde y)
\end{equation}
\item \textbf{(LiLAW geometry separation)} disagreement is 0 for clean labels and $> 0$ for noisy labels:
\begin{equation}\tag{10}
m(x,\tilde y) = \max(s) - s[\tilde y] \quad\implies\quad m(x_c,\tilde y)=0 \wedge m(x_n,\tilde y)=T_{ii}-T_{i\tilde y}>0
\end{equation}
\item \textbf{(LiLAW weights separation)} if $\alpha\ge 1$ and $\beta\ge 1$, then:
\begin{equation}\tag{11}
\mathcal{W}_{\alpha}(s(x_c),\tilde y) \;>\; \mathcal{W}_{\alpha}(s(x_n),\tilde y)
\quad\text{and}\quad
\mathcal{W}_{\beta}(s(x_c),\tilde y) \;<\; \mathcal{W}_{\beta}(s(x_n),\tilde y).
\end{equation}
\end{enumerate}
Thus, diagonal dominance is sufficient to guarantee that \(s[\tilde y]\), \(\max(s)\), and the disagreement gap $m(x,\tilde y)$ contain a strict clean vs. noisy separation signal for samples sharing the same observed label. Moreover, when \(\alpha,\beta \geq 1\), the monotone LiLAW components \(W_\alpha\) and \(W_\beta\) preserve this ordering. The localized component \(W_\delta\) instead acts as a smooth transition-band weight.
\end{corollary}

\begin{proof}
Lemma~2 (used in the proofs of Theorems~\ref{thma1} and~\ref{thma2} in the Appendix of~\cite{gui2021towards}) states that the deep neural network $g^*$ minimizing the expected loss satisfies:
\begin{equation}
\label{eq:gstar-row}
s(g^*(x))[j] = T_{f^*(x),j} \forall j.
\end{equation}

\noindent\emph{(i)}
If $f^*(x_c)=\tilde y$, then by \eqref{eq:gstar-row}, $s(g^*(x_c))[\tilde y]=T_{\tilde y\tilde y}$.
If $f^*(x_n)=i\neq \tilde y$, then $s(g^*(x_n))[\tilde y]=T_{i\tilde y}$.

By \eqref{eq:dd}, we have $T_{\tilde y\tilde y} > T_{i\tilde y} \implies -\log T_{\tilde y\tilde y} < -\log T_{i\tilde y} \implies \ell(g^*(x_c),\tilde y) < \ell(g^*(x_n),\tilde y)$.

This is similar to the proof of Theorem~\ref{thma1} in the Appendix of~\cite{gui2021towards}.

\noindent\emph{(ii)}
For $x_c$ with true label $\tilde y$, \eqref{eq:gstar-row} and diagonal dominance \eqref{eq:dd} imply
\begin{equation}
\max_j s(g^*(x_c))[j] = \max_j T_{\tilde y j} = T_{\tilde y\tilde y} = s(g^*(x_c))[\tilde y] \implies m(x_c,\tilde y)=0
\end{equation}
For $x_n$ with true label $i\neq \tilde y$, \eqref{eq:gstar-row} and diagonal dominance \eqref{eq:dd} imply
\begin{equation}
\max_j s(g^*(x_n))[j] = \max_j T_{i j} = T_{ii} \text{ and } s(g^*(x_n))[\tilde y]=T_{i\tilde y} 
\end{equation}
\begin{equation}
\implies m(x_n,\tilde y)=T_{ii}-T_{i\tilde y}>0 \text{ since } T_{ii}>\max_{j\neq i}T_{ij}\ge T_{i\tilde y}
\end{equation}

\noindent\emph{(iii)}
Because the sigmoid function is strictly increasing, it suffices to compare its arguments.

Let $s_c=s(g^*(x_c))[\tilde y]=T_{\tilde y\tilde y}$, $s_n=s(g^*(x_n))[\tilde y]=T_{i\tilde y}$, and \\ $M_c=\max_j s(g^*(x_c))[j]=s_c$, $M_n=\max_j s(g^*(x_n))[j]=T_{ii}$.

Then,
\begin{equation}
\bigl(\alpha s_c - M_c\bigr) - \bigl(\alpha s_n - M_n\bigr)
= \alpha(s_c-s_n) - (M_c-M_n) \\
= (\alpha-1)(s_c-s_n) + (M_n-s_n).
\end{equation}
By \eqref{eq:dd}, $s_c>s_n$ and $M_n-s_n=T_{ii}-T_{i\tilde y}>0$, so for $\alpha\ge 1$, $\mathcal{W}_{\alpha}(s(x_c),\tilde y)>\mathcal{W}_{\alpha}(s(x_n),\tilde y)$.

Similarly,
\begin{equation}
\bigl(M_n-\beta s_n\bigr) - \bigl(M_c-\beta s_c\bigr)
= (M_n-M_c) - \beta(s_n-s_c) \\
= (M_n-s_n) + (\beta-1)(s_c-s_n).
\end{equation}
Again, we have $M_n-s_n>0$ and $s_c-s_n>0$, so for $\beta\ge 1$, $\mathcal{W}_{\beta}(s(x_c),\tilde y)<\mathcal{W}_{\beta}(s(x_n),\tilde y)$.

Since \emph{(i), (ii), and (iii)} hold, we have shown that it is possible to obtain a strict clean vs. noisy separation signal for samples with the same observed label, but different true labels using LiLAW under the condition that we have diagonally dominant noise.

\end{proof}

\subsection{Geometric interpretation of the weight functions} \label{supp:geometric_interpretation} 

LiLAW learns three global scalar parameters, $\alpha,\beta,\delta$, that parameterize a continuous per-sample weighting function. The parameters are global and shared, but the actual weight is different for every sample and changes throughout training because it depends on the sample's current confidence / disagreement coordinates, namely $s_i[\tilde y_i]$ and $\max(s_i)$.

In other words, LiLAW does not place samples into only three fixed difficulty categories, but defines a continuous weighting rule over the difficulty geometry. So, LiLAW captures instance-dependent difficulty. Two samples from the same class can receive very different weights if one is confidently correct and the other is low confidence or confidently incorrect. In~\ref{proof:diag}, we further show that, under diagonally dominant noise, the LiLAW geometry provides a strict clean vs. noisy separation signal for samples sharing the same observed label. Empirically, the $\mathcal{W}_\alpha$ weights also achieve strong AUROC / AUPRC for mislabeled-sample detection at the early and late training stages. 

LiLAW also captures class-dependent difficulty, but not through explicit class-specific parameters. The term $s_i[\tilde y_i]$ is already class-conditional since it is the model score assigned to the observed class of that sample. Therefore, classes that are noisier, more ambiguous, or more underrepresented occupy different regions of the $(s_i[\tilde y_i], \max(s_i))$ plane and are reweighted differently through the same shared mapping. This gives class-dependent behavior through the logits rather than through separate parameters for each class, keeping the method lightweight.

The 3 LiLAW components correspond to different geometries in the confidence/disagreement plane: 

\begin{enumerate}
\item a one-sided low-disagreement region, $\mathcal{W}_\alpha$;
\item a one-sided high-disagreement region, $\mathcal{W}_\beta$; and
\item a localized transition band, $\mathcal{W}_\delta$, around an intermediate region.
\end{enumerate}

The first two components are naturally modeled by smooth threshold functions, while the third requires a smooth localized bump. Thus, the functional forms arise from decomposing the difficulty plane into two half-space detectors and one band-pass detector. For $\mathcal{W}_\alpha$ and $\mathcal{W}_\beta$, the desired regions are one-sided inequalities corresponding to easy and hard regions. We need a function that is strictly monotone, always positive, bounded in $(0,1)$, and differentiable. A sigmoid is a natural choice for this role. This is important because the weight multiplies the loss, so positivity and boundedness prevent negative or unstable loss scaling. The sigmoid also preserves the clean vs. noisy ordering established by the confidence/disagreement geometry as shown in the previous section.

The role of $\mathcal{W}_\delta$ is fundamentally different. The moderate region is not a half-space. It is an interior transition band. For this component, we want a function with zero gradient at the center of the band, a sign change in the gradient across the band, bounded magnitude, and differentiability. A simple radial basis function is a natural choice for this localized band-pass behavior.

We also provide a geometric justification for the chosen functional forms. Let:
\begin{equation}
x_i = s_i[\tilde y_i],
\quad
m_i = \max(s_i) - s_i[\tilde y_i] \ge 0,
\end{equation}
where $x_i$ is the model confidence in the observed label and $m_i$ is the disagreement gap relative to the top predicted class. Then the three LiLAW terms can be rewritten as
\begin{equation}
W_{\alpha,i}
=
\sigma\big((\alpha-1)x_i - m_i\big),
\end{equation}
\begin{equation}
W_{\beta,i}
=
\sigma\big(m_i - (\beta-1)x_i\big),
\end{equation}
\begin{equation}
W_{\delta,i}
=
\exp\left(
-\frac{\big((\delta-1)x_i - m_i\big)^2}{2}
\right).
\end{equation}

In these coordinates, the easy, hard, and moderate sample regions are approximated by the following indicator sets:
\begin{equation}
\mathbf{1}\{m_i \le (\alpha-1)x_i\},
\quad
\mathbf{1}\{m_i \ge (\beta-1)x_i\},
\quad
\mathbf{1}\{|m_i - (\delta-1)x_i| \le \epsilon\},
\end{equation}
for some small $\epsilon>0$.

This explains why the three terms use different functional forms. The first two regions are naturally represented by one-sided threshold detectors, for which the sigmoid is a canonical smooth surrogate. The sigmoid guarantees positive, bounded, differentiable weights in $(0,1)$, which is important because the weight multiplies the loss. It also preserves the clean vs. noisy ordering derived in~\ref{proof:diag}. Under diagonally dominant noise, for two samples with the same observed label, $\tilde y$, a clean sample $(x_c,m_c)$ and a noisy sample $(x_n,m_n)$, the arguments of the sigmoid are ordered so that $\mathcal{W}_\alpha(x_c,m_c) > \mathcal{W}_\alpha(x_n,m_n)$ and $\mathcal{W}_\beta(x_c,m_c) < \mathcal{W}_\beta(x_n,m_n)$, for $\alpha,\beta\ge 1$.

The sigmoid also yields stable updates. Let:
\begin{equation}
z_{\alpha,i}=(\alpha-1)x_i-m_i,
\quad
z_{\beta,i}=m_i-(\beta-1)x_i.
\end{equation}
Then,
\begin{equation}
\frac{\partial W_{\alpha,i}}{\partial \alpha}
=
x_i\sigma(z_{\alpha,i})(1-\sigma(z_{\alpha,i}))
=
x_i W_{\alpha,i}(1-W_{\alpha,i}),
\end{equation}
\begin{equation}
\frac{\partial W_{\beta,i}}{\partial \beta}
=
-x_i\sigma(z_{\beta,i})(1-\sigma(z_{\beta,i}))
=
-x_i W_{\beta,i}(1-W_{\beta,i}).
\end{equation}
Since $x_i\in[0,1]$ and $u(1-u)\le 1/4$ for all $u\in[0,1]$, we obtain:
\begin{equation}
\left|
\frac{\partial W_{\alpha,i}}{\partial \alpha}
\right|
\le \frac{1}{4},
\quad
\left|
\frac{\partial W_{\beta,i}}{\partial \beta}
\right|
\le \frac{1}{4}.
\end{equation}
This bounded-gradient property is desirable for bilevel and alternating optimization~\cite{lin2023non}.

Since $\mathcal{W}_\delta$ corresponds to the moderate region, it can be seen as an interior localized transition band, not a half-space, so a monotone function cannot isolate it. Let $t_{\delta,i}=(\delta-1)x_i-m_i$.

Then $W_{\delta,i} = \exp\left(-\frac{t_{\delta,i}^2}{2}\right)$
is a smooth band-pass detector centered around $m_i=(\delta-1)x_i$.

The derivative with respect to $\delta$ is:
\begin{equation}
\frac{\partial W_{\delta,i}}{\partial \delta}
=
-x_i t_{\delta,i}
\exp\left(-\frac{t_{\delta,i}^2}{2}\right)
=
-x_i t_{\delta,i} W_{\delta,i}.
\end{equation}
Again, the gradient is bounded:
\begin{equation}
\left|
\frac{\partial W_{\delta,i}}{\partial \delta}
\right|
\le
\max_{u\in\mathbb{R}} |u|e^{-u^2/2}
=
e^{-1/2}.
\end{equation}
Thus, the radial basis function provides a stable and adaptive localized update for the moderate region.

\subsection{Bounds of the weight functions and their derivatives}
\label{supp:derivative}

Note that our weight functions are defined as in Section~\ref{sec:method}:
\begin{align}\tag{2}
    \mathcal{W}_\alpha (s_i, \widetilde{y_i}) = \sigma(\alpha \cdot s_i[\widetilde{y_i}] - \max(s_i))
\end{align}
\begin{align}\tag{3}
    \mathcal{W}_\beta (s_i, \widetilde{y_i}) = \sigma(-(\beta \cdot s_i[\widetilde{y_i}] - \max(s_i)))
\end{align}
\begin{align}\tag{4}
    \mathcal{W}_\delta (s_i, \widetilde{y_i}) = \exp\biggl(-\frac{(\delta \cdot s_i[\widetilde{y_i}] - \max(s_i))^2}{2}\biggr)
\end{align}
\noindent We calculated the weight for each sample as follows:
\begin{align}\tag{5}
    \mathcal{W}(s_i, \widetilde{y_i}) = \mathcal{W}_\alpha (s_i, \widetilde{y_i}) + \mathcal{W}_\beta (s_i, \widetilde{y_i}) + \mathcal{W}_\delta (s_i, \widetilde{y_i})
\end{align}
\noindent and defined our LiLAW weighted loss function as follows:
\begin{align}\tag{6}
\mathcal{L}_{W} &= \mathcal{W}(s_i, \widetilde{y_i}) \cdot \mathcal{L}
\end{align}

Let us consider $\mathcal{W}_\alpha = \sigma(\alpha \cdot s_i[\widetilde{y_i}] - \max(s_i))$. $\mathcal{W}_\alpha < 0.5$ when:
\begin{align}
    \alpha \cdot s_i[\widetilde{y_i}] - \max(s_i) &< 0
    %\alpha \cdot s_i[\widetilde{y_i}] &< \max(s_i) \\
     \implies\alpha \cdot s_i[\widetilde{y_i}] < \max(s_i) \leq 1 \text{ by (ii)}
\end{align}
$\mathcal{W}_\alpha \geq 0.5$ when:
\begin{align}
    \alpha \cdot s_i[\widetilde{y_i}] - \max(s_i) &\geq 0
    %\alpha \cdot s_i[\widetilde{y_i}] &\geq \max(s_i) \\
    \implies \alpha \cdot s_i[\widetilde{y_i}] \geq \max(s_i) \geq s_i[\widetilde{y_i}] \text{ by (i)}
\end{align}
As such, the upper and lower bounds on $\mathcal{W}_\alpha$, as seen in Figure~\ref{fig:graph}, are as follows:
\begin{align}
    -1 \leq \alpha \cdot s_i[\widetilde{y_i}] - \max(s_i) &\leq \alpha-1 
    \implies 0.2689 \approx \sigma(-1) \leq \mathcal{W}_\alpha \leq \sigma(\alpha-1) < 1
\end{align}
Now, let's consider $\mathcal{W}_\beta = \sigma(-(\beta \cdot s_i[\widetilde{y_i}] - \max(s_i)))$. $\mathcal{W}_\beta < 0.5$ when:
\begin{align}
    \beta \cdot s_i[\widetilde{y_i}] - \max(s_i) &> 0 
    \implies \beta \cdot s_i[\widetilde{y_i}] > \max(s_i) \geq s_i[\widetilde{y_i}] \text{ by (i)}
    %\beta \cdot s_i[\widetilde{y_i}] &> \max(s_i) \\
\end{align}
$\mathcal{W}_\beta \geq 0.5$ when:
\begin{align}
    \beta \cdot s_i[\widetilde{y_i}] - \max(s_i) &\leq 0
    %\beta \cdot s_i[\widetilde{y_i}] &\leq \max(s_i) \\
    \implies \beta \cdot s_i[\widetilde{y_i}] \leq \max(s_i) \leq 1 \text{ by (ii)}
\end{align}
The upper and lower bounds on $\mathcal{W}_\beta$, as seen in Figure~\ref{fig:graph}, are as follows:
\begin{align}
    1-\beta \leq -(\beta \cdot s_i[\widetilde{y_i}] - \max(s_i)) &\leq 1 \implies 0 < \sigma(1-\beta) \leq \mathcal{W}_\beta \leq \sigma(1) \approx 0.7311
\end{align}
Finally, consider $\mathcal{W}_\delta (s_i, \widetilde{y_i}) = \exp\bigl(-\frac{(\delta \cdot s_i[\widetilde{y_i}] - \max(s_i))^2}{2}\bigr)$. $\mathcal{W}_\delta = 1$ when:
\begin{align}
    \delta \cdot s_i[\widetilde{y_i}] - \max(s_i) = 0 \implies
    \delta \cdot s_i[\widetilde{y_i}] = \max(s_i)
\end{align}
Otherwise, $0 < \mathcal{W}_\delta < 1$, with $\mathcal{W}_\delta$ tending towards zero as $\delta \cdot s_i[\widetilde{y_i}]$ and $\max(s_i)$ grow apart.

We now consider the derivatives of our weight functions based on $\alpha, \beta, \delta$ with respect to the LiLAW weighted loss function to study how they grow with those three parameters. Based on the definitions of the weight functions, we have the following derivatives:
\begin{align}
    \nabla_{\alpha}\mathcal{L}_{W} &= \diffp{\mathcal{L}_{W}}{\alpha} = \diffp{}{\alpha} (\mathcal{W}_\alpha (s_i, \widetilde{y_i}) \cdot \mathcal{L}) = \mathcal{L} \cdot \diffp{}{\alpha} \mathcal{W}_\alpha (s_i, \widetilde{y_i}) \\ &= \mathcal{L} \cdot \diffp{}{\alpha} (\sigma(\alpha \cdot s_i[\widetilde{y_i}] - \max(s_i))) \\ &= \frac{\mathcal{L} \cdot (\sigma(\alpha \cdot s_i[\widetilde{y_i}] - \max(s_i)))^2 \cdot s_i[\widetilde{y_i}]}{\exp\biggl(\alpha \cdot s_i[\widetilde{y_i}] - \max(s_i)\biggl)} \\ &= \frac{\mathcal{L} \cdot \mathcal{W}_\alpha (s_i, \widetilde{y_i})^2 \cdot s_i[\widetilde{y_i}]}{\exp\biggl(\alpha \cdot s_i[\widetilde{y_i}] - \max(s_i)\biggl)}
\end{align}

\noindent Note: $\nabla_{\alpha}\mathcal{L}_{W} \geq 0$ as $\mathcal{L} \geq 0$, $\mathcal{W}_\alpha (s_i, \widetilde{y_i})^2 > 0$, $s_i[\widetilde{y_i}] \geq 0$, and $\exp\biggl(\alpha \cdot s_i[\widetilde{y_i}] - \max(s_i)\biggl) > 0$.

\begin{align}
    \nabla_{\beta}\mathcal{L}_{W} &= \diffp{\mathcal{L}_{W}}{\beta} = \diffp{}{\beta} (\mathcal{W}_\beta (s_i, \widetilde{y_i}) \cdot \mathcal{L})= \mathcal{L} \cdot \diffp{}{\beta} \mathcal{W}_\beta (s_i, \widetilde{y_i}) \\ &= \mathcal{L} \cdot \diffp{}{\beta} (\sigma(-(\beta \cdot s_i[\widetilde{y_i}] - \max(s_i)))) \\ &= -\frac{\mathcal{L} \cdot (\sigma(-(\beta \cdot s_i[\widetilde{y_i}] - \max(s_i))))^2 \cdot s_i[\widetilde{y_i}]}{\exp\biggl(-(\beta \cdot s_i[\widetilde{y_i}] - \max(s_i))\biggl)} \\ &= -\frac{\mathcal{L} \cdot \mathcal{W}_\beta (s_i, \widetilde{y_i})^2 \cdot s_i[\widetilde{y_i}]}{\exp\biggl(-(\beta \cdot s_i[\widetilde{y_i}] - \max(s_i))\biggl)}
\end{align}

\noindent Note: $\nabla_{\beta}\mathcal{L}_{W} \leq 0$ as $\mathcal{L} \geq 0$, $\mathcal{W}_\beta (s_i, \widetilde{y_i})^2 > 0$, $s_i[\widetilde{y_i}] \geq 0$, and $\exp\biggl(-(\beta \cdot s_i[\widetilde{y_i}] - \max(s_i))\biggl) > 0$.

\begin{align}\nabla_{\delta}\mathcal{L}_{W} &= \diffp{\mathcal{L}_{W}}{\delta} = \diffp{}{\delta} (\mathcal{W}_\delta (s_i, \widetilde{y_i}) \cdot \mathcal{L}) = \mathcal{L} \cdot \diffp{}{\delta} \mathcal{W}_\delta (s_i, \widetilde{y_i}) \\ &= \mathcal{L} \cdot \diffp{}{\delta} \biggl(\exp\biggl(-\frac{(\delta \cdot s_i[\widetilde{y_i}] - \max(s_i))^2}{2}\biggr)\biggr) \\ &= -\frac{\mathcal{L} \cdot (\delta \cdot s_i[\widetilde{y_i}] - \max(s_i)) \cdot s_i[\widetilde{y_i}]}{\exp\biggl(\frac{(\delta \cdot s_i[\widetilde{y_i}] - \max(s_i))^2}{2}\biggr)} \\ &= -\mathcal{L} \cdot \mathcal{W}_\delta (s_i, \widetilde{y_i}) \cdot (\delta \cdot s_i[\widetilde{y_i}] - \max(s_i)) \cdot s_i[\widetilde{y_i}]
\end{align}

\noindent Note: $\mathcal{L} \geq 0$, $\mathcal{W}_\delta (s_i, \widetilde{y_i}) > 0$, and $s_i[\widetilde{y_i}] \geq 0$, we see that $\nabla_{\delta}\mathcal{L}_{W} \leq 0$ when $\delta \cdot s_i[\widetilde{y_i}] \geq \max(s_i)$ and $\nabla_{\delta}\mathcal{L}_{W} > 0$ when $\delta \cdot s_i[\widetilde{y_i}] < \max(s_i)$.

\begin{figure}[ht!]
    \centering
    \includegraphics[scale=0.23]{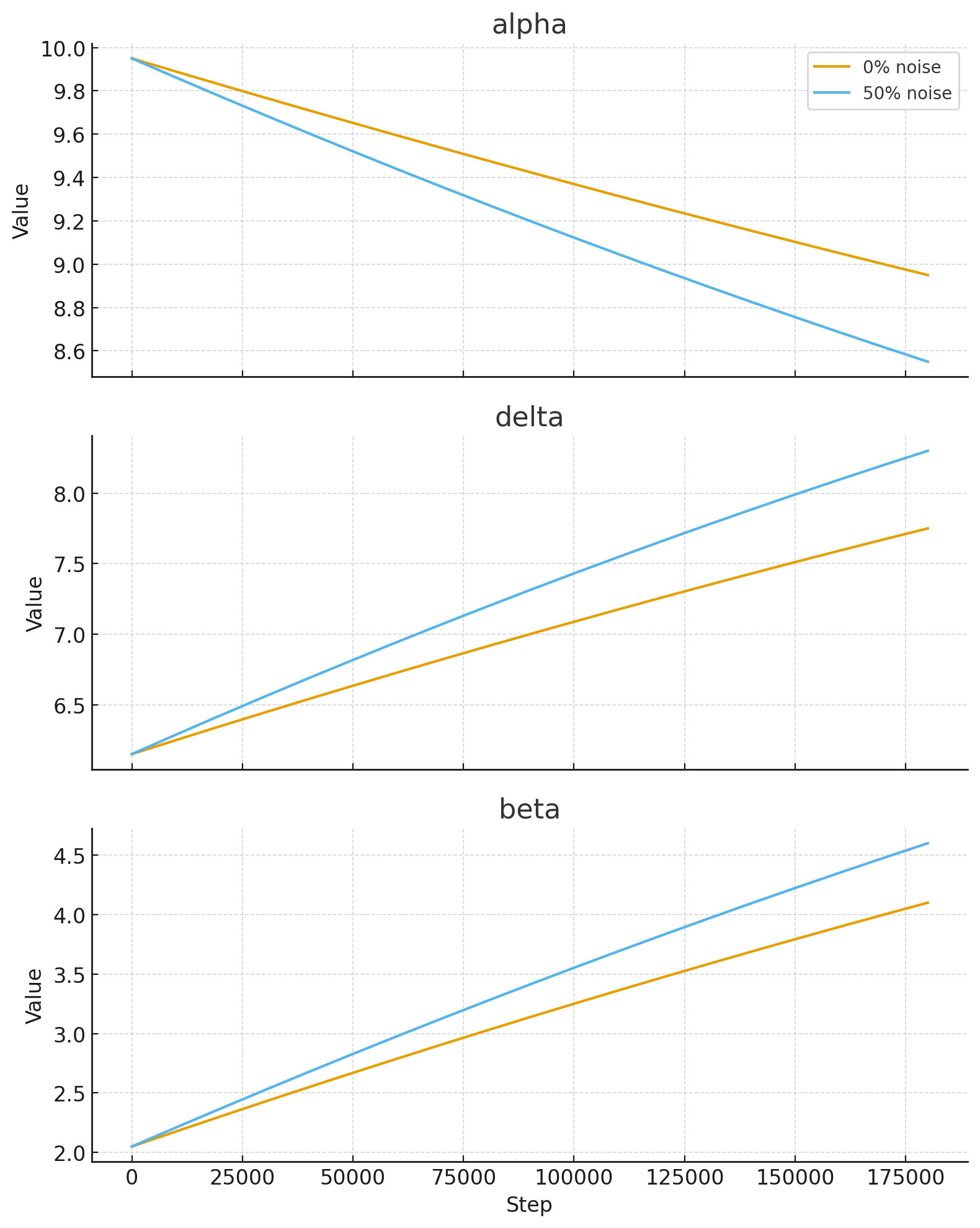}
    \caption{Plots showing how $\alpha,\beta,\delta$ change during training with 0\% symmetric noise and 50\% symmetric noise on CIFAR-100-M.}
    \label{fig:abd}
\end{figure}

In summary, the gradients for the parameters are updated using autograd, but we show why $\alpha$ always decreases and $\beta$ always increases at different rates. On the other hand, $\delta$ could increase or decrease depending on the conditions above. This is the reason for our choice of high $\alpha$, low $\beta$, and $\delta$ in between (by default, $\alpha=10,\beta=2,\delta=6$). Figure~\ref{fig:abd} shows that $\alpha$ decreases, $\beta$ increases, and $\delta$ increases as discussed above. The reason $\alpha$ decreases faster, $\delta$ increases faster, and $\beta$ increases faster with 50\% noise than 0\% noise is because easy samples are less reliable sooner, moderate cases are more informative faster, and hard samples are also more informative faster, respectively.

\newpage
\subsection{Time complexity analysis}
\label{supp:timecomplexity}

Without LiLAW, we have the following time-complexity for Algorithm~\ref{algo}. Let $B_t$ be the training batch size and let $T_\theta(B_t)$ denote the cost of the forward pass, loss calculation, backward pass, and update step for a mini-batch of size $B_t$. As this cost scales linearly with the number of model parameters and the batch size, we have $T_\theta(B_t)=\mathcal{O}(|\theta| \cdot B_t)$. Since there are $\lceil |\mathcal{D}_t|/B_t\rceil$ training mini-batches per epoch, the per-epoch runtime without LiLAW is:
\begin{equation} 
\mathcal{O}\left(\frac{|\mathcal{D}_t|}{B_t}  \cdot T_\theta(B_t)\right)
=
\mathcal{O}\left(\frac{|\mathcal{D}_t|}{B_t}  \cdot |\theta| \cdot B_t\right)
=
\mathcal{O}(|\theta| \cdot |\mathcal{D}_t|).
\end{equation}
Thus, the total runtime for $n$ epochs without LiLAW is $\mathcal{O}(n \cdot |\theta| \cdot |\mathcal{D}_t|)$.

With LiLAW, each training mini-batch is paired with exactly one sampled validation mini-batch. Let $B_v$ be the validation batch size. The per-step runtime is therefore: $\mathcal{O}\left(T_\theta(B_t) + T_\theta(B_v)\right)$,
up to constant-time operations involving the three LiLAW parameters $\alpha,\beta,\delta$. Since the number of steps per epoch is still determined by the number of training mini-batches, the per-epoch runtime is:
\begin{equation} 
\mathcal{O}\left(\frac{|\mathcal{D}_t|}{B_t} \cdot \left(T_\theta(B_t)+T_\theta(B_v)\right)\right).
\end{equation}
Assuming equal training and validation batch sizes, $B_t=B_v=B$, and using $T_\theta(B)=\mathcal{O}(|\theta| \cdot B)$:
\begin{equation} 
\mathcal{O}\left(\frac{|\mathcal{D}_t|}{B}\cdot 2 \cdot |\theta| \cdot B\right)
=
\mathcal{O}(|\theta| \cdot |\mathcal{D}_t|).
\end{equation}
Therefore, the total runtime for $n$ epochs with LiLAW is also $\mathcal{O}(n \cdot |\theta| \cdot |\mathcal{D}_t|)$. Importantly, the per-epoch runtime is independent of the total validation-set size $|\mathcal{D}_v|$, since $|\mathcal{D}_v|$ only determines the pool from which the validation mini-batch is sampled.

We confirm empirically that across several runs of ViT-Base-16-224 with ImageNet-1K pretraining, on 224$\times$224 inputs with CIFAR-100-M, we observe only a small increase in per-epoch training time of approximately $18$ seconds, from roughly $3{:}01$ minutes without LiLAW to roughly $3{:}19$ minutes with LiLAW, on one NVIDIA L40S GPU. This overhead comes from processing one additional validation mini-batch after each training mini-batch and does not change the asymptotic per-epoch time complexity of standard training.

\subsection{Space complexity analysis}
\label{supp:spacecomplexity}

Without LiLAW, the space complexity for Algorithm~\ref{algo} consists of memory for the model parameters, their gradients, optimizer states, and intermediate activations. The parameter and gradient memory are each $\mathcal{O}(|\theta|)$, and the optimizer state is also $\mathcal{O}(|\theta|)$ for standard optimizers up to constant factors. Let $A_\theta(B_t)$ denote the activation memory required by the base model for a training mini-batch of size $B_t$. This term depends on the architecture, parameters, input resolution, hidden dimensions, etc. Thus, the total space complexity without LiLAW is $\mathcal{O}(|\theta| + A_\theta(B_t))$.

With LiLAW, the three additional parameters $\alpha,\beta,\delta$ and their gradients add only $\mathcal{O}(1)$ memory. The model parameter, gradient, and optimizer-state memory therefore remains $\mathcal{O}(|\theta|)$. In addition, LiLAW processes one validation mini-batch for each training mini-batch. Let $A_\theta(B_v)$ denote the activation memory for a validation mini-batch of size $B_v$. So, the total activation memory is:
\begin{equation} 
\mathcal{O}(\max\{A_\theta(B_t), A_\theta(B_v)\}).
\end{equation}

In either case, if $B_t=B_v$, the space complexity with LiLAW is also $\mathcal{O}(|\theta| + A_\theta(B_t))$.

Across several runs of ViT-Base-16-224 with ImageNet-1K pretraining, on 224$\times$224 inputs with CIFAR-100-M, we observe a negligible memory increase of about $30$ MB, corresponding to roughly $1.34\%$, from about $2161$ MB without LiLAW to about $2190$ MB with LiLAW, on one NVIDIA L40S GPU. This overhead comes from the additional validation mini-batch and related intermediate quantities, but it does not change the asymptotic space complexity relative to standard training.

\newpage

\subsection{Extension of LiLAW to multi-label classification}
\label{sec:multilabel}
In the $c$-class multi-label classification case, we extend LiLAW as follows.

Let $\mathcal{D}_{t} = \{(x_i, \widetilde{y_i})\}^N_{i=1}$ represent the training set and $\mathcal{D}_{v} = \{(x_j, \widetilde{y_j})\}^{N+M}_{j=N+1}$ represent the validation set. As before, note that $(x_i, \widetilde{y_i})$ represents the pairs of inputs and observed (potentially synthetic/noisy) targets. For multi-label classification, $x_i \in \mathcal{X}$, where $\mathcal{X}$ is the input space and $\widetilde{y_i} \in \mathcal{Y} = \{0,1\}^c,$ is the output space with $c \in \mathbb{N}$ such that $c \ge 1$ is the total number of labels. Note that $\widetilde{y_i}[z]$ from $z \in \{0,...,c-1\}$ is 0 or 1 depending on whether the input has the observed label. Let $f_{\theta}: \mathcal{X} \rightarrow \mathbb{R}^c$ be the neural network model, $\theta$ be its parameters, and $\sigma_i = \sigma(f_{\theta}(x_i))$ be the sigmoid of its logits. We keep a similar setup as before, except in $\mathcal{W}_\alpha$, $\mathcal{W}_\beta$, $\mathcal{W}_\delta$, we replace $s_i[\widetilde{y_i}]$ with $\frac{1}{k}\sum_{j=0}^{c-1}\widetilde{y}_i[j]\sigma_i[j]$ and $\max(s_i)$ with $\frac{1}{k}\sum_{j\in \mathrm{Top}\text{-}k(\sigma_i)}\sigma_i[j]$, where $k=\sum_{j=0}^{c-1}\widetilde{y}_i[j]$.

Instead of using the score of a single observed class as in the multi-class classification case, we summarize the model's confidence in the observed positive label set by averaging the predicted probabilities assigned to all $k$ positive observed labels. We similarly replace the single maximum probability with the average of the top $k$ predicted probabilities, since multiple labels may be relevant for each sample. The LiLAW weighting functions then use the same confidence / disagreement comparison as before, but with these two averaged quantities. This allows LiLAW to assign one scalar weight per multi-label sample while accounting for multiple observed positive labels. We assume $k\ge 1$, so samples with no observed positive labels require a different approach.

\subsection{Extension of LiLAW to regression}
\label{sec:regression}
In the regression case, we extend LiLAW as follows:

Let $\mathcal{D}_{t} = \{(x_i, \widetilde{y_i})\}^N_{i=1}$ represent the training set and $\mathcal{D}_{v} = \{(x_j, \widetilde{y_j})\}^{N+M}_{j=N+1}$ represent the validation set. As before, note that $(x_i, \widetilde{y_i})$ represents the pairs of inputs and observed (potentially synthetic/noisy) targets. However, for regression, $x_i \in \mathcal{X}$, where $\mathcal{X}$ is the input space and $\widetilde{y_i} \in \mathcal{Y} = \mathbb{R},$ is the output space. Let $f_{\theta}: \mathcal{X} \rightarrow \mathcal{Y}$ be the neural network model, $\theta$ be its parameters, and $f_{\theta}(x_i)$ be its output. Also, let $R$ be the range of the true targets, which is usually well-established in any given regression problem. We keep the setup similar to before, except in $\mathcal{W}_\alpha$, $\mathcal{W}_\beta$, $\mathcal{W}_\delta$, we replace $\alpha \cdot s_i[\widetilde{y_i}] - \max(s_i)$ with $\frac{1}{R}\cdot (\alpha \cdot f_{\theta}(x_i) - \widetilde{y_i})$, replace $\beta \cdot s_i[\widetilde{y_i}] - \max(s_i)$ with $\frac{1}{R}\cdot (\beta \cdot f_{\theta}(x_i) - \widetilde{y_i})$, and replace $\delta \cdot s_i[\widetilde{y_i}] - \max(s_i)$ with $\frac{1}{R}\cdot (\delta \cdot f_{\theta}(x_i) - \widetilde{y_i})$.

Unlike a symmetric residual-based formulation, this regression extension is not intended to depend only on $|f_{\theta}(x_i)-\widetilde{y_i}|$. Instead, we have signed, target-scaled margins comparing the parameter-scaled prediction to the observed target. Therefore, over-predictions and under-predictions of the same absolute magnitude may receive different weights. This asymmetry is useful because the target magnitude itself carries meaning and different directions of error may deserve different training emphasis.

\subsection{Comparison of the three versions of LiLAW}
\label{comparison}
We compare the three weight functions for the multi-class classification, multi-label classification, and regression cases. Note that the corresponding baseline loss function also changes accordingly.

\subsubsection{Multi-class classification}
\begin{align}
    \mathcal{W}_\alpha (s_i, \widetilde{y_i}) = \sigma(\alpha \cdot s_i[\widetilde{y_i}] - \max(s_i))\\
    \mathcal{W}_\beta (s_i, \widetilde{y_i}) = \sigma(-(\beta \cdot s_i[\widetilde{y_i}] - \max(s_i)))\\
    \mathcal{W}_\delta (s_i, \widetilde{y_i}) = \exp\biggl(-\frac{(\delta \cdot s_i[\widetilde{y_i}] - \max(s_i))^2}{2}\biggr)
\end{align}

\subsubsection{Multi-label classification}
\begin{align}
\mathcal{W}_\alpha(\sigma_i,\widetilde{y}_i)
&=
\sigma\left(
\alpha \cdot
\frac{1}{k}\sum_{j=0}^{c-1}\widetilde{y}_i[j]\sigma_i[j]
-
\frac{1}{k}\sum_{j\in \mathrm{Top}\text{-}k(\sigma_i)}\sigma_i[j]
\right)\\
\mathcal{W}_\beta(\sigma_i,\widetilde{y}_i)
&=
\sigma\left(
-\left(
\beta \cdot
\frac{1}{k}\sum_{j=0}^{c-1}\widetilde{y}_i[j]\sigma_i[j]
-
\frac{1}{k}\sum_{j\in \mathrm{Top}\text{-}k(\sigma_i)}\sigma_i[j]
\right)
\right)\\
\mathcal{W}_\delta(\sigma_i,\widetilde{y}_i)
&=
\exp\left(
-\frac{
\left(
\delta \cdot
\frac{1}{k}\sum_{j=0}^{c-1}\widetilde{y}_i[j]\sigma_i[j]
-
\frac{1}{k}\sum_{j\in \mathrm{Top}\text{-}k(\sigma_i)}\sigma_i[j]
\right)^2
}{2}
\right)
\end{align}

\subsubsection{Regression}
\begin{align}
    \mathcal{W}_\alpha (s_i, \widetilde{y_i}) = \sigma\biggl(\frac{1}{R}\cdot (\alpha \cdot f_{\theta}(x_i) - \widetilde{y_i})\biggr)\\
    \mathcal{W}_\beta (s_i, \widetilde{y_i}) = \sigma\biggl(-\biggl(\frac{1}{R}\cdot (\beta \cdot f_{\theta}(x_i) - \widetilde{y_i})\biggr)\biggr)\\
    \mathcal{W}_\delta (s_i, \widetilde{y_i}) = \exp\biggl(-\frac{(\frac{1}{R}\cdot (\delta \cdot f_{\theta}(x_i) - \widetilde{y_i}))^2}{2}\biggr)
\end{align}

\subsection{Implementation Details for General Imaging and Medical Imaging Dataset}
\label{imp}

\begin{enumerate}[leftmargin=*,labelindent=0pt,label=\textbf{--}]
    \itemsep-0.1em
    \item ViT-Base-16-224~\citep{vit}, with ImageNet-1K pretraining, on 224$\times$224 inputs with CIFAR-100-M, CIFAR-10-M, FashionMNIST-M, and MNIST-M. We trained for 10 epochs with batch size 16, learning rate $0.0002$, and weight decay $0$, using a linear learning rate scheduler. 
    \item ResNet-18~\citep{resnet}, with ImageNet-1K pretraining, on 224$\times$224 inputs with the MedMNISTv2 datasets. We trained for 100 epochs with batch size 128, learning rate $0.0001$, and weight decay $0$, using a multi-step learning rate scheduler with a $0.1$ decay at 50 epochs and 75 epochs, as mentioned in~\citet{medmnistv2}. Note that ResNet-18 was used for the MedMNISTv2 datasets because it achieves the highest average accuracy of all the benchmark models in~\citet{medmnistv2}.
\end{enumerate}

All of the above models use the Adam optimizer~\citep{kingma2014adam} with early stopping. We use a warmup period of 1 epoch to ensure that the model briefly learns from the data before using LiLAW. The parameters are initialized to $\alpha_{init}, \beta_{init}, \delta_{init} = 10, 2, 6$, with learning rates $\alpha_{lr}, \beta_{lr}, \delta_{lr} = 0.005$, and weight decays $\alpha_{wd}, \beta_{wd}, \delta_{wd} = 0.0001$ for manual gradient descent. Our method is not highly sensitive to these choices as shown in~\ref{supp:abd_init}. We use cross-entropy loss, but also evaluate with focal loss in~\ref{supp:loss}.

%Additionally, the CIFAR-100-M, CIFAR-10-M, FashionMNIST-M, and MNIST-M experiments with ViT-Base-16-224, with ImageNet-1K pretraining, on 224x224 inputs were compared with: fine-tuning on the full pretrained model (without LiLAW) and fine-tuning solely on the last 2 layers of the pretrained model (with LiLAW).

\subsection{Performance with various noise levels}
\label{supp:noiselevel}

\begin{table}[ht!]
\centering
\caption{Results with different levels of symmetric noise on CIFAR-100-M. \label{table:noiselevels}}
\footnotesize{
\begin{tabular}{cccc}
\hline
\textbf{Training Noise (\%)} & \textbf{Top-1 Acc. (\%)} & \textbf{Top-5 Acc. (\%)} & \textbf{AUROC} \\
\hline
0  & 80.85 \textcolor{BrickRed}{$\downarrow$} 0.01 & 96.07 \textcolor{Green}{$\uparrow$} 0.16 & 0.9888 \textcolor{BrickRed}{$\downarrow$} 0.0005 \\
10 & 79.59 \textcolor{Green}{$\uparrow$} 0.20 & 95.32 \textcolor{Green}{$\uparrow$} 0.14 & 0.9860 \textcolor{BrickRed}{$\downarrow$} 0.0002 \\
20 & 78.72 \textcolor{Green}{$\uparrow$} 0.66 & 94.80 \textcolor{Green}{$\uparrow$} 0.23 & 0.9870 \textcolor{Green}{$\uparrow$} 0.0001 \\
30 & 77.86 \textcolor{Green}{$\uparrow$} 0.74 & 94.29 \textcolor{Green}{$\uparrow$} 0.30 & 0.9833 \textcolor{Green}{$\uparrow$} 0.0006 \\
40 & 76.77 \textcolor{Green}{$\uparrow$} 0.88 & 93.55 \textcolor{Green}{$\uparrow$} 0.49 & 0.9835 \textcolor{Green}{$\uparrow$} 0.0010 \\
50 & 75.20 \textcolor{Green}{$\uparrow$} 1.35 & 92.61 \textcolor{Green}{$\uparrow$} 0.67 & 0.9811 \textcolor{Green}{$\uparrow$} 0.0013 \\
60 & 73.62 \textcolor{Green}{$\uparrow$} 1.80 & 91.22 \textcolor{Green}{$\uparrow$} 1.10 & 0.9771 \textcolor{Green}{$\uparrow$} 0.0022 \\
70 & 70.83 \textcolor{Green}{$\uparrow$} 2.63 & 88.98 \textcolor{Green}{$\uparrow$} 1.39 & 0.9713 \textcolor{Green}{$\uparrow$} 0.0033 \\
80 & 66.64 \textcolor{Green}{$\uparrow$} 3.20 & 85.37 \textcolor{Green}{$\uparrow$} 1.93 & 0.9591 \textcolor{Green}{$\uparrow$} 0.0052 \\
90 & 53.03 \textcolor{Green}{$\uparrow$} 2.66 & 73.54 \textcolor{Green}{$\uparrow$} 1.74 & 0.9220 \textcolor{Green}{$\uparrow$} 0.0036 \\
\hline
\end{tabular}}
\end{table}

In Table~\ref{table:noiselevels}, we report test performance across noise levels from 0\% to 90\%. In nearly every setting, LiLAW improves both accuracy and AUROC over the baseline with larger gains at higher noise levels. In particular, linear probing with LiLAW, top-1 accuracy receives accuracy similar to linear probing without LiLAW at a 10\% reduced noise level. Under noisy settings, how a pretrained model is used while fine-tuning has a large effect on the performance. As mentioned in~\citet{marrium2024implicit}, full fine-tuning adapts all parameters of a pre-trained model to a downstream task with high time and space complexity while also severely overfitting to noise. On the other hand, linear probing only tunes a few parameters to the downstream task and the risk of overfitting is drastically reduced. In addition to linear probing, we use LiLAW to further improve results.

\subsection{Cost-Performance Trade-off on Clothing-1M}
\label{supp:clothing1m_tradeoff}

As shown in Table~\ref{tab:clothing1m_complexity}, LiLAW has a strong overall accuracy-runtime-space trade-off. On Clothing-1M, LiLAW improves over cross-entropy from $69.21$ to $71.44$ ($+2.23$), while having asymptotic runtime and space complexity no larger than standard cross-entropy training and no larger than the other directly comparable robust-learning or meta-reweighting baselines under the same model and training setup. Some methods with slightly higher accuracy use additional data, auxiliary models, instance-wise weighting networks, sample pruning, synthetic representations, or class-dependent correction terms, and are therefore not directly comparable in implementation cost.

Here, $E$ is the number of epochs, $N$ is the dataset size, $P$ is the time for a forward/backward pass, $C$ is the number of classes, $M$ is the number of model parameters/activations, $B$ is the batch size, $K$ is the number of nearest neighbors, $s$ is the MetaCleaner subset size, $q$ is the number of dataset correction passes in LRT, $d$ is the feature dimension, and $r$ is the number of self-evolution rounds.

Note that methods are defined as ``comparable to LiLAW'' if the training data is not significantly changed or augmented, although most methods with similar training setups still have a larger total runtime and total space complexity compared to LiLAW.

\begin{table}[ht!]
\centering
\caption{
Complexity comparison for 16 baseline noisy learning methods on Clothing-1M. $\Delta$ denotes the relative accuracy change with respect to CE. These are normalized big-$\mathcal{O}$ derivations in shared variables. The top 3 results are in \textbf{bold}, but note that the top 2 results have a \textit{significantly} higher total runtime and space complexity. LiLAW has the best results of methods with the same runtime and space complexity. Results are from the respective papers.
}
\setlength{\tabcolsep}{3pt}
\renewcommand{\arraystretch}{1.12}
\resizebox{\textwidth}{!}{
\begin{tabular}{@{}lcclll@{}}
\toprule
Method & Acc. (\%) & $\Delta$ w.r.t. CE & Total runtime & Total space & Comparable to LiLAW? \\
\midrule
Cross-entropy (CE)
& $69.21$ & -- & $\mathcal{O}(ENP)$ & $\mathcal{O}(M)$ & Yes \\

APNL~\citep{ma2020normalized}
& $69.35$ & $+0.14$ & $\mathcal{O}(ENP)$ & $\mathcal{O}(M)$ & Yes \\

GCE~\citep{zhang2018generalized}
& $69.75$ & $+0.54$ & $\mathcal{O}(ENP)$ & $\mathcal{O}(M)$ & Yes \\

Mixup~\citep{zhang_mixup_2018}
& $70.48$ & $+1.27$ & $\mathcal{O}(ENP)$ & $\mathcal{O}(M)$ & No (dataset augmented) \\

SCE~\citep{wang2019symmetric}
& $71.02$ & $+1.81$ & $\mathcal{O}(ENP)$ & $\mathcal{O}(M)$ & Yes \\

Weakly Supervised~\citep{zhuang2017attend}
& $71.36$ & $+2.15$ & $\mathcal{O}(ENP)$ & $\mathcal{O}(M)$ & No (randomly stacks samples) \\

\midrule

Backward~\citep{patrini2016making}
& $69.13$ & $-0.08$ & $\mathcal{O}(ENP + ENC^2 + C^3)$ & $\mathcal{O}(M + C^2)$ & Yes, but larger runtime and space \\

Forward~\citep{patrini2016making}
& $69.84$ & $+0.63$ & $\mathcal{O}(ENP + ENC^2)$ & $\mathcal{O}(M + C^2)$ & Yes, but larger runtime and space \\

Co-teaching~\citep{han2018co}
& $70.15$ & $+0.94$ & $\mathcal{O}(ENP + EN\log B)$ & $\mathcal{O}(M+B)$ & Yes, but larger runtime and space \\

Dyn-Mixup~\citep{arazo2019unsupervised}
& $70.33$ & $+1.12$ & $\mathcal{O}(ENP + EN)$ & $\mathcal{O}(M+N)$ & No (dataset augmented) \\

EG~\citep{majidi2021exponentiated}
& $70.52$ & $+1.31$ & $\mathcal{O}(ENP)$ & $\mathcal{O}(M+N)$ & Yes, but larger space \\

CIW~\citep{kumar2021constrained}
& $70.52$ & $+1.31$ & $\mathcal{O}(ENP)$ & $\mathcal{O}(M+B)$ & Yes, but larger space \\

SEAL~\citep{chen2021beyond}
& $70.63$ & $+1.42$ & $\mathcal{O}(rENP)$ & $\mathcal{O}(M+NC)$ & Yes, but larger runtime and space \\

CICW~\citep{kumar2021constrained}
& $70.84$ & $+1.63$ & $\mathcal{O}(ENP + ENC)$ & $\mathcal{O}(M+B+C)$ & Yes, but larger runtime and space \\

LRT~\citep{zheng2020error}
& \textbf{71.74} & $+2.53$ & $\mathcal{O}(ENP + qNC)$ & $\mathcal{O}(M+NC)$ & Yes, but larger runtime and space \\

Joint-Optim~\citep{tanaka2018joint}
& \textbf{72.16} & $+2.95$ & $\mathcal{O}(ENP + ENC)$ & $\mathcal{O}(M+NC)$ & Yes, but larger runtime and space \\

\midrule

\textbf{LiLAW (ours)}
& \textbf{71.44} & \textbf{$+2.23$} & $\mathcal{O}(ENP)$ & $\mathcal{O}(M)$ & - \\

%MixNN
%& $72.39$ & $+3.18$ &
%$\mathcal{O}(ENP + EN\log N + ENK)$ &
%$\mathcal{O}(M+Nd+NK+NC)$ &
%No (synthetic samples) \\

%MetaCleaner
%& $72.50$ & $+3.29$ &
%$\mathcal{O}(ENP + ENd)$ &
%$\mathcal{O}(M+sd)$ &
%No (synthetic representations) \\
\bottomrule
\end{tabular}
}

\label{tab:clothing1m_complexity}
\end{table}

\subsection{Sensitivity to the initialization of parameters}
\label{supp:abd_init}

Tables~\ref{tab:abd_sensitivity} and~\ref{tab:abd_sensitivity_summary} show that LiLAW is not highly sensitive to the initialization of $\alpha,\beta,\delta$, even across large multiplicative changes in scale. Across initializations from $(5,1,2)$ to $(100,10,20)$, the variation in Top-1 accuracy remains within $\pm 0.18\%$ at $0\%$ noise and $\pm 0.13\%$ at $50\%$ noise, with negligible variation in Top-5 accuracy and AUROC as well. This indicates that performance is governed primarily by the learned weighting dynamics rather than initial parameter values. Although $\alpha,\beta,\delta$ define the geometry of the confidence-disagreement weighting function, the validation update step rapidly adapts them to a stable region of the parameter space, yielding consistent effective weight distributions across runs. This is supported by the bounded and smooth gradients of $\mathcal{W}_\alpha$, $\mathcal{W}_\beta$, and $\mathcal{W}_\delta$, which prevents severe fluctuations and ensures stable updates. So, LiLAW avoids the brittle hyperparameter sensitivity typical of many reweighting methods, and can be reliably used with simple initializations. After some further tuning, we use $\alpha=10$, $\beta=2$, and $\delta=6$ as the default initialization.

\begin{table}[ht!]
\centering
\caption{
Sensitivity of LiLAW to the initialization of $\alpha,\beta,\delta$ on CIFAR-100-M. Performance remains stable across a broad range of initializations.
}
\label{tab:abd_sensitivity}
\setlength{\tabcolsep}{1pt}
\renewcommand{\arraystretch}{1}
\footnotesize{\begin{tabular}{@{}ccccccc@{}}
\toprule
$\alpha$ & $\beta$ & $\delta$ & Noise (\%) & Top-1 Acc. (\%) & Top-5 Acc. (\%) & AUROC \\
\midrule
5   & 1  & 2  & 0  & 80.89 & 96.07 & 0.9886 \\
5   & 1  & 2  & 50 & 76.59 & 93.17 & 0.9821 \\
\midrule
10  & 1  & 2  & 0  & 80.72 & 96.15 & 0.9886 \\
10  & 1  & 2  & 50 & 76.72 & 93.29 & 0.9822 \\

\midrule
10  & 2  & 5  & 0  & 80.83 & 96.15 & 0.9881 \\
10  & 2  & 5  & 50 & 76.56 & 93.24 & 0.9824 \\

\midrule
20  & 2  & 5  & 0  & 80.79 & 96.17 & 0.9881 \\
20  & 2  & 5  & 50 & 76.49 & 93.35 & 0.9826 \\

\midrule
50  & 5  & 10 & 0  & 80.84 & 96.01 & 0.9875 \\
50  & 5  & 10 & 50 & 76.71 & 93.61 & 0.9833 \\

\midrule
50  & 2  & 10 & 0  & 80.35 & 96.05 & 0.9880 \\
50  & 2  & 10 & 50 & 76.76 & 93.60 & 0.9833 \\
\midrule
100 & 10 & 20 & 0  & 80.54 & 95.85 & 0.9871 \\
100 & 10 & 20 & 50 & 76.35 & 93.65 & 0.9835 \\
\midrule
100 & 5  & 10 & 0  & 80.51 & 95.91 & 0.9876 \\
100 & 5  & 10 & 50 & 76.61 & 93.78 & 0.9840 \\
\bottomrule
\end{tabular}}
\end{table}

\begin{table}[ht!]
\centering

\caption{
Mean and standard deviation across the initialization sweep in Table~\ref{tab:abd_sensitivity}.
}
\label{tab:abd_sensitivity_summary}
\setlength{\tabcolsep}{8pt}
\renewcommand{\arraystretch}{1.1}
\footnotesize{\begin{tabular}{@{}cccc@{}}
\toprule
Noise (\%) & Top-1 Acc. (\%) & Top-5 Acc. (\%) & AUROC \\
\midrule
0  & 80.68 $\pm$ 0.18 & 96.05 $\pm$ 0.12 & 0.9880 $\pm$ 0.0005 \\
50 & 76.60 $\pm$ 0.13 & 93.46 $\pm$ 0.22 & 0.9832 $\pm$ 0.0006 \\
\bottomrule
\end{tabular}}
\end{table}

\subsection{Performance with and without calibration}
\label{supp:calibration}

According to Table \ref{table:calibration}, when noise is present, applying LiLAW leads to performance gains, regardless of calibration. When calibration is combined with LiLAW, it works synergistically to enhance robustness to noise. The performance boost at 50\% noise indicates that LiLAW effectively mitigates the effects of label noise. We note that there is a boost in top-5 test accuracy when using LiLAW even when there is 0\% noise since we are pushing low confidence predictions to be more confident. There is a slight decrease in AUROC in nearly all cases where there is 0\% noise since the model may need to be slightly less confident on the thresholds to improve accuracy.

\begin{table*}[ht!] \centering \caption{Results with and without calibration with two levels of symmetric noise on CIFAR-100-M.\label{table:calibration}} \footnotesize{ \begin{tabular}{l|cccc} \hline \textbf{Calibration} & \textbf{Noise Level (\%)} & \textbf{Top-1 Acc. (\%)} & \textbf{Top-5 Acc. (\%)} & \textbf{AUROC} \\ \hline \multirow{2}{*}{Without calibration} 
& 0  & 80.85 \textcolor{BrickRed}{$\downarrow$} 0.01 & 96.07 \textcolor{Green}{$\uparrow$} 0.16 & 0.9888 \textcolor{BrickRed}{$\downarrow$} 0.0005 \\
& 50 & 75.20 \textcolor{Green}{$\uparrow$} 1.35 & 92.61 \textcolor{Green}{$\uparrow$} 0.67 & 0.9811 \textcolor{Green}{$\uparrow$} 0.0013 \\ \hline

\multirow{2}{*}{With calibration} 
& 0  & 80.03 \textcolor{BrickRed}{$\downarrow$} 0.01 & 95.45 \textcolor{Green}{$\uparrow$} 0.07 & 0.9967 \textcolor{BrickRed}{$\downarrow$} 0.0001 \\
& 50 & 75.20 \textcolor{Green}{$\uparrow$} 1.32 & 92.61 \textcolor{Green}{$\uparrow$} 0.67 & 0.9827 \textcolor{Green}{$\uparrow$} 0.0013 \\ \hline \end{tabular}}  \end{table*}

\subsection{Performance with and without a clean validation set}
\label{supp:cleanvalset}

In Table \ref{table:cleanval}, we show test performance with and without a clean validation set. We see that LiLAW enhances performance even without a clean validation set, demonstrating its robustness in practical scenarios where obtaining a clean validation set may be challenging. The improvements with a clean validation set are comparable to those without one, indicating that LiLAW does not heavily rely on validation set cleanliness and can easily adapt to noisy validation data. In addition, training only on the 50\% clean subset achieves lower top-1 accuracy than using LiLAW without a clean validation set, although it has higher top-5 accuracy and AUROC.

\begin{table*}[ht!]
\centering
\caption{Results with and without a clean validation set on CIFAR-100-M with 50\% symmetric noise. We also compare these results to when the model is trained only on the 50\% of the clean data. \label{table:cleanval}}
\footnotesize{
\begin{tabular}{l|cccc}
\hline
\textbf{Validation set cleanliness} & \textbf{Noise Level (\%)} & \textbf{Top-1 Acc. (\%)} & \textbf{Top-5 Acc. (\%)} & \textbf{AUROC} \\
\hline
\multirow{2}{*}{Without a clean validation set} 
& 0  & 80.85 \textcolor{BrickRed}{$\downarrow$} 0.01 & 96.07 \textcolor{Green}{$\uparrow$} 0.16 & 0.9888 \textcolor{BrickRed}{$\downarrow$} 0.0005 \\
& 50 & 75.20 \textcolor{Green}{$\uparrow$} 1.35 & 92.61 \textcolor{Green}{$\uparrow$} 0.67 & 0.9811 \textcolor{Green}{$\uparrow$} 0.0013 \\ \hline

\multirow{2}{*}{With a clean validation set} 
& 0  & 80.03 \textcolor{Green}{$\uparrow$} 0.01 & 95.67 \textcolor{BrickRed}{$\downarrow$} 0.07 & 0.9882 \textcolor{BrickRed}{$\downarrow$} 0.0001 \\
& 50 & 75.20 \textcolor{Green}{$\uparrow$} 1.29 & 92.61 \textcolor{Green}{$\uparrow$} 0.70 & 0.9811 \textcolor{Green}{$\uparrow$} 0.0013 \\ \hline

Trained only on 50\% clean data & - & 74.34 & 94.94 & 0.9919 \\
\hline
\end{tabular}}
\end{table*}

\subsection{Performance under different random seeds}
\label{supp:randseed}

Table~\ref{table:seeds} motivates our reporting choice for the general imaging datasets. Full fine-tuning without LiLAW is included only as a diagnostic baseline. At 50\% noise, full fine-tuning is substantially worse and less stable than linear probing, while LiLAW gives an additional gain on top of linear probing. 

This supports using linear probing as the main evaluation setting and reporting LiLAW as the improvement over that setting for the general imaging datasets. Note that we use full fine-tuning or training from scratch for the medical datasets since ImageNet-1K pretraining is not as representative of those datasets and therefore may not tend to overfit as drastically.

\begin{table}[ht!]
\centering
\caption{Performance (mean $\pm$ std over 5 runs with different random seeds) on CIFAR-100-M across different noise levels with (\color{Green}{\checkmark}\color{Black}) and without (\color{BrickRed}{\texttimes}\color{Black}) LiLAW.}
\label{table:seeds}
\footnotesize{\begin{tabular}{cccccc}
\hline
\textbf{Noise Level (\%)} & \textbf{LiLAW} & \textbf{Top-1 Acc. (\%)} & \textbf{Top-5 Acc. (\%)} & \textbf{AUROC} \\
\hline
\multirow{2}{*}{0}  & {\color{BrickRed}\texttimes} (full fine-tuning) & 74.93 $\pm$ 1.07 & 94.56 $\pm$ 0.51 & 0.9918 $\pm$ 0.0009 \\
& {\color{BrickRed}\texttimes} (linear probing) & 80.99 $\pm$ 0.15 & 95.98 $\pm$ 0.27 & 0.9879 $\pm$ 0.0003 \\
  & {\color{Green}\checkmark} (linear probing) & 81.00 $\pm$ 0.21 & 95.92 $\pm$ 0.15 & 0.9874 $\pm$ 0.0003 \\ \hline
\multirow{2}{*}{50} & {\color{BrickRed}\texttimes} (full fine-tuning) & 46.32 $\pm$ 7.82 & 76.12 $\pm$ 5.98 & 0.9537 $\pm$ 0.0142 \\
& {\color{BrickRed}\texttimes} (linear probing) & 74.71 $\pm$ 0.14 & 91.94 $\pm$ 0.34 & 0.9773 $\pm$ 0.0008 \\
 & {\color{Green}\checkmark} (linear probing) & 76.21 $\pm$ 0.28 & 92.53 $\pm$ 0.26 & 0.9776 $\pm$ 0.0008 \\
\hline
\end{tabular}}
\end{table}

\iffalse
\begin{figure}[ht!]
    \centering
    \includegraphics[scale=0.38]{top1vnoise.png}
    \includegraphics[scale=0.38]{top5vnoise.png}
    \includegraphics[scale=0.38]{aurocvnoise.png}
    \caption{Plots with means and standard deviations of top-1 accuracy, top-5 accuracy, and AUROC across 5 runs with different random seeds on CIFAR-100-M under different noise levels.}
    \label{fig:seeds}
\end{figure}
\fi

\subsection{Performance under different loss functions}
\label{supp:loss}

In Table \ref{table:losses}, we see that LiLAW provides performance gains with both cross-entropy and focal loss functions, indicating its versatility even when we use different loss landscapes that are already designed to handle issues such as class imbalance. Although noisy labels can negatively affect training with either of these two losses, LiLAW's adaptive weighting helps mitigate the impact of mislabeled or noisy data by dynamically adjusting the loss weight of each sample. Note that the boosts with LiLAW with cross-entropy loss and with focal loss reach similar accuracies.

\begin{table*}[ht!] \centering \caption{Results of two loss functions with varying levels of symmetric noise on CIFAR-100-M. \label{table:losses}} \footnotesize{\begin{tabular}{lccccc} \hline \textbf{Loss Function} & \textbf{Noise Level (\%)} & \textbf{Top-1 Acc. (\%)} & \textbf{Top-5 Acc. (\%)} & \textbf{AUROC} \\ \hline \multirow{2}{*}{Cross-Entropy} 
& 0  & 80.85 \textcolor{BrickRed}{$\downarrow$} 0.01 & 96.07 \textcolor{Green}{$\uparrow$} 0.16 & 0.9888 \textcolor{BrickRed}{$\downarrow$} 0.0005 \\
& 50 & 75.20 \textcolor{Green}{$\uparrow$} 1.35 & 92.61 \textcolor{Green}{$\uparrow$} 0.67 & 0.9811 \textcolor{Green}{$\uparrow$} 0.0013 \\ \hline

\multirow{2}{*}{Focal Loss~\citep{focalloss}} 
& 0  & 80.14 \textcolor{BrickRed}{$\downarrow$} 0.18 & 97.54 \textcolor{BrickRed}{$\downarrow$} 0.06 & 1.0000 \textcolor{BrickRed}{$\downarrow$} 0.0001 \\
& 50 & 75.27 \textcolor{Green}{$\uparrow$} 1.15 & 92.70 \textcolor{Green}{$\uparrow$} 0.62 & 0.9793 \textcolor{Green}{$\uparrow$} 0.0014 \\ \hline \end{tabular}} \end{table*}

\subsection{Effect of validation set size}
\label{supp:valsize}

In Table \ref{table:valsize}, we analyze the effect of varying the validation set size (as a percentage of the training set size) on the test performance. We see that we do not need too much validation data to obtain high performance using LiLAW. Note that we only use one random batch from the validation set. We see that a 15\% validation set size strikes a good balance between having enough validation data for LiLAW while leaving sufficient data for model training. Simply increasing the validation set size does not guarantee better performance. As a result, we conclude that there is minor variability in the boost from LiLAW depending on the validation set size, but LiLAW consistently improves accuracy across all validation set sizes, demonstrating its robustness in noisy settings.

\begin{table}[ht!] \centering \caption{Results with different validation set sizes (as a \% of the training set size) on CIFAR-100-M with 50\% symmetric noise. \label{table:valsize}} \footnotesize{\begin{tabular}{cccc} \hline \textbf{Validation Set Size (\%)} & \textbf{Top-1 Acc. (\%)} & \textbf{Top-5 Acc. (\%)} & \textbf{AUROC} \\ \hline 5  & 75.83 \textcolor{Green}{$\uparrow$} 1.13 & 92.70 \textcolor{Green}{$\uparrow$} 0.59 & 0.9813 \textcolor{Green}{$\uparrow$} 0.0012 \\
10 & 75.20 \textcolor{Green}{$\uparrow$} 1.62 & 92.29 \textcolor{Green}{$\uparrow$} 1.03 & 0.9804 \textcolor{Green}{$\uparrow$} 0.0015 \\
15 & 75.20 \textcolor{Green}{$\uparrow$} 1.35 & 92.61 \textcolor{Green}{$\uparrow$} 0.67 & 0.9811 \textcolor{Green}{$\uparrow$} 0.0013 \\
20 & 75.10 \textcolor{Green}{$\uparrow$} 0.90 & 92.67 \textcolor{Green}{$\uparrow$} 0.56 & 0.9800 \textcolor{Green}{$\uparrow$} 0.0051 \\
25 & 75.21 \textcolor{Green}{$\uparrow$} 1.37 & 92.28 \textcolor{Green}{$\uparrow$} 0.73 & 0.9796 \textcolor{Green}{$\uparrow$} 0.0015 \\
30 & 75.12 \textcolor{Green}{$\uparrow$} 1.56 & 91.87 \textcolor{Green}{$\uparrow$} 1.06 & 0.9783 \textcolor{Green}{$\uparrow$} 0.0017 \\ \hline \end{tabular}}  \end{table}

\subsection{Effect of validation set noise}
\label{supp:valnoise}

As shown in Table~\ref{tab:valnoise}, LiLAW's performance does not deteriorate severely even when the validation set is substantially noisier. Across a range of validation noise from $0\%$ to $90\%$, performance is essentially unchanged with and without LiLAW for Top-1 accuracy, Top-5 accuracy, and AUROC.

\begin{table}[ht!]
\centering
\caption{
Effect of different amounts of symmetric noise in the validation set on CIFAR-100-M, with training noise fixed at 20\%. LiLAW remains stable even when the validation set is more noisy.
}
\label{tab:valnoise}
\setlength{\tabcolsep}{5pt}
\renewcommand{\arraystretch}{1}
\footnotesize{\begin{tabular}{@{}cccc@{}}
\toprule
Validation Noise (\%) & Top-1 Acc. (\%) & Top-5 Acc. (\%) & AUROC \\
\midrule
0  & 78.72 \textcolor{Green}{$\uparrow$} 0.67 & 94.81 \textcolor{Green}{$\uparrow$} 0.23 & 0.9870 \textcolor{Green}{$\uparrow$} 0.0001 \\
10 & 78.72 \textcolor{Green}{$\uparrow$} 0.66 & 94.81 \textcolor{Green}{$\uparrow$} 0.21 & 0.9870 \textcolor{Green}{$\uparrow$} 0.0001 \\
20 & 78.72 \textcolor{Green}{$\uparrow$} 0.65 & 94.81 \textcolor{Green}{$\uparrow$} 0.24 & 0.9870 \textcolor{Green}{$\uparrow$} 0.0001 \\
30 & 78.72 \textcolor{Green}{$\uparrow$} 0.62 & 94.81 \textcolor{Green}{$\uparrow$} 0.21 & 0.9870 \textcolor{Green}{$\uparrow$} 0.0001 \\
40 & 78.72 \textcolor{Green}{$\uparrow$} 0.60 & 94.81 \textcolor{Green}{$\uparrow$} 0.21 & 0.9870 \textcolor{Green}{$\uparrow$} 0.0001 \\
50 & 78.72 \textcolor{Green}{$\uparrow$} 0.58 & 94.81 \textcolor{Green}{$\uparrow$} 0.20 & 0.9870 \textcolor{Green}{$\uparrow$} 0.0001 \\
60 & 78.72 \textcolor{Green}{$\uparrow$} 0.58 & 94.81 \textcolor{Green}{$\uparrow$} 0.21 & 0.9870 \textcolor{Green}{$\uparrow$} 0.0002 \\
70 & 78.72 \textcolor{Green}{$\uparrow$} 0.60 & 94.81 \textcolor{Green}{$\uparrow$} 0.21 & 0.9870 \textcolor{Green}{$\uparrow$} 0.0002 \\
80 & 78.72 \textcolor{Green}{$\uparrow$} 0.56 & 94.81 \textcolor{Green}{$\uparrow$} 0.20 & 0.9870 \textcolor{Green}{$\uparrow$} 0.0002 \\
90 & 78.72 \textcolor{Green}{$\uparrow$} 0.59 & 94.81 \textcolor{Green}{$\uparrow$} 0.20 & 0.9870 \textcolor{Green}{$\uparrow$} 0.0002 \\
\midrule
Mean $\pm$ Std.
& 78.72 \textcolor{Green}{$\uparrow$} 0.611 $\pm$ 0.036
& 94.81 \textcolor{Green}{$\uparrow$} 0.214 $\pm$ 0.012
& 0.9870 \textcolor{Green}{$\uparrow$} 0.00014 $\pm$ 0.00005 \\
\bottomrule
\end{tabular}}
\end{table}

\subsection{Effect of validation set distribution}
\label{supp:valdist}

In Table \ref{table:valdist}, we explore the effect of the validation set distribution on test performance. Namely, we compare a validation set drawn from the same distribution as the training set versus one drawn from a different distribution (created through augmentations like random flipping, rotations, and color jitter). Results indicate that LiLAW is slightly more effective when the validation set matches the training set distribution, though the difference is relatively small. We conclude that the distribution of the validation set has a minor impact on the effectiveness of LiLAW and that LiLAW remains robust even when the validation set distribution differs, which may be useful in cases where all the training data has to be used for training. This suggests that while matching the validation set distribution to the training set is ideal, LiLAW can still provide improvements in noisy settings even when this condition is not perfectly met.

\begin{table*}[ht!] 
\centering \caption{Results with a validation set from the same distribution and from a different distribution than the training set on CIFAR-100-M with 50\% symmetric noise. \label{table:valdist}} \footnotesize{\begin{tabular}{lcccc} 
\hline \textbf{Validation Set Distribution} & \textbf{Noise Level (\%)} & \textbf{Top-1 Acc. (\%)} & \textbf{Top-5 Acc. (\%)} & \textbf{AUROC} \\ \hline 
\multirow{2}{*}{Same distribution} 
& 0  & 80.85 \textcolor{BrickRed}{$\downarrow$} 0.01 & 96.07 \textcolor{Green}{$\uparrow$} 0.16 & 0.9888 \textcolor{BrickRed}{$\downarrow$} 0.0005 \\
& 50 & 75.20 \textcolor{Green}{$\uparrow$} 1.35 & 92.61 \textcolor{Green}{$\uparrow$} 0.67 & 0.9811 \textcolor{Green}{$\uparrow$} 0.0013 \\ \hline

\multirow{2}{*}{Different distribution} 
& 0  & 80.85 \textcolor{BrickRed}{$\downarrow$} 0.08 & 95.99 \textcolor{BrickRed}{$\downarrow$} 0.48 & 0.9964 \textcolor{BrickRed}{$\downarrow$} 0.0007 \\
& 50 & 75.52 \textcolor{Green}{$\uparrow$} 1.01 & 92.38 \textcolor{Green}{$\uparrow$} 0.91 & 0.9804 \textcolor{Green}{$\uparrow$} 0.0020 \\ \hline \end{tabular}} \end{table*}

\subsection{Results on additional general imaging datasets}
\label{supp:moredatasets}

In Table \ref{table:datasets}, LiLAW improves accuracy on all of the datasets (with only a minor drop in some cases). This suggests that LiLAW is beneficial for both simple and complex classification tasks.

\begin{table*}[ht!] \centering \caption{Results on various datasets with different levels of symmetric noise. \label{table:datasets}}  \footnotesize{ \begin{tabular}{l|cccc} \hline \textbf{Dataset} & \textbf{Noise Level (\%)} & \textbf{Top-1 Acc. (\%)} & \textbf{Top-5 Acc. (\%)} & \textbf{AUROC} \\ \hline
\multirow{2}{*}{MNIST-M} 
& 0  & 97.76 \textcolor{BrickRed}{$\downarrow$} 0.08 & 99.97 \textcolor{Green}{$\uparrow$} 0.01 & 0.9946 \textcolor{BrickRed}{$\downarrow$} 0.0002 \\
& 50 & 95.34 \textcolor{Green}{$\uparrow$} 0.55 & 99.85 \textcolor{Green}{$\uparrow$} 0.04 & 0.9959 \textcolor{Green}{$\uparrow$} 0.0002 \\ \hline

\multirow{2}{*}{FashionMNIST-M} 
& 0  & 91.53 \textcolor{BrickRed}{$\downarrow$} 0.05 & 99.94 \textcolor{Green}{$\uparrow$} 0.00 & 0.9874 \textcolor{BrickRed}{$\downarrow$} 0.0019 \\
& 50 & 87.74 \textcolor{Green}{$\uparrow$} 0.02 & 99.46 \textcolor{Green}{$\uparrow$} 0.11 & 0.9837 \textcolor{Green}{$\uparrow$} 0.0019 \\ \hline

\multirow{2}{*}{CIFAR-10-M} 
& 0  & 94.63 \textcolor{Green}{$\uparrow$} 0.03 & 99.80 \textcolor{Green}{$\uparrow$} 0.02 & 0.9941 \textcolor{BrickRed}{$\downarrow$} 0.0002 \\
& 50 & 92.30 \textcolor{Green}{$\uparrow$} 0.40 & 99.26 \textcolor{Green}{$\uparrow$} 0.05 & 0.9898 \textcolor{Green}{$\uparrow$} 0.0007 \\ \hline

\multirow{2}{*}{CIFAR-100-M} 
& 0  & 80.85 \textcolor{BrickRed}{$\downarrow$} 0.01 & 96.07 \textcolor{Green}{$\uparrow$} 0.16 & 0.9888 \textcolor{BrickRed}{$\downarrow$} 0.0005 \\
& 50 & 75.20 \textcolor{Green}{$\uparrow$} 1.35 & 92.61 \textcolor{Green}{$\uparrow$} 0.67 & 0.9811 \textcolor{Green}{$\uparrow$} 0.0013 \\ \hline
\end{tabular}} \end{table*}

\subsection{AUROC and Accuracy for MedMNISTv2}
\label{supp:medmnist}

\begin{figure}[ht!]
    \centering
    \includegraphics[scale=0.174]{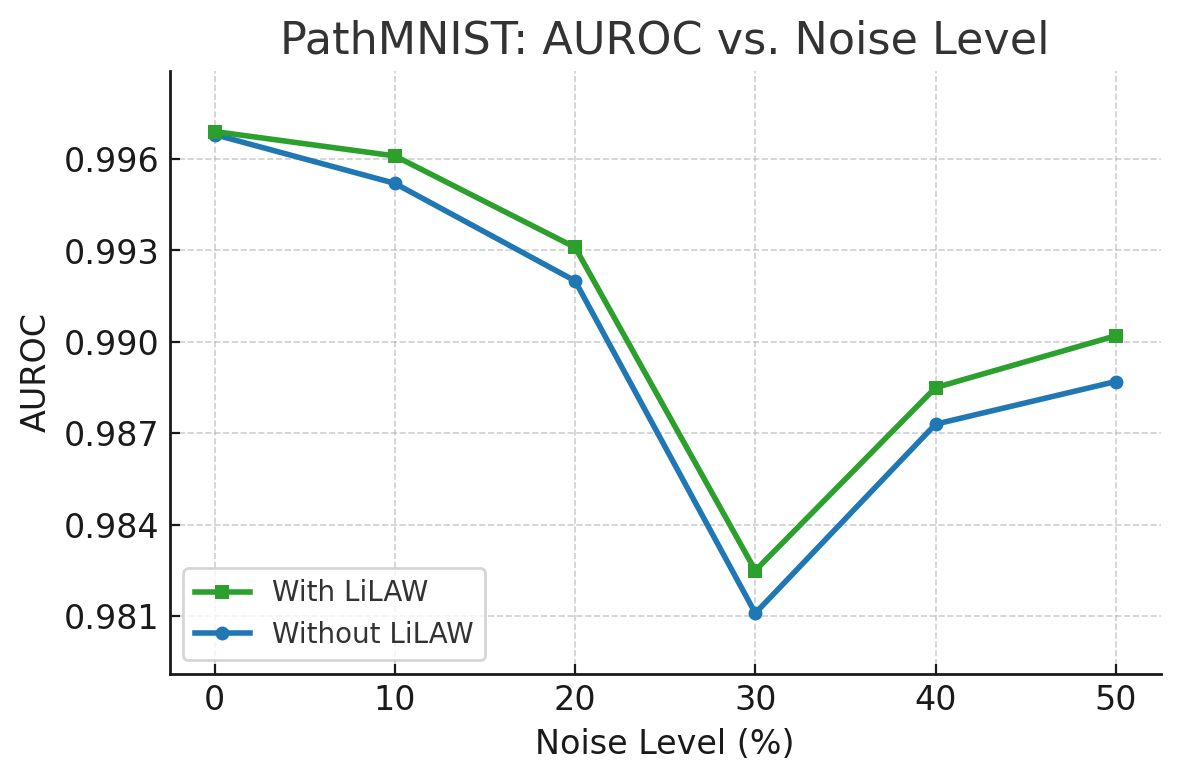}
    \includegraphics[scale=0.174]{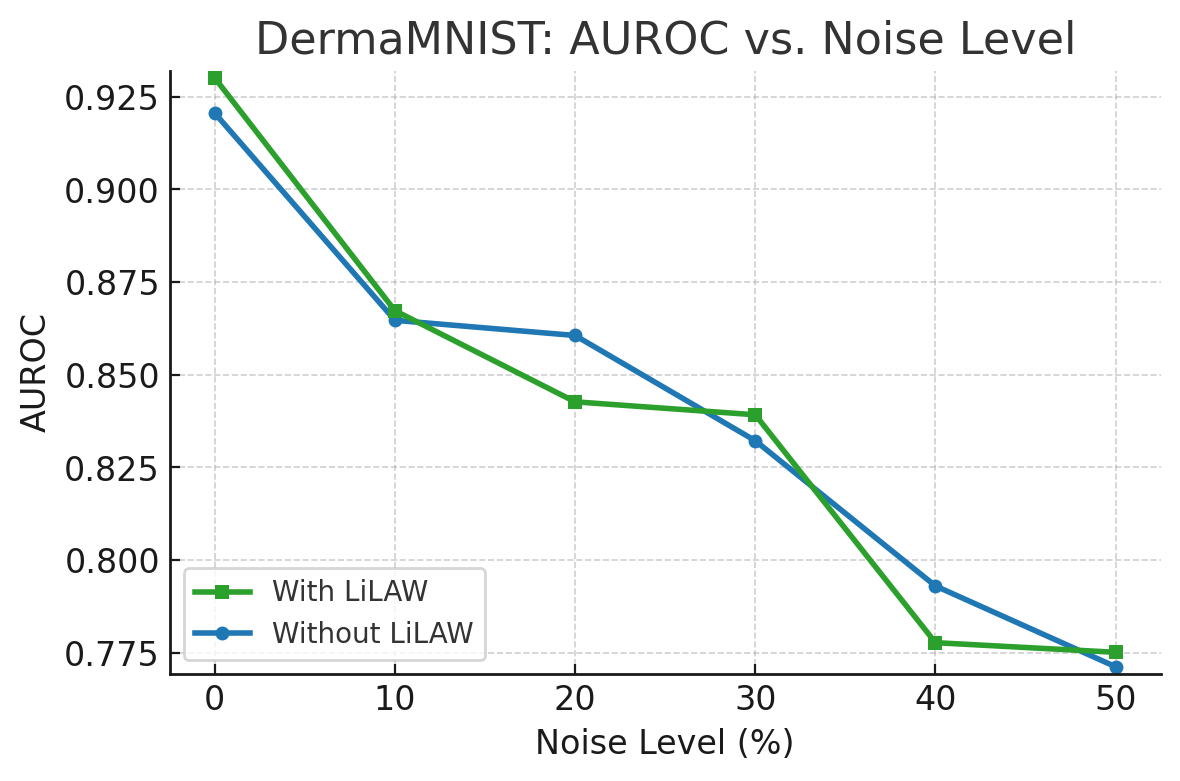}
    \includegraphics[scale=0.174]{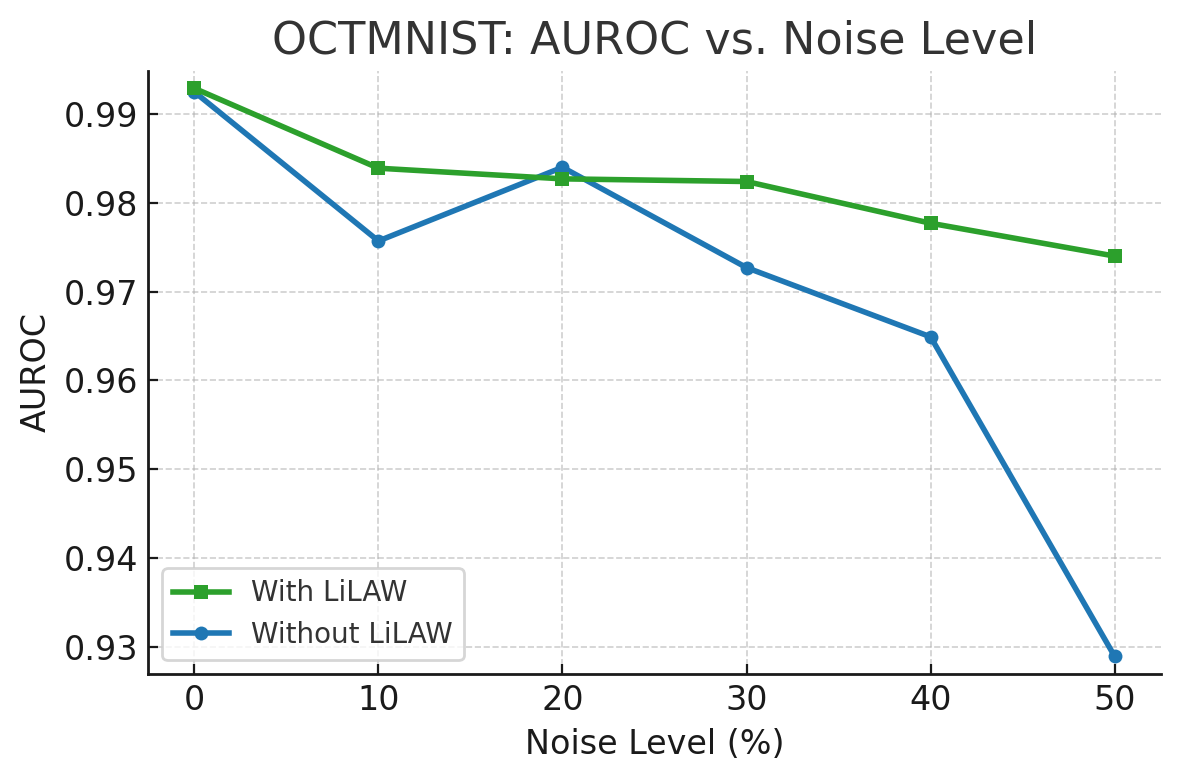}
    \includegraphics[scale=0.174]{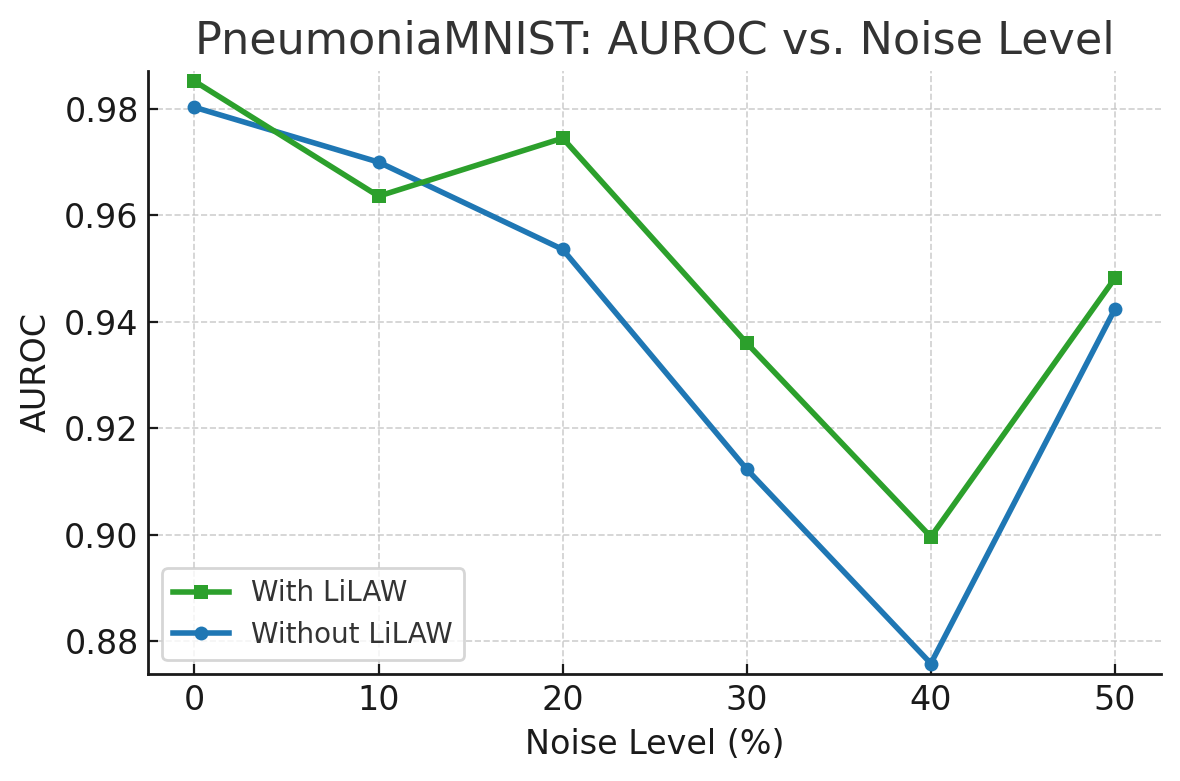}
    \includegraphics[scale=0.174]{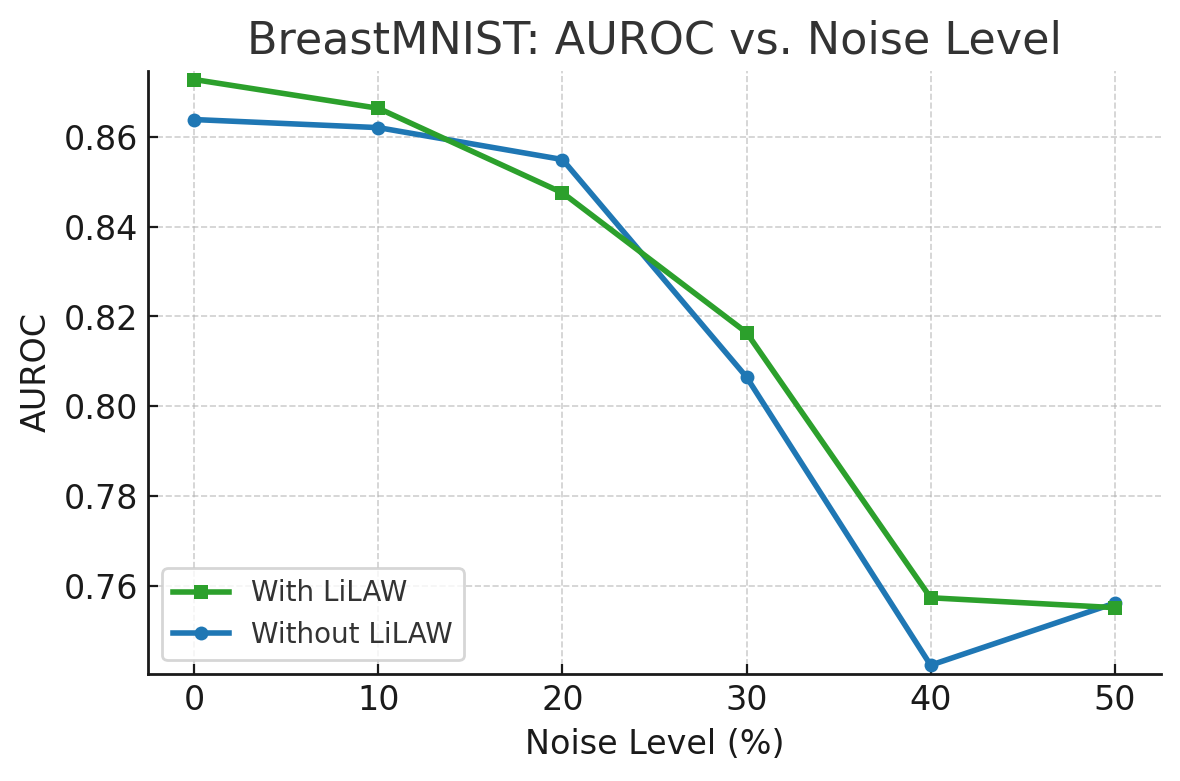}\\
    \includegraphics[scale=0.174]{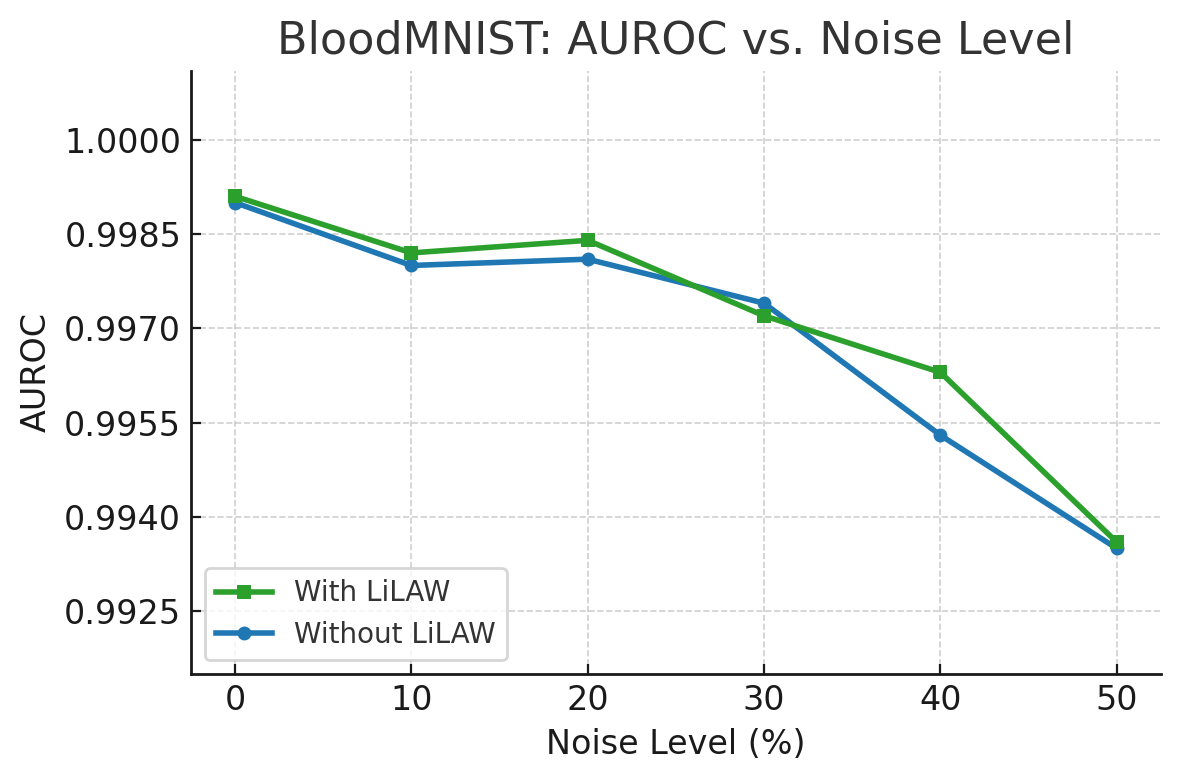}
    \includegraphics[scale=0.174]{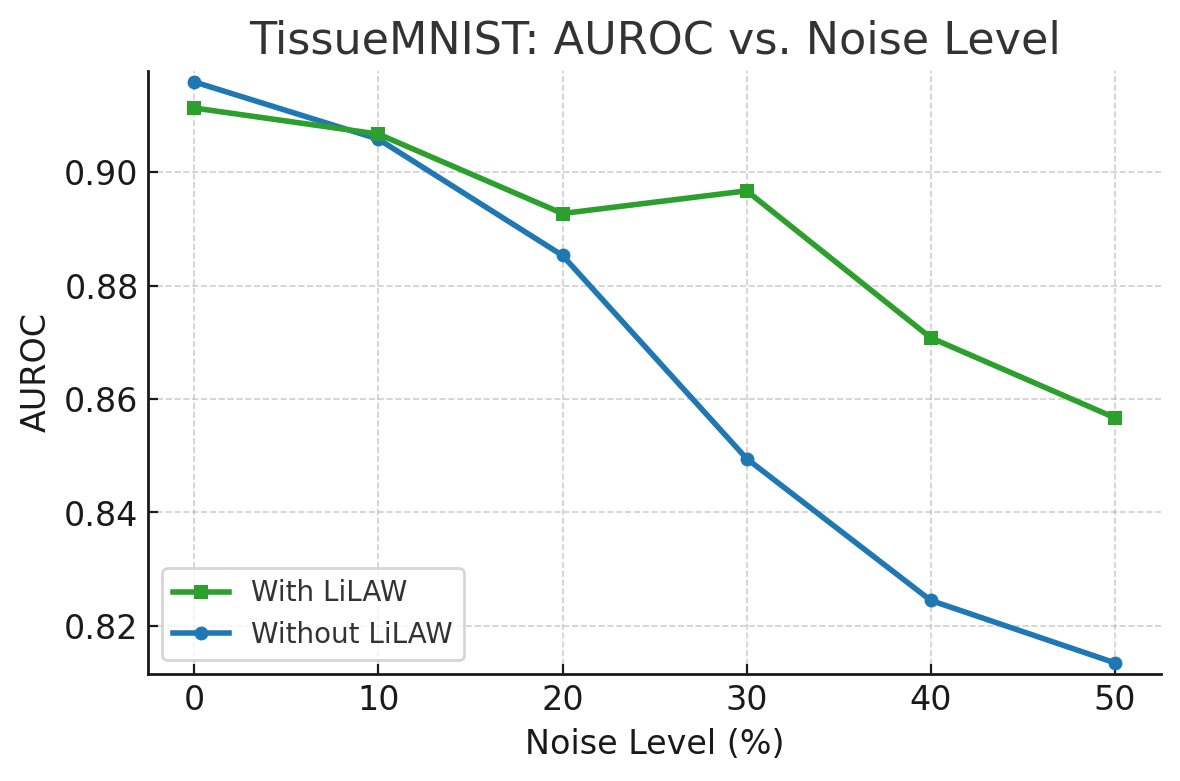}
    \includegraphics[scale=0.174]{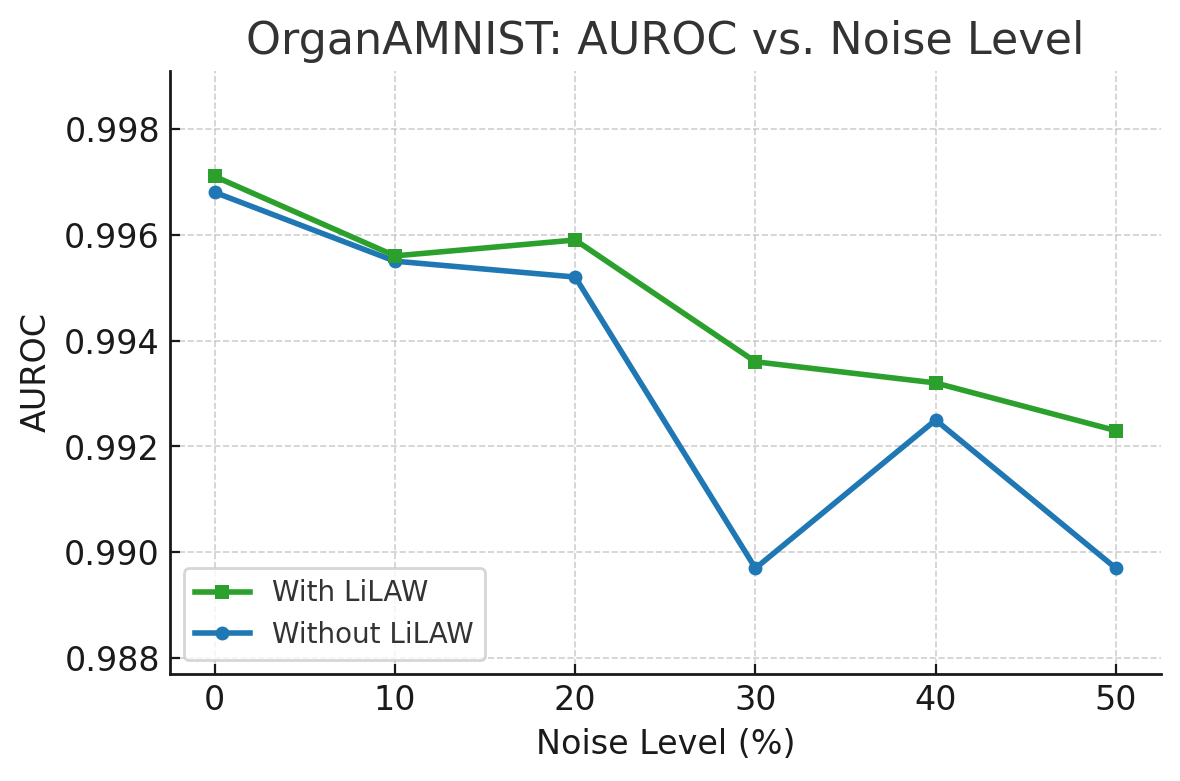}
    \includegraphics[scale=0.174]{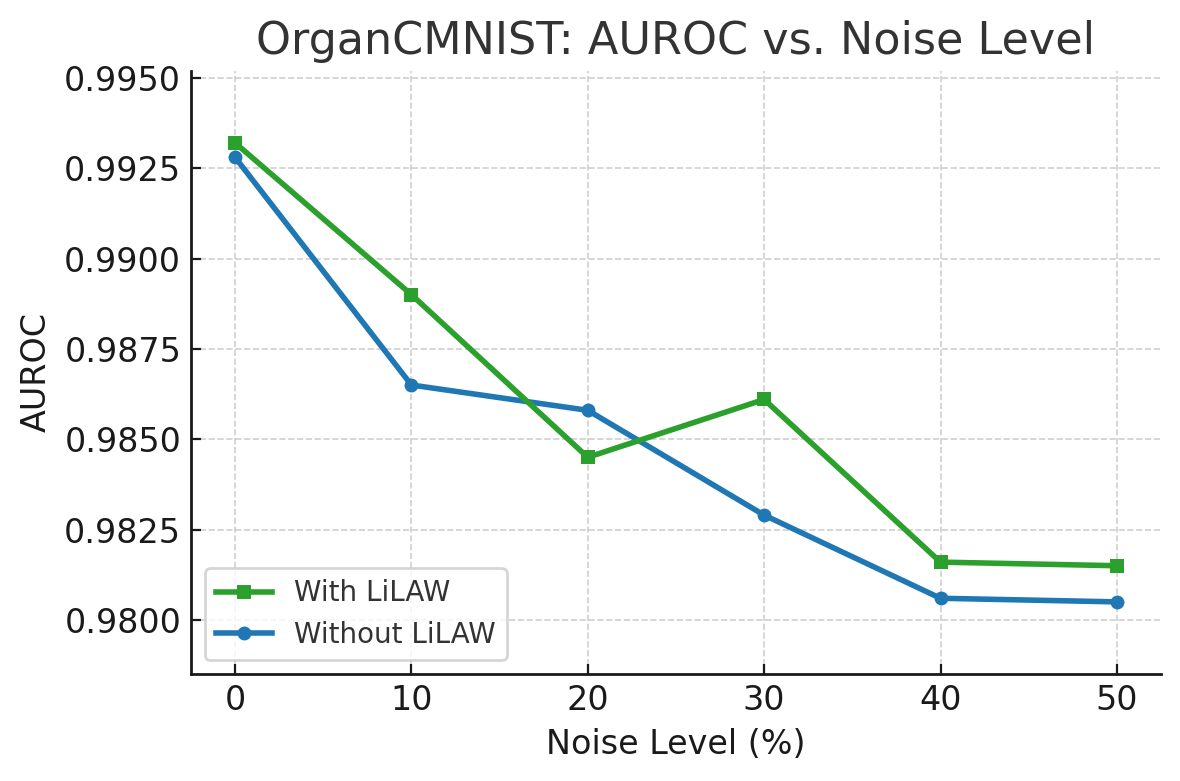}
    \includegraphics[scale=0.174]{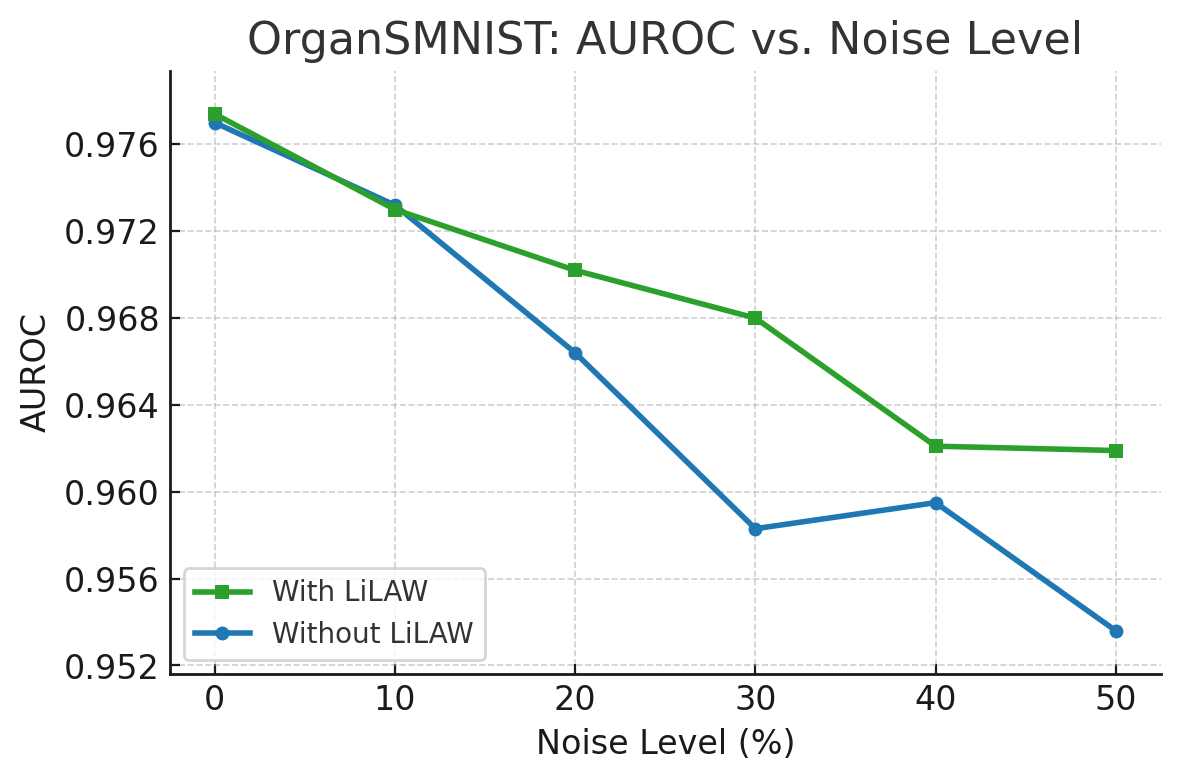}
    \caption{AUROC with and without LiLAW on ten 2D datasets from MedMNISTv2.}
    \label{fig:med2}
\end{figure}

\begin{figure*}[ht!]
    \centering
    \includegraphics[scale=0.171]{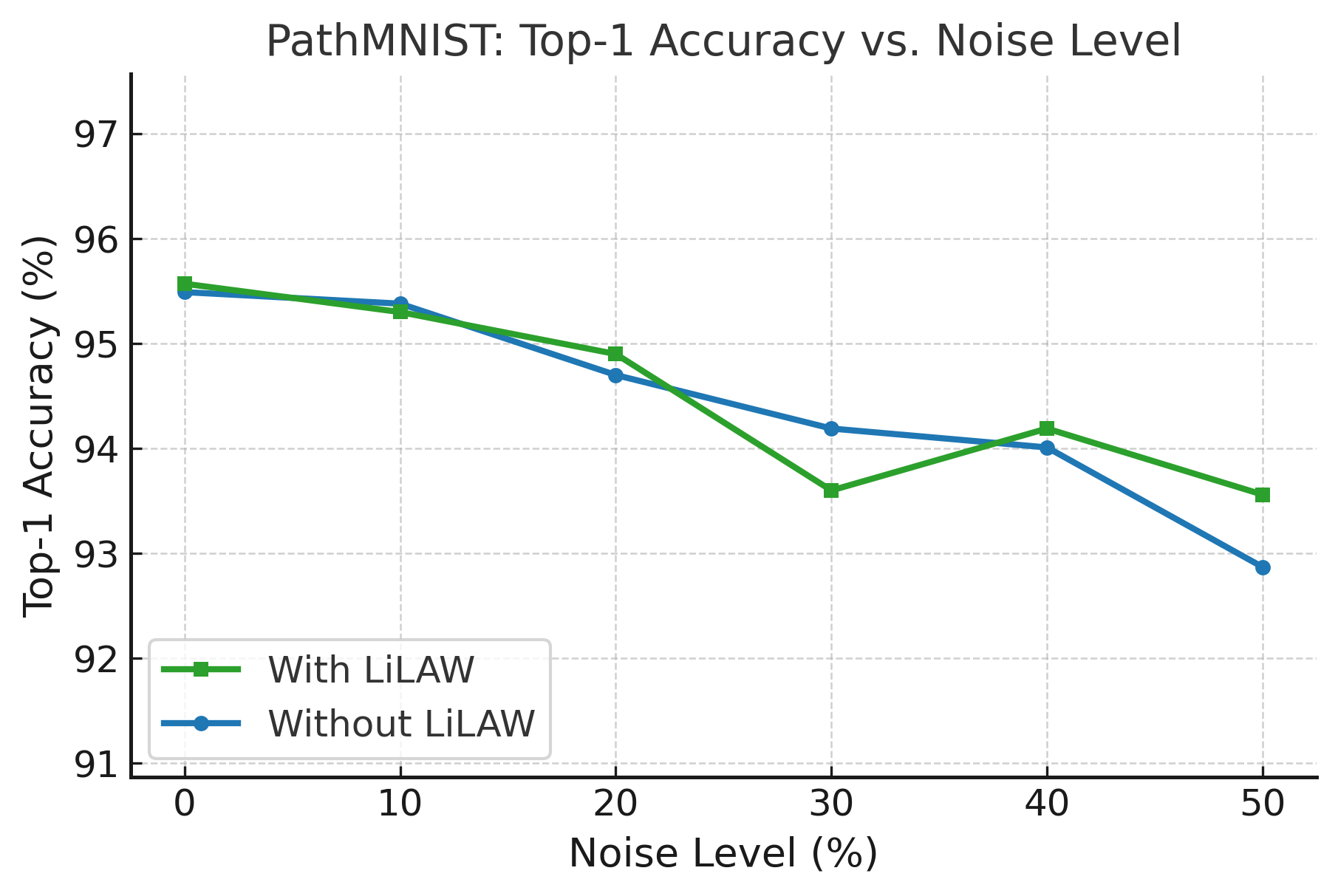}
    \includegraphics[scale=0.171]{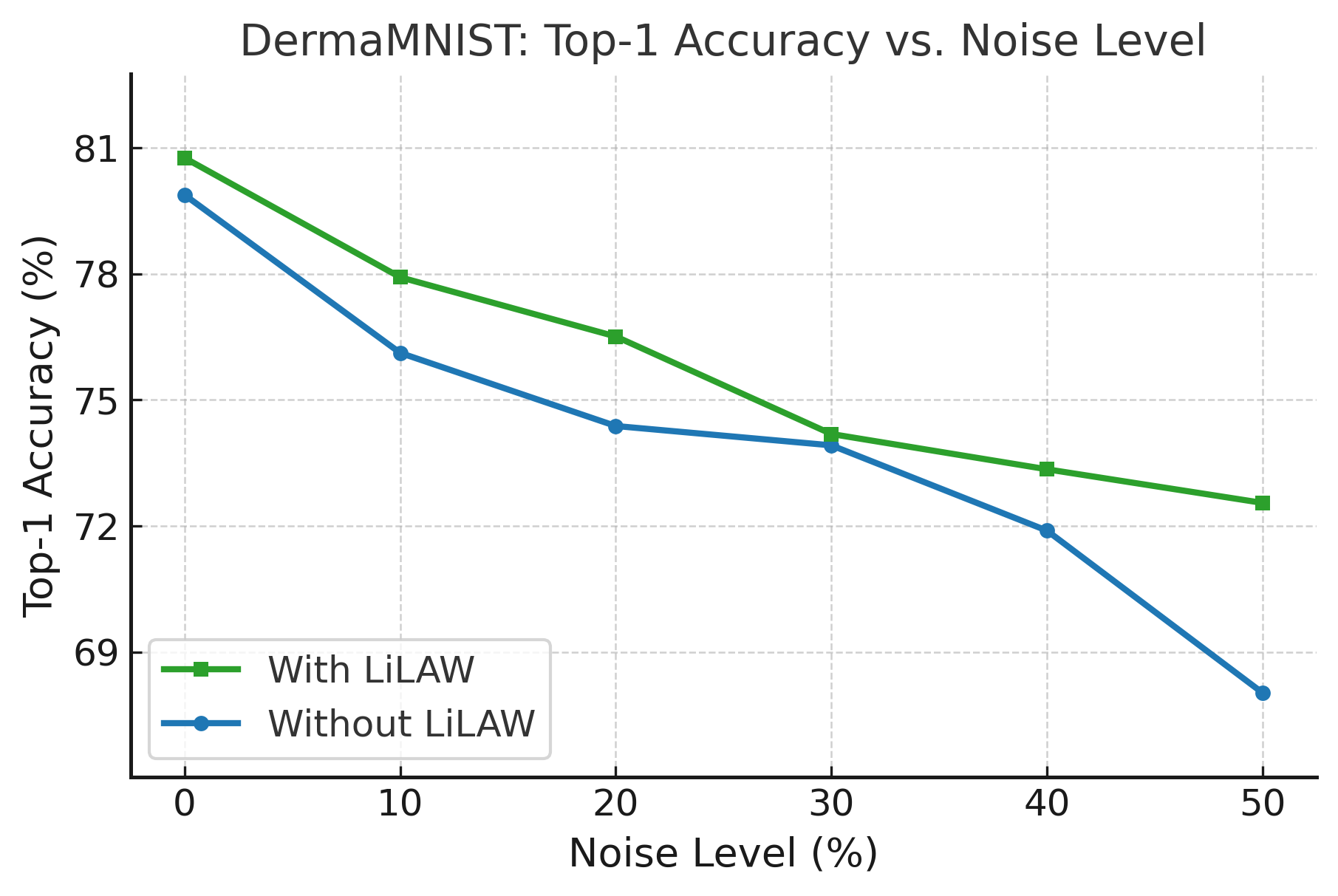}
    \includegraphics[scale=0.171]{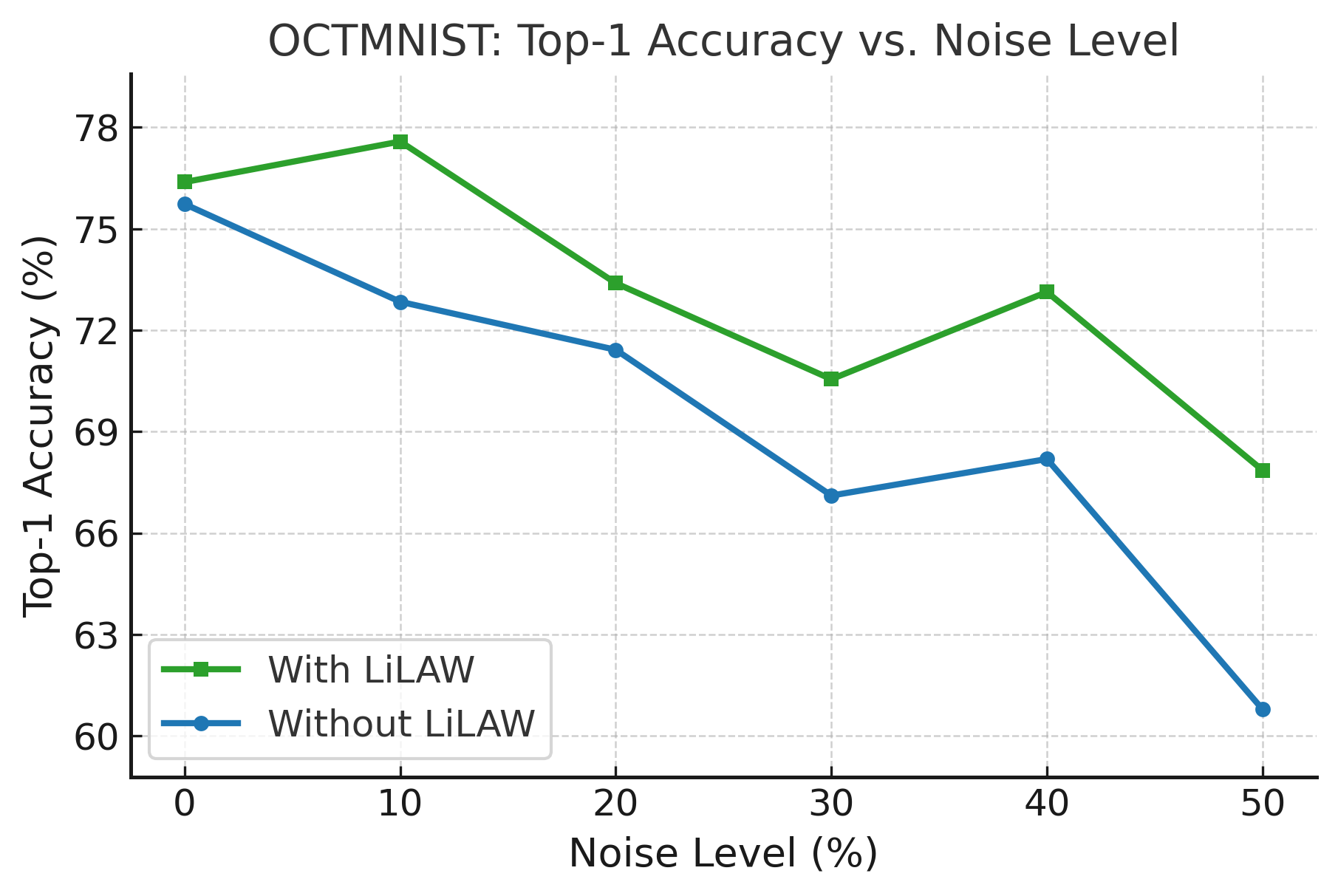}
    \includegraphics[scale=0.171]{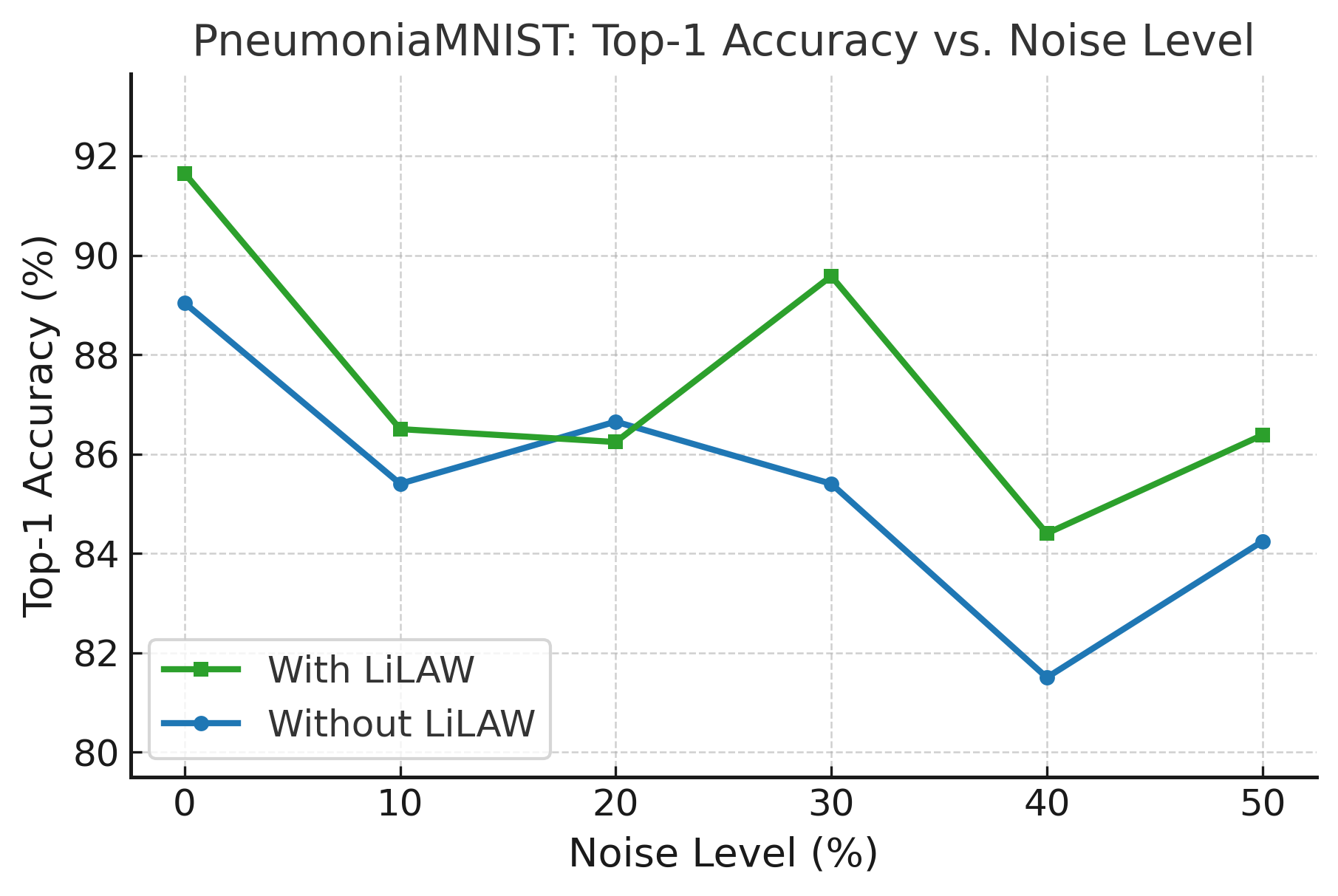}
    \includegraphics[scale=0.171]{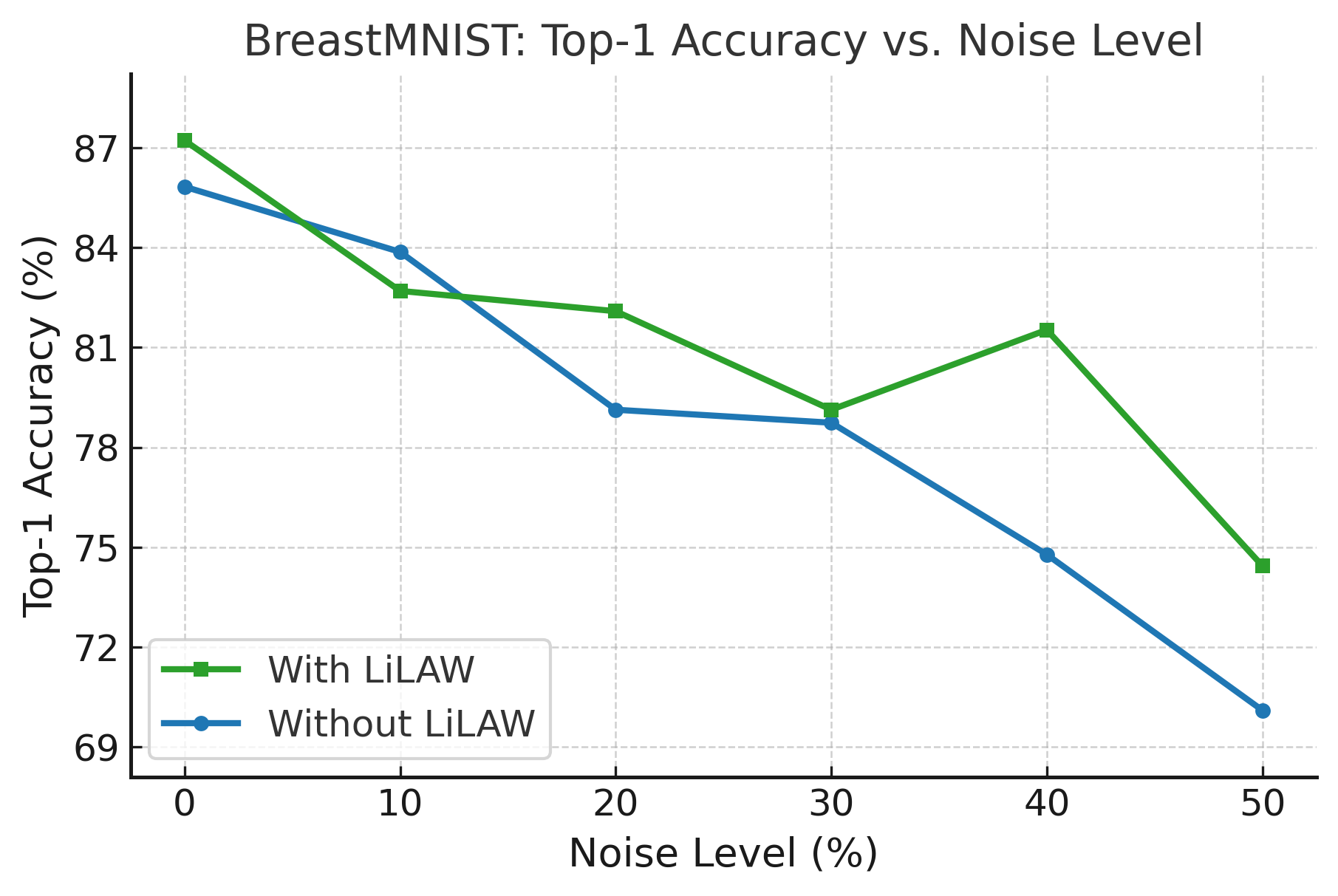}\\
    \includegraphics[scale=0.171]{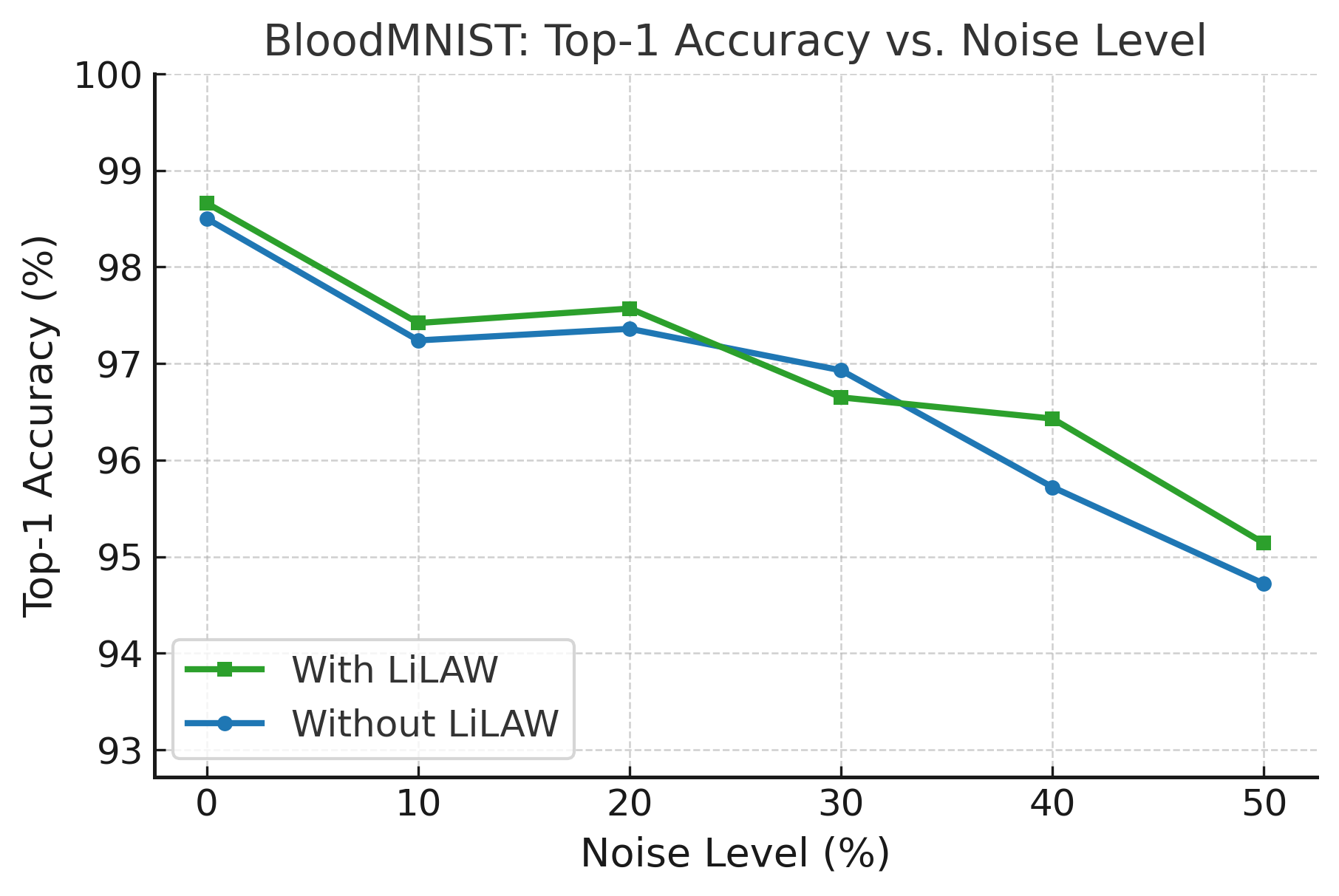}
    \includegraphics[scale=0.171]{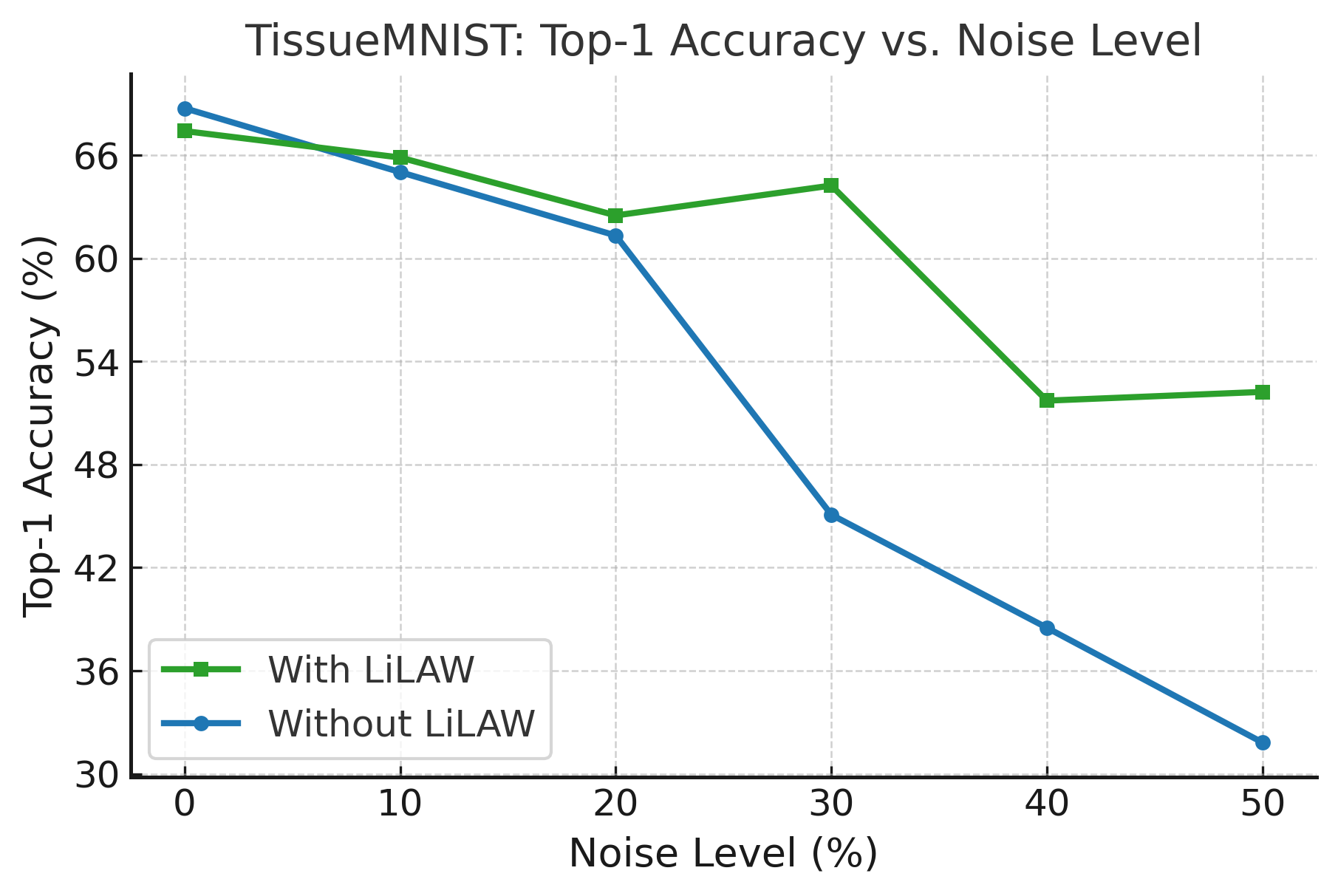}
    \includegraphics[scale=0.171]{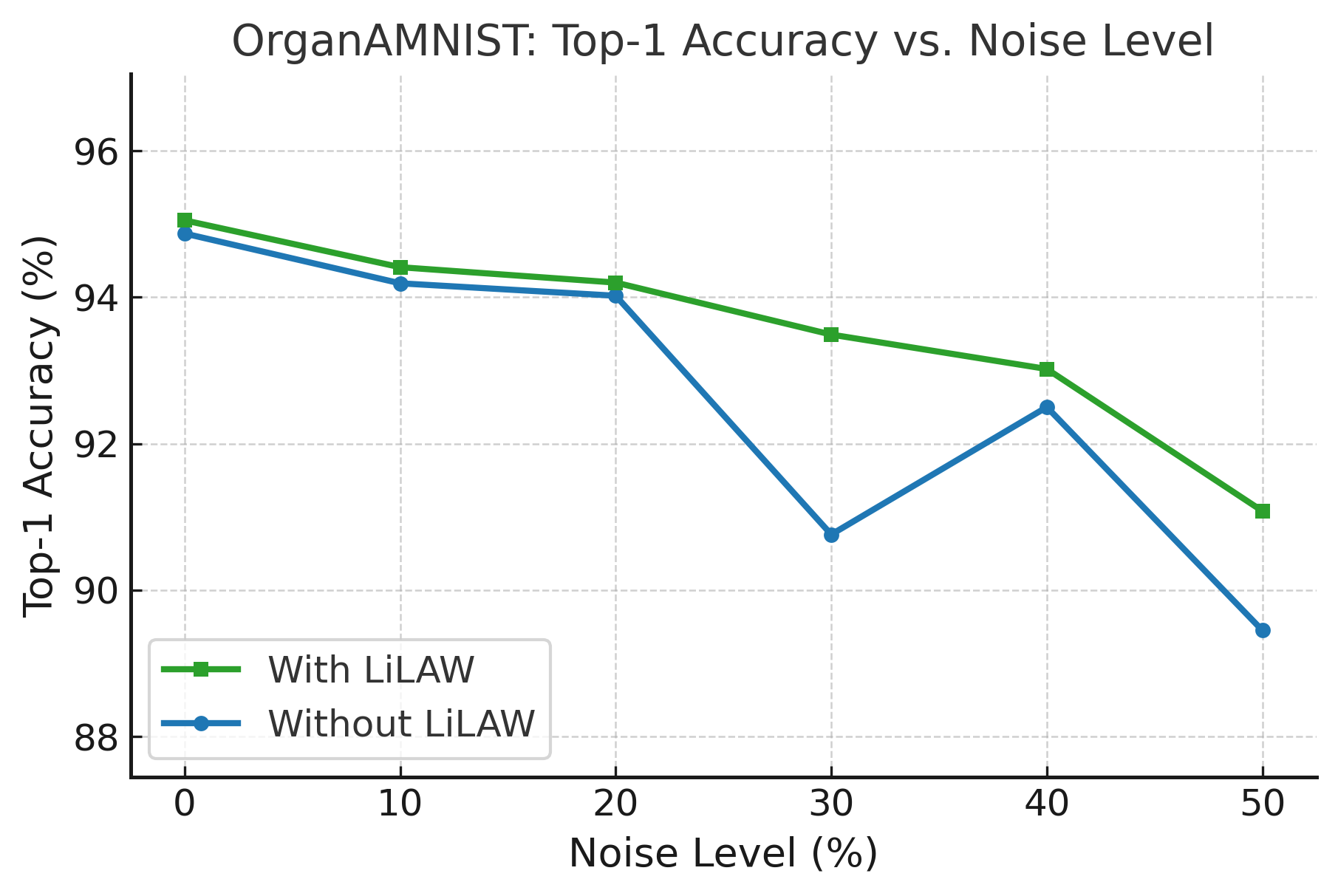}
    \includegraphics[scale=0.171]{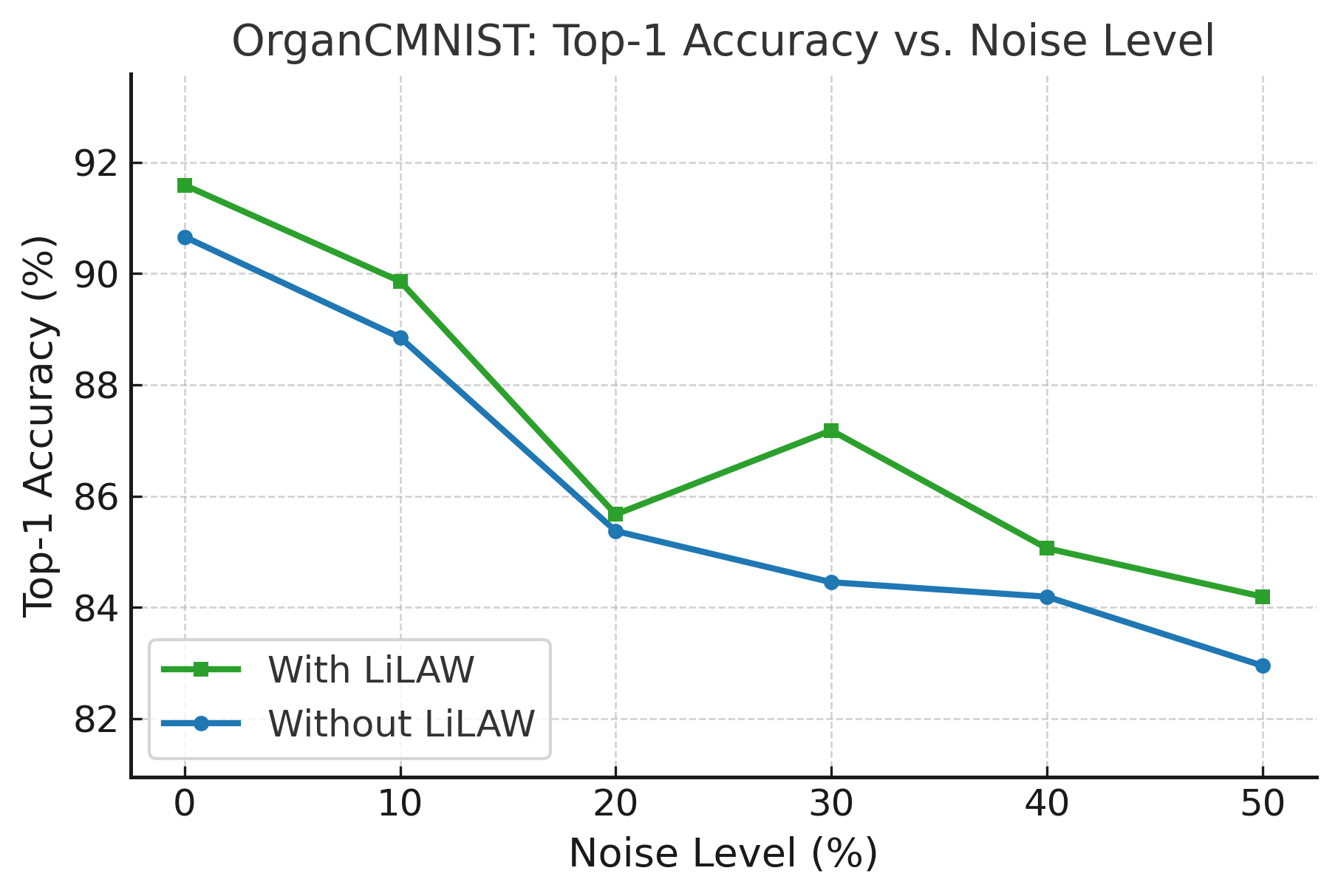}
    \includegraphics[scale=0.171]{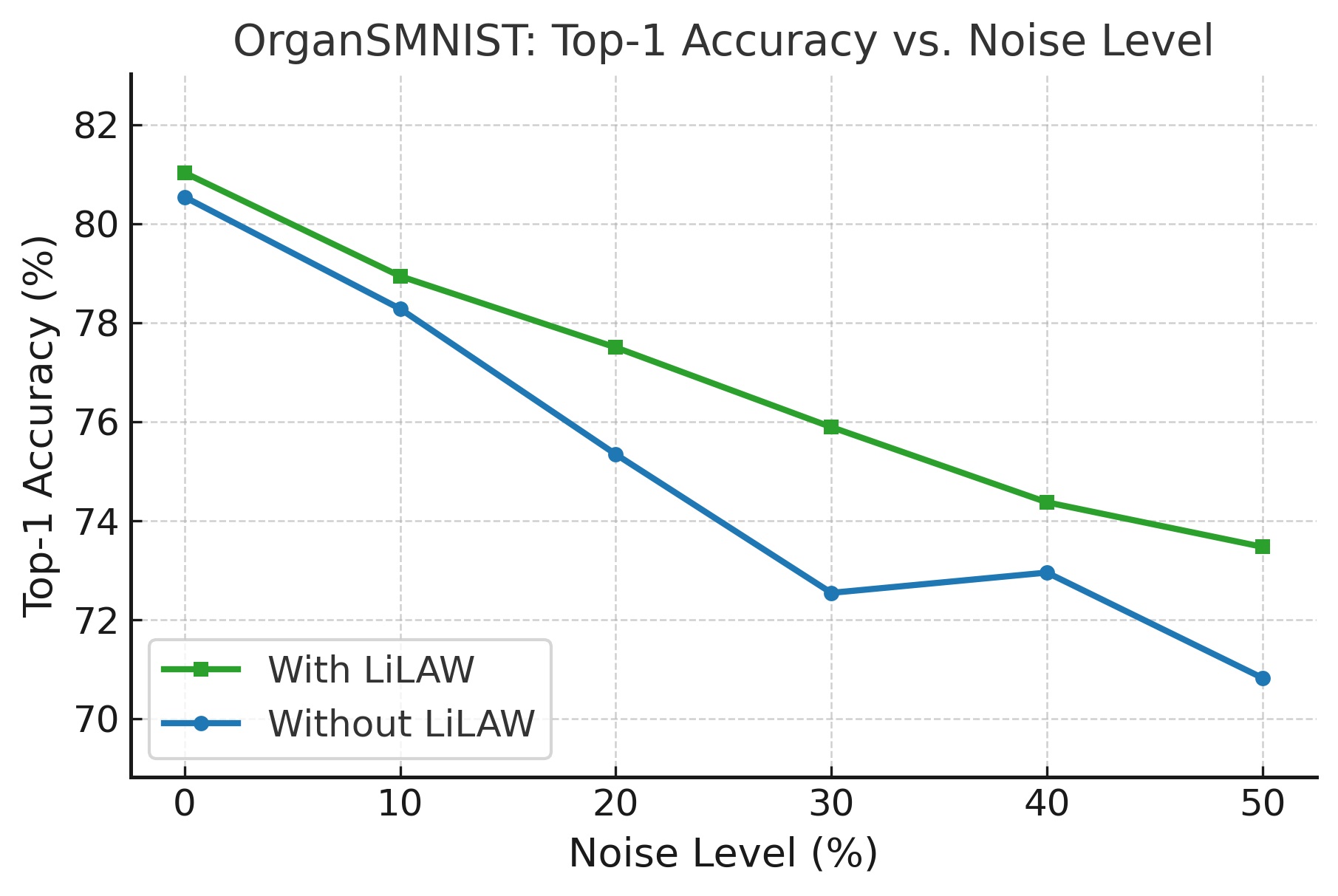}
    \caption{Accuracy with and without LiLAW on ten 2D datasets from MedMNISTv2.}
    \label{fig:med1}
\end{figure*}

In Table \ref{table:medmnist2} and Figure \ref{fig:med2}, we present the AUROC for ten 2D medical imaging datasets from MedMNISTv2 at varying levels of label noise. Across most datasets and noise levels, LiLAW enhances AUROC ranging from minor to substantial gains. LiLAW provides improvements or minor deteriorations across all noise levels. In Figure \ref{fig:med1}, we present the accuracy results shown in Table \ref{table:medmnist}.

\begin{table*}[ht!]
\centering
\caption{AUROC on ten 2D datasets from MedMNISTv2 at different levels of symmetric noise. \label{table:medmnist2}}
\resizebox{\textwidth}{!}{\begin{tabular}{l|cccccc}
\hline
 & & & \textbf{AUROC} & & & \\
\hline
\textbf{Dataset} & \textbf{0\% Noise} & \textbf{10\% Noise} & \textbf{20\% Noise} & \textbf{30\% Noise} & \textbf{40\% Noise} & \textbf{50\% Noise} \\
\hline
PathMNIST & 0.9968 \textcolor{Green}{$\uparrow$} 0.0001 & 0.9952 \textcolor{Green}{$\uparrow$} 0.0009 & 0.9920 \textcolor{Green}{$\uparrow$} 0.0011 & 0.9811 \textcolor{Green}{$\uparrow$} 0.0014 & 0.9873 \textcolor{Green}{$\uparrow$} 0.0012 & 0.9887 \textcolor{Green}{$\uparrow$} 0.0015 \\
\hline
DermaMNIST & 0.9206 \textcolor{Green}{$\uparrow$} 0.0095 & 0.8647 \textcolor{Green}{$\uparrow$} 0.0026 & 0.8606 \textcolor{BrickRed}{$\downarrow$} 0.0179 & 0.8322 \textcolor{Green}{$\uparrow$} 0.0070 & 0.7930 \textcolor{BrickRed}{$\downarrow$} 0.0153 & 0.7712 \textcolor{Green}{$\uparrow$} 0.0039 \\
\hline
OCTMNIST & 0.9925 \textcolor{Green}{$\uparrow$} 0.0004 & 0.9757 \textcolor{Green}{$\uparrow$} 0.0082 & 0.9840 \textcolor{BrickRed}{$\downarrow$} 0.0013 & 0.9727 \textcolor{Green}{$\uparrow$} 0.0097 & 0.9649 \textcolor{Green}{$\uparrow$} 0.0128 & 0.9289 \textcolor{Green}{$\uparrow$} 0.0451 \\
\hline
PneumoniaMNIST & 0.9803 \textcolor{Green}{$\uparrow$} 0.0049 & 0.9700 \textcolor{BrickRed}{$\downarrow$} 0.0064 & 0.9536 \textcolor{Green}{$\uparrow$} 0.0209 & 0.9123 \textcolor{Green}{$\uparrow$} 0.0237 & 0.8758 \textcolor{Green}{$\uparrow$} 0.0238 & 0.9424 \textcolor{Green}{$\uparrow$} 0.0058 \\
\hline
BreastMNIST & 0.8638 \textcolor{Green}{$\uparrow$} 0.0089 & 0.8620 \textcolor{Green}{$\uparrow$} 0.0043 & 0.8549 \textcolor{BrickRed}{$\downarrow$} 0.0074 & 0.8065 \textcolor{Green}{$\uparrow$} 0.0099 & 0.7424 \textcolor{Green}{$\uparrow$} 0.0150 & 0.7562 \textcolor{BrickRed}{$\downarrow$} 0.0010 \\
\hline
BloodMNIST & 0.9990 \textcolor{Green}{$\uparrow$} 0.0001 & 0.9980 \textcolor{Green}{$\uparrow$} 0.0002 & 0.9981 \textcolor{Green}{$\uparrow$} 0.0003 & 0.9974 \textcolor{BrickRed}{$\downarrow$} 0.0002 & 0.9953 \textcolor{Green}{$\uparrow$} 0.0010 & 0.9935 \textcolor{Green}{$\uparrow$} 0.0001 \\
\hline
TissueMNIST & 0.9159 \textcolor{BrickRed}{$\downarrow$} 0.0046 & 0.9058 \textcolor{Green}{$\uparrow$} 0.0009 & 0.8853 \textcolor{Green}{$\uparrow$} 0.0074 & 0.8495 \textcolor{Green}{$\uparrow$} 0.0472 & 0.8245 \textcolor{Green}{$\uparrow$} 0.0463 & 0.8135 \textcolor{Green}{$\uparrow$} 0.0432 \\
\hline
OrganAMNIST & 0.9968 \textcolor{Green}{$\uparrow$} 0.0003 & 0.9955 \textcolor{Green}{$\uparrow$} 0.0001 & 0.9952 \textcolor{Green}{$\uparrow$} 0.0007 & 0.9897 \textcolor{Green}{$\uparrow$} 0.0039 & 0.9925 \textcolor{Green}{$\uparrow$} 0.0007 & 0.9897 \textcolor{Green}{$\uparrow$} 0.0026 \\
\hline
OrganCMNIST & 0.9928 \textcolor{Green}{$\uparrow$} 0.0004 & 0.9865 \textcolor{Green}{$\uparrow$} 0.0025 & 0.9858 \textcolor{BrickRed}{$\downarrow$} 0.0013 & 0.9829 \textcolor{Green}{$\uparrow$} 0.0032 & 0.9806 \textcolor{Green}{$\uparrow$} 0.0010 & 0.9805 \textcolor{Green}{$\uparrow$} 0.0010 \\
\hline
OrganSMNIST & 0.9770 \textcolor{Green}{$\uparrow$} 0.0004 & 0.9732 \textcolor{BrickRed}{$\downarrow$} 0.0002 & 0.9664 \textcolor{Green}{$\uparrow$} 0.0038 & 0.9583 \textcolor{Green}{$\uparrow$} 0.0097 & 0.9595 \textcolor{Green}{$\uparrow$} 0.0026 & 0.9536 \textcolor{Green}{$\uparrow$} 0.0083 \\
\hline
\end{tabular}}
\end{table*}

\subsection{Comparison to baselines for identifying mislabels}
\label{supp:comp}

\begin{table*}[ht!]
\centering
\caption{AUROC and AUPRC comparison of metrics from our method ($\mathcal{W}_\alpha$, $\mathcal{W}_\beta$, $\mathcal{W}_\delta$, $\mathcal{W}$) to identify mislabels (in CIFAR-100 with 50\% symmetric noise) in early-stage (epoch 3) and late-stage (epoch 10) training with 7 other difficulty estimation methods.} % (Note: non-LiLAW metrics were obtained while training without LiLAW).}
\label{table:comp}
\footnotesize{
\begin{tabular}{l|cc|cc}
\toprule
\multirow{2}{*}{\textbf{Method}} & \multicolumn{2}{c|}{\textbf{AUROC}} & \multicolumn{2}{c}{\textbf{AUPRC}} \\
 & \textbf{Epoch 3} & \textbf{Epoch 10} & \textbf{Epoch 3} & \textbf{Epoch 10} \\
\midrule
Data-IQ \citep{seedat_data-iq_2022}      & 0.9363 & 0.8348 & 0.9290 & 0.9746 \\
DataMaps \citep{swayamdipta_dataset_2020}     & 0.5000 & 0.6989 & 0.5000 & 0.8320 \\
CNLCU-S \citep{xia_sample_2022}        & 0.9438 & 0.9026 & 0.3119 & 0.3181 \\
AUM \citep{pleiss_identifying_2020}        & 0.9680 & 0.9635 & 0.9649 & 0.9610 \\
EL2N \citep{paul2021deep}        & 0.9180 & 0.7567 & 0.3161 & 0.3604 \\
GraNd \citep{paul2021deep}       & 0.6402 & 0.7034 & 0.4820 & 0.4054 \\
Forgetting \citep{toneva_empirical_2019}  & 0.5000 & 0.5782 & 0.5000 & 0.5740 \\ \midrule
$\mathcal{W}_\alpha$        & \textbf{0.9838} & \textbf{0.9782} & \textbf{0.9810} & \textbf{0.9755} \\
$\mathcal{W}_\beta$         & 0.9719 & 0.9768 & 0.3086 & 0.3085 \\
$\mathcal{W}_\delta$        & 0.8434 & 0.9258 & 0.3454 & 0.3164 \\
$\mathcal{W}$       & 0.7103 & 0.9069 & 0.4985 & 0.3268 \\
\bottomrule
\end{tabular}}
\end{table*}

In Table~\ref{table:comp}, we evaluate mislabel detection on CIFAR-100 with 50\% symmetric noise at an early training stage (epoch~3) and later training stage (epoch~10). Early-stage signals are very informative for LiLAW with $\mathcal{W}_\alpha$ achieving the best AUROC and the best AUPRC at both stages, with minor changes, compared to all 7 other methods. This indicates that, early in training, we can surface mislabeled points and $\mathcal{W}_\alpha$ captures this separation robustly over time. $\mathcal{W}_\beta$ attains high AUROC but low AUPRC, suggesting it ranks many mislabeled items highly but also hard and correct samples highly. $\mathcal{W}_\delta$ improves AUROC with training while maintaining low AUPRC, suggesting that hard samples get easier to identify over time. % but there may be more false positives. %Note that $\mathcal{W}$ takes the sum of $\mathcal{W}_\alpha$, $\mathcal{W}_\beta$, and $\mathcal{W}_\delta$ to aggregate signal from easy, moderate, and hard samples.

\subsection{Synthetic Datasets}
\label{sec:synthetic}

Synthetic data generation methods include Generative Adversarial Networks (GANs)~\citep{gan}, Variational Autoencoders (VAEs)~\citep{vae}, and diffusion models~\citep{diffusion}, and help address data scarcity, privacy issues, and demographic imbalance~\citep{bauer2024comprehensive}. GANs, such as StyleGAN~\citep{stylegan1,stylegan2}, Progressive GAN~\citep{pggan}, etc., have synthesized high-fidelity images but struggle with diversity and mode collapse. VAEs have also been successful in data generation with controlled latent variables but typically achieve lower fidelity compared to GANs. Diffusion models have achieved very high fidelity compared to GANs and VAEs by starting with random noise and learning to denoise to generate high-quality data. They are usually combined with text-encoders to allow for text prompts~\cite{zhang2024texttoimagediffusionmodelsgenerative}, but may be unable to fully capture physiological or biological data.

Recent methods, such as \synpain~\cite{taati2025synpain}, address these limitations with a diverse synthetic dataset specifically designed for pain classification, thus overcoming demographic bias prevalent in traditional datasets. Similarly, \gaitgen~\cite{adeli2025gaitgen} introduces clinically relevant synthetic gait data conditioned on pathological severity, addressing the scarcity of high-quality labeled clinical data and underrepresentation of severe pathology cases. Despite this, synthetic datasets need to be handled carefully due to variability in generation quality and distributional discrepancies with real data. Incorporating synthetic data does not always improve performance. \textit{Our method helps effectively use synthetic data and data augmentations during training.} %These issues are not fully addressed by standard data augmentation methods.

To demonstrate LiLAW's effectiveness when augmenting real data with synthetic data, we use real datasets (UNBC-McMaster~\citep{lucey2011painful} (CC BY-NC-ND 4.0) and UofR~\citep{rezaei2020unobtrusive} (by request)) and a synthetic dataset (\synpain~\citep{taati2025synpain} (MIT license)) for pain detection. UofR comprises images from people with dementia (Dementia) and without dementia (Healthy). \synpain also contains a subset of images, \synpain-Old, which contain samples from adults of age 75+. For gait classification, we use a real dataset (PD-GaM~\citep{adeli2025gaitgen} (CC BY-NC 4.0)) and a synthetic dataset (\gaitgen~\citep{adeli2025gaitgen} (CC BY-SA 4.0)). We then use the ECG5000~\citep{goldberger2000physiobank} dataset (ODC-By 1.0) with simple augmentations (call this ECG5000-A) containing time-series data for heartbeat classification. We use the following models for experiments on synthetic/augmented data: the Pairwise with Contrastive Training (PwCT) model~\cite{rezaei2020unobtrusive} for the pain detection datasets, the MotionClassifier model~\cite{adeli2025gaitgen} for the gait classification datasets, and a simple Stacked LSTM model for the ECG5000-A dataset.

We achieve state-of-the-art results using LiLAW on the UofR dataset for pain detection with the incorporation of real data from UNBC and UofR and synthetic data from \synpain (see~\ref{sec:pain}, Table~\ref{pain}). Also, we achieve state-of-the-art results using LiLAW on the PD-GaM dataset for gait classification with the incorporation of real data from the PD-GaM dataset and synthetic data from the \gaitgen dataset (see~\ref{sec:gait}, Table~\ref{tab:gait}). And finally, we achieve state-of-the-art results using LiLAW on the ECG5000 dataset using real data from ECG5000 and augmented data from ECG5000-A for heartbeat classification (see~\ref{sec:heart}, Table~\ref{tab:ecg}).

\subsubsection{Pain Detection}
\label{sec:pain}

\textbf{UNBC}-McMaster~\citep{lucey2011painful} (we use UNBC in this paper) contains video data from 25 participants (13 females) with shoulder injuries, recorded during both painful and non-painful movements. The videos were recorded at 30 fps and had a total of 48,391 frames. Each frame is manually annotated with FACS~\citep{ekman1978facial} codes, allowing calculation of the Prkachin and Solomon Pain Intensity (PSPI)~\citep{prkachin2008structure} score. This dataset is publicly available and widely used as a benchmark for pain expression recognition research.

\textbf{UofR}~\citep{rezaei2020unobtrusive} contains video recordings of 102 older adult participants, both with and without dementia. Each session was recorded at 15 fps across two conditions: a baseline lying state and an examination state where a licensed physiotherapist assisted movements to locate painful areas. After removing non-frontal frames, UofR has a total of 162,629 frames. Manual annotations were provided for 95 participants (74 females) using both PSPI and PACSLAC-II pain rating scales, of which 47 cognitively healthy older adults and 48 were residents of long-term care with severe dementia. The test results are reported on the subset of participants with dementia (\textbf{Dementia}) and without dementia (\textbf{Healthy}). Additionally, results on the entire test set are also reported (\textbf{All}). 

\textbf{\synpain}~\citep{taati2025synpain} contains synthetic image pairs consisting of one neutral expression and one expressive image for each identity. Using Ideogram 2.0's commercial API, the authors generated synthetic identities along with their corresponding neutral and expressive image pairs. The dataset encompasses the following demographic and expression categories: age groups include young (20-35) and old (75+); ethnicity/race covers White, Black, South Asian, East Asian, and Middle Eastern; gender represents male and female; expression types span pain (proxy PSPI score of 1) and non-pain (proxy PSPI score of 0). Each synthetic identity comprises two corresponding images (neutral and expressive), with a total of 10,710 images across 5,355 pairs. \textbf{\synpain-Old} refers to the subset of \synpain which only contains images of the old (75+) age group. There are a total of 5,790 images across 2,895 pairs in \synpain-Old.

\textit{Model.} Pairwise with Contrastive Training (\textbf{PwCT}~\citep{rezaei2020unobtrusive}) is the current state-of-the-art model for detecting painful expressions in older adults. It is trained on UNBC and UofR datasets for regression (so we use R=16 since PSPI $\in [0,16]$, as mentioned in \ref{sec:regression}), but we threshold the results to obtain binary pain classification. The two key ideas of PwCT are personalized neutral baselines comparing each test expression to that person's own neutral face to reduce age-related idiosyncrasies and contrastive representation learning to improve cross-dataset generalization. The model has been externally validated in vivo and is currently being evaluated in situ. We use the pre-trained PwCT model and fine-tune it with \synpain in this paper using the setup described in~\citet{rezaei2020unobtrusive}.

\begin{table*}[ht!]
\centering
\caption{Results of fine-tuning the PwCT model with and without LiLAW under different dataset configurations (UofR, UNBC, \synpain, and \synpain-Old). The top 2 results from each column of the UofR test set (Dementia, Healthy, and All) are in \textbf{bold}.}
\label{pain}
\begin{footnotesize}{
\begin{tabular}{cc|ccc}
\toprule
 \multicolumn{2}{c}{\textbf{Dataset configurations}} & \multicolumn{3}{c}{\textbf{AUROC}} \\ \midrule
\textbf{Training} & \textbf{Validation (LiLAW)} & \textbf{Dementia} & \textbf{Healthy} & \textbf{All} \\ \midrule
UofR, UNBC & - & \textbf{0.787} & 0.763 & 0.775 \\ \midrule
& - & 0.761 & 0.774 & 0.767 \\
UofR, UNBC, \synpain & UofR & 0.767 & 0.801 & 0.784 \\
& UofR, \synpain & 0.778 & 0.794 & 0.786 \\ \midrule
& - & 0.778 & 0.779 & 0.778 \\
& UofR, \synpain-Old & 0.782 & 0.796 & \textbf{0.789} \\
UofR, UNBC, \synpain-Old  & UofR, UNBC, \synpain-Old & 0.763 & 0.796 & 0.780 \\
 & \synpain-Old & 0.754 & \textbf{0.803} & 0.779 \\
 & UofR & \textbf{0.784} & \textbf{0.803} & \textbf{0.793} \\
\bottomrule
\end{tabular}}

\end{footnotesize}
\end{table*}

The results on pain classification in Table~\ref{pain} demonstrate the interaction between heterogeneous real datasets, synthetic data augmentations, and LiLAW within the PwCT framework. When trained only on UofR and UNBC, PwCT achieves strong baseline AUROC on the Dementia subgroup (0.787), but relatively weaker performance on Healthy participants (0.763). Incorporating synthetic data from \synpain or \synpain-Old without LiLAW gives inconsistent results: while Healthy AUROC occasionally improves, Dementia AUROC and All AUROC often stagnate or decline, suggesting that naively combining real and synthetic samples risks degradation. With LiLAW, however, the validation-guided update  effectively reweighs real and synthetic examples. The best outcome is observed when training on UofR, UNBC, and \synpain-Old while validating on UofR, resulting in AUROCs of 0.784 (Dementia), 0.803 (Healthy), and 0.793 (All). These results highlight that LiLAW selectively downweights misleading or difficult synthetic samples. We also achieve state-of-the-art results on the UofR test set. LiLAW enables PwCT to integrate synthetic data in a controlled and clinically meaningful way.

\subsubsection{Gait classification}
\label{sec:gait}
\textbf{PD-GaM}~\citep{adeli2025gaitgen} is a fully anonymized, publicly available 3D mesh dataset of parkinsonian gait derived from PD4T~\citep{dadashzadeh2023pecopparameterefficientcontinual}. It contains 1,701 segmented walking sequences from 30 individuals with Parkinson's disease, each labeled by an expert with UPDRS-gait scores from 0 to 3 (score 4 is typically non-ambulatory). Sequences are extracted from video at 25 fps, with post-processing to correct global trajectory artifacts. PD-GaM is the largest public UPDRS-gait-annotated mesh dataset, designed to mitigate data scarcity (especially at higher severities) and to support clinically relevant evaluation metrics. We use its validation set for LiLAW.

\textbf{\gaitgen}~\citep{adeli2025gaitgen} is a generative framework that synthesizes realistic gait sequences conditioned on Parkinson's severity (UPDRS-gait scores from 0 to 3). It disentangles motion dynamics from pathology using a Conditional Residual VQ-VAE, then generates base sequences with a Mask Transformer and refines details via a Residual Transformer, enabling precise control over impairment level. Clinician studies report near-chance discrimination between real and synthetic clips.

\textit{Model.} We use the \textbf{MotionClassifier}~\citep{adeli2025gaitgen} model to train a lightweight supervised head on a fixed 512-D motion embeddings extracted from each gait sequence by the evaluation backbone. The classifier is a 3-layer MLP that produces class logits optimized using Adam for 7000 epochs with batch size 256 and learning rate 0.00001 when synthetic data is used and 0.0001 when synthetic data is not used. We train the MLP from scratch using the same setup as before.

\begin{table*}[ht!] \centering
\caption{Results of the MotionClassifier model with PD-GaM and with both PD-GaM + \gaitgen without LiLAW along with the increase (signified by \textcolor{Green}{$\uparrow$}) in performance with LiLAW.}
\label{tab:gait}
\footnotesize{\begin{tabular}{c|ccc}
\hline
\textbf{Dataset} & \textbf{Recall} & \textbf{Precision} & \textbf{Macro-F1} \\
\hline
PD-GaM & 70.09 \textcolor{Green}{$\uparrow$} 0.03\hspace{0.5em} & 75.32 \textcolor{Green}{$\uparrow$} 1.15\hspace{0.5em} & 70.29 \textcolor{Green}{$\uparrow$} 0.98\hspace{0.5em} \\
PD-GaM + \gaitgen & 73.94 \textcolor{Green}{$\uparrow$} 3.51\hspace{0.5em} & 80.20 \textcolor{Green}{$\uparrow$} 1.17\hspace{0.5em} & 75.95 \textcolor{Green}{$\uparrow$} 2.60\hspace{0.5em} \\
\hline
\end{tabular}}
\end{table*}

\iffalse
\begin{figure}
    \centering
    \includegraphics[scale=0.4]{motionclassifier.png}
    \caption{Recall, precision, and macro-F1 scores of the MotionClassifer with and without synthetic data and with and without LiLAW.}
    \label{fig:motionclassifier}
\end{figure}
\fi 

In the gait classification task, Table~\ref{tab:gait} shows how LiLAW enhances the MotionClassifier trained on PD-GaM and synthetic gait sequences generated by \gaitgen. On PD-GaM alone, the MotionClassifier achieves recall of 70.09, precision of 75.32, and macro-F1 of 70.29. With LiLAW, each metric improves modestly, stabilizing training even without synthetic data. However, when synthetic \gaitgen samples are introduced, baseline performance rises substantially to a recall of 73.94, precision of 80.20, macro-F1 of 75.95, but LiLAW amplifies these gains even further, with recall increasing by 3.51, precision by 1.17, and macro-F1 by 2.60. These results achieve state-of-the-art for the PD-GaM test dataset. These gains are especially important clinically, as recall corresponds to sensitivity in detecting higher-severity Parkinsonian gait impairments. Rather than overfitting to synthetic distributions, LiLAW leverages them and demonstrates that synthetic augmentation paired with LiLAW can help improve subgroup generalization.

\subsubsection{Heartbeat classification}
\label{sec:heart}
\textbf{ECG5000}~\citep{goldberger2000physiobank} is a five-class heartbeat dataset derived from a single 20-hour ECG tracing in the BIDMC Congestive Heart Failure Database (in PhysioNet). The preprocessing isolates individual heartbeats and interpolates each to a fixed length. 5,000 heartbeats are randomly selected. Labels are obtained via automated annotation. 5\% of the training set is reserved for the validation-guided update.

To introduce variability and simulate sensor and physiological noise, we augment 50\% of the training heartbeats with additive Gaussian perturbations that combine a random offset and stochastic variance. We call this augmented dataset \textbf{ECG5000-A}. Class labels are retained for augmented samples. For a heartbeat sequence $x$, the augmented sample is:
\[
\tilde{x} \;=\; x \;+\; b \;+\; \sigma \varepsilon,\quad
b \sim \mathcal{U}[-0.2,\,0.2],\ \ \sigma=\sqrt{u},\ u \sim \mathcal{U}[0.05,\,0.1],\ \ \varepsilon \sim \mathcal{N}(0,I).
\]

\textit{Model.} We train on ECG5000 using a \textbf{Stacked LSTM} model (containing 2 LSTM layers) from scratch. We use the Adam optimizer for 500 epochs with batch size 512 and learning rate 0.001. %We train the model using the same setup as before.

\begin{table}[ht]
\begin{footnotesize}
\centering
\caption{Results of the Stacked LSTM model with ECG5000 and with ECG5000-A without LiLAW along with the increase (signified by \textcolor{Green}{$\uparrow$}) in performance with LiLAW.}
\label{tab:ecg}
\begin{tabular}{c|cc}
\hline
\textbf{Dataset} & \textbf{Acc. (\%)} & \textbf{AUROC} \\
\hline
ECG5000 & 93.60 \textcolor{Green}{$\uparrow$} 3.20 & 0.9982 \textcolor{Green}{$\uparrow$} 0.0005 \\
ECG5000-A & 96.80 \textcolor{Green}{$\uparrow$} 3.20 & 0.9907 \textcolor{Green}{$\uparrow$} 0.0093 \\
\hline
\end{tabular}

\end{footnotesize}
\end{table}

\iffalse
\begin{figure}
    \centering
    \includegraphics[scale=0.4]{ecg5000_accuracy.png}
    \includegraphics[scale=0.4]{ecg5000_auroc.png}
    \caption{Accuracy and AUROC of Stacked LSTM model with and without augmented data and with and without LiLAW.}
    \label{fig:ecg5000}
\end{figure}
\fi

The ECG heartbeat classification results in Table~\ref{tab:ecg} highlight LiLAW's utility in time-series domains where baseline performance is already saturated. Using a Stacked LSTM trained from scratch, ECG5000 alone achieves 93.60\% accuracy and an AUROC of 0.9982. Even in this high-performing setting, LiLAW yields improvements (accuracy up by 3.20\%, AUROC  up by 0.0005). When training on ECG5000-A, which contains classical data augmentations, accuracy increases to 96.80\%, but AUROC drops to 0.9907, suggesting that the augmentation increases accuracy but decreases discrimination/ranking performance. LiLAW mitigates this imbalance by improving the AUROC by 0.0093 while preserving accuracy gains. These results achieve state-of-the-art for the ECG5000 test dataset.

Compared to standard training, LiLAW yields a substantial improvement in accuracy and AUROC, demonstrating its ability to enhance predictive performance and discrimination in inherently noisy domains. Namely, LiLAW can dynamically reweight samples based on difficulty to effectively mitigate the impact of inherent noise in physiological signals. This suggests that LiLAW provides a generalizable framework to improve robustness across modalities where label quality and data heterogeneity are common challenges.

\subsection{Improving Fairness}
\label{fairness}

Reweighting is also used as a lightweight method to improve fairness since it can give different training samples different importance in the loss. Early methods assigned weights to groups and labels proportionally during pre-processing~\citep{kamiran}, however later methods have made reweighting adaptive to better satisfy fairness objectives while training~\citep{krasanakis, chai}. More recent methods aim to achieve better fairness through meta-learning, where weights are optimized jointly with the model parameters~\citep{yan2020forml}. \textit{Our method also assigns the learned adaptive weights to each sample's loss and improves fairness across subgroups.}

We show that LiLAW improves performance and generally improves fairness across sex, race, and age with the Adult dataset~\citep{adult} for income prediction. We use the following model for experiments on fairness: a simple 3-layer MLP for the Adult census dataset.

Although it may seem that sample reweighting may downweight underrepresented classes or subgroups, LiLAW does not optimize weights directly as a function of class frequency, subgroup identity, or protected attributes. Instead, LiLAW optimizes $W(s_i,\tilde y_i)$, which depends on the sample's current confidence/disagreement geometry. Specifically, the weight depends on the model score assigned to the observed label, $s_i[\tilde y_i]$, and the maximum predicted score, $\max(s_i)$. As a result, LiLAW is not a class-frequency reweighting rule and does not explicitly encode protected attributes. If an underrepresented class or subgroup is systematically harder because of imbalance, distribution shift, or label noise, then LiLAW can increase its effective training signal by assigning larger weights to samples that fall into harder or more ambiguous regions of the confidence/disagreement plane. This is the intuition behind the Adult dataset results below, which show that LiLAW can improve fairness empirically when subgroup disadvantage manifests as greater sample difficulty or ambiguity.

\newcommand{\Up}{\textcolor{Green}{\ensuremath{\uparrow}}}
\newcommand{\Upr}{\textcolor{BrickRed}{\ensuremath{\uparrow}}}
\newcommand{\Downg}{\textcolor{Green}{\ensuremath{\downarrow}}}
\newcommand{\Down}{\textcolor{BrickRed}{\ensuremath{\downarrow}}}
\newcommand{\Eq}{\textcolor{gray}{\ensuremath{\leftrightarrow}}}

\begin{table}[ht]
\centering
\small
\begin{subtable}{}
\centering
\caption{Performance metrics for the Adult dataset trained with a 3-layer MLP. For each metric, the left value shows the metric without LiLAW, and the right value shows the change with LiLAW. Upward arrows indicate increases, while downward arrows indicate decreases.}
\label{tab:adult}
\begin{tabular}{cc}
\toprule
Accuracy & Support \\
\midrule
0.8541 \textcolor{Green}{\ensuremath{\uparrow}} 0.0026 & 9769 \\
\bottomrule
\end{tabular}
%\caption{Overall accuracy}
\end{subtable}

\begin{subtable}{}
\centering
\begin{tabular}{lcccr}
\toprule
Class & Precision & Recall & F1-score & Support \\
\midrule
$\leq$50K
& 0.8827 \textcolor{Green}{\ensuremath{\uparrow}} 0.0002
& 0.9316 \textcolor{Green}{\ensuremath{\uparrow}} 0.0037
& 0.9065 \textcolor{Green}{\ensuremath{\uparrow}} 0.0018
& 7417 \\

$>$50K
& 0.7388 \textcolor{Green}{\ensuremath{\uparrow}} 0.0102
& 0.6097 \textcolor{BrickRed}{\ensuremath{\downarrow}} 0.0009
& 0.6681 \textcolor{Green}{\ensuremath{\uparrow}} 0.0036
& 2352 \\

\midrule
Average &  &  &  &  \\
\midrule
Macro average
& 0.8108 \textcolor{Green}{\ensuremath{\uparrow}} 0.0051
& 0.7707 \textcolor{Green}{\ensuremath{\uparrow}} 0.0014
& 0.7873 \textcolor{Green}{\ensuremath{\uparrow}} 0.0027
& 9769 \\

Weighted average
& 0.8481 \textcolor{Green}{\ensuremath{\uparrow}} 0.0026
& 0.8541 \textcolor{Green}{\ensuremath{\uparrow}} 0.0026
& 0.8491 \textcolor{Green}{\ensuremath{\uparrow}} 0.0023
& 9769 \\
\bottomrule
\end{tabular}
\end{subtable}
\end{table}

Using the Adult dataset for income prediction with a 70-30 train-test split (with 10\% of the training set used for validation), we study how using LiLAW on a 3-layer MLP generally improves performance and fairness across five different common fairness metrics~\cite{garg2020fairness}. We first report the performance boost when we use LiLAW in Table~\ref{tab:adult}.

We see a consistent improvement in performance, with an increase in the accuracy and the macro and weighted averages of precision, recall, and F1-score. While the $\leq$50K class sees consistent improvements since it is the majority label, the $>$50K class exhibits higher precision and F1-score at the cost of a small decrease in recall.

Now, let $A$ be a sensitive attribute, $Y\in\{0,1\}$ be the true label, and $\hat{Y}$ be the prediction. The disparity in the metrics relative to a reference group $r$ for each group $g \neq r$ is: $\Delta m_g = m_g - m_r$. Predictions are fairer when $\Delta m_g$ is near 0. Note: For sex, $r =$ Male. For race, $r =$ White. For age, $r =$ 45--54. These categories have the highest income in the Adult dataset.

We report the disparities in disadvantaged groups (based on sensitive attribute) relative to the reference group or privileged group with and without LiLAW for sex, race, and age in Tables~\ref{tab:sex},~\ref{tab:race}, and~\ref{tab:age}, respectively. We consider five fairness metrics:
\begin{enumerate}
\vspace{-0.3em}
    \item $\Delta DP_g = DP_g - DP_r$, where $DP_a = \Pr(\hat{Y}=1|A=a)$\newline
    This measures gap in \textit{demographic parity}, i.e. positive prediction rates are similar across two groups.
\vspace{-0.3em}
    \item $\Delta TPR_g = TPR_g - TPR_r$, where $TPR_a = \Pr(\hat{Y}=1|Y=1,A=a)$\newline
    This measures gap in the true positive rate and is referred to as \textit{equal opportunity} when it is equal across groups (true positives are identified at similar rates across two groups).
\vspace{-0.3em}
    \item $\Delta FPR_g = FPR_g - FPR_r$, where $FPR_a = \Pr(\hat{Y}=1|Y=0,A=a)$\newline
    This measures gap in the false positive rate (false positives are mistakenly identified at similar rates across two groups). Note that combining 2 and 3 gives us \textit{equalized odds}.
\vspace{-0.3em}
    \item $\Delta PPV_g = PPV_g - PPV_r$, where $PPV_a = \Pr(Y=1|\hat{Y}=1,A=a)$\newline
    This measures gap in the positive predictive value and called \textit{predictive parity} (probability of a positive outcome given a positive prediction is similar across two groups).
\vspace{-0.3em}
    \item $\Delta ACC_g = ACC_g - ACC_r$, where $ACC_a = \Pr(Y=\hat{Y}|A=a)$\newline
    This measures gap in accuracy and is referred to as \textit{accuracy parity}, i.e. accuracy is similar across two groups.
\end{enumerate}

\begin{table}[ht!]
\centering
\caption{Fairness disparities by sex relative to the reference group (Male) in the Adult Dataset trained with a 3-layer MLP. For each metric, the left value shows the disparity without LiLAW, and the right value shows the magnitude of the change with LiLAW. Upward arrows indicate that the metric increased with LiLAW, while downward arrows indicate a decrease. Green arrows indicate decreased disparity and red arrows indicate increased disparity.}
\label{tab:sex}
\scriptsize
\setlength{\tabcolsep}{3pt}
\renewcommand{\arraystretch}{1.15}
{%
\begin{tabular}{lrccccc}
\toprule
sex & n & $\Delta\,DP$ & $\Delta\,TPR$ & $\Delta\,FPR$ & $\Delta\,PPV$ & $\Delta\,ACC$ \\
\midrule
Male   & 6537 & $-$ & $-$ & $-$ & $-$ & $-$ \\
Female & 3232 & $\,-0.1712\,\Up\,0.0086\,$ & $\,-0.0522\,\Up\,0.0277\,$ & $\,-0.0703\,\Up\,0.0065\,$ & $\,-0.0083\,\Down\,0.0061\,$ & $\,0.1135\,\Downg\,0.0006\,$ \\
\bottomrule
\end{tabular}%
}
\end{table}

\begin{table}[ht!]
\centering
\caption{Fairness disparities by race relative to the reference group (White) in the Adult Dataset trained with a 3-layer MLP. For each metric, the left value shows the disparity without LiLAW, and the right value shows the magnitude of the change with LiLAW. Upward arrows indicate that the metric increased with LiLAW, while downward arrows indicate a decrease. Green arrows indicate decreased disparity and red arrows indicate increased disparity.}
\label{tab:race}
\scriptsize
\setlength{\tabcolsep}{2.25pt}
\renewcommand{\arraystretch}{1.15}
{%
\begin{tabular}{lrccccc}
\toprule
race & n & $\Delta\,DP$ & $\Delta\,TPR$ & $\Delta\,FPR$ & $\Delta\,PPV$ & $\Delta\,ACC$\\
\midrule
White              & 8304 & $-$ & $-$ & $-$ & $-$ & $-$ \\
Black              & 968  & $\,-0.1211\,\Up\,0.0062\,$ & $\,-0.0742\,\Up\,0.0005\,$ & $\,-0.0411\,\Up\,0.0076\,$ & $\,-0.0471\,\Down\,0.0341\,$ & $\,0.0718\,\Downg\,0.0060\,$\\
Asian-Pac-Islander & 331  & \hspace{1em}$\,0.0398\,\Downg\,0.0059\,$ & $\,-0.0307\,\Up\,0.0124\,$ & \hspace{1em}$\,0.0662\,\Downg\,0.0121\,$ & $\,-0.1579\,\Up\,0.0239\,$ & \hspace{-1em}$\,-0.0561\,\Up\,0.0122\,$\\
Amer-Indian%-Eskimo
& 94   & $\,-0.0727\,\Down\,0.0289\,$ & \hspace{1em}$\,0.0288\,\Down\,0.1424\,$ & $\,-0.0215\,\Down\,0.0084\,$ & $\,-0.0560\,\Down\,0.0030\,$ & $\,0.0566\,\Downg\,0.0135\,$\\
Other              & 72   & $\,-0.1276\,\Up\,0.0031\,$ & \hspace{1em}$\,0.0526\,\Up\,0.0005\,$ & $\,-0.0412\,\Up\,0.0041\,$ & $\,-0.0816\,\Down\,0.0107\,$ & $\,0.0968\,\Downg\,0.0029\,$\\
\bottomrule
\end{tabular}%
}
\end{table}

\begin{table}[ht!]
\centering

\caption{Fairness disparities by age relative to the reference group (45--54) in the Adult Dataset trained with a 3-layer MLP. For each metric, the left value shows the disparity without LiLAW, and the right value shows the magnitude of the change with LiLAW. Upward arrows indicate that the metric increased with LiLAW, while downward arrows indicate a decrease. Green arrows indicate decreased disparity and red arrows indicate increased disparity.}
\label{tab:age}
\scriptsize
\setlength{\tabcolsep}{3pt}
\renewcommand{\arraystretch}{1.15}
{%
\begin{tabular}{lrccccc}
\toprule
age & n & $\Delta\,DP$ & $\Delta\,TPR$ & $\Delta\,FPR$ & $\Delta\,PPV$ & $\Delta\,ACC$\\
\midrule
$\leq 24$     & 1658 & $\,-0.3456\,\Up\,0.0110\,$ & $\,-0.3295\,\Up\,0.0100\,$ & $\,-0.1391\,\Up\,0.0116\,$ & \hspace{1em}$\,0.0960\,\Downg\,0.0127\,$ & $\,0.2112\,\Downg\,0.0029\,$\\
25--34        & 2621 & $\,-0.2411\,\Up\,0.0102\,$ & $\,-0.2065\,\Up\,0.0123\,$ & $\,-0.1024\,\Up\,0.0102\,$ & $\,-0.0453\,\Down\,0.0041\,$ & $\,0.0954\,\Downg\,0.0014\,$ \\
35--44        & 2401 & $\,-0.0450\,\Up\,0.0076\,$ & \hspace{1em}$\,0.0185\,\Upr\,0.0126\,$ & $\,-0.0208\,\Up\,0.0053\,$ & $\,-0.0220\,\Down\,0.0018\,$ & $\,0.0341\,\Upr\,0.0021\,$  \\
45--54        & 1735 & $-$ & $-$ & $-$ & $-$ & $-$ \\
55--64        & 947  & $\,-0.0859\,\Up\,0.0078\,$ & $\,-0.0951\,\Up\,0.0100\,$ & $\,-0.0235\,\Up\,0.0069\,$ & $\,-0.0571\,\Down\,0.0041\,$ & \hspace{-1em}$\,-0.0006\,\Up\,0.0003\,$  \\
$65+$         & 407  & $\,-0.2024\,\Up\,0.0183\,$ & $\,-0.1396\,\Up\,0.0333\,$ & $\,-0.0930\,\Up\,0.0147\,$ & $\,-0.0111\,\Down\,0.0167\,$ & $\,0.0814\,\Downg\,0.0004\,$  \\
\bottomrule
\end{tabular}%
}
\end{table}

We see in Tables~\ref{tab:sex},~\ref{tab:race}, and~\ref{tab:age} that we compare each group in each table to a privileged reference group, where values closer to 0 imply smaller disparities. In general, adding LiLAW tends to move most non-reference group disparities closer to 0 for improved demographic parity, equalized odds (inc. equal opportunity), and accuracy parity. Some groups have an increase in predictive parity with LiLAW, since it is generally impossible to simultaneously achieve demographic parity, equalized odds, and predictive parity due to the impossibility theorem~\citep{imposs}. Results from small groups should be treated cautiously as they can fluctuate more, although in our case, we still see improvements.

\subsubsection{Proof that LiLAW helps improve fairness}

Intuitively, each sample loss is multiplied by a differentiable weight
$\mathcal{W}(s_i,\tilde y_i)$ that depends on $s_i[\tilde y_i]$ and $\max(s_i)$ as mentioned in Section~\ref{sec:method}. Since $\mathcal{W}(s_i,\tilde y_i)$ is designed / updated to emphasize low confidence and/or disagreement with the observed label, optimization focuses more on informative hard / moderate examples while reducing the influence of easy points, typically improving generalization. 

And, since underrepresented groups contain a higher fraction of these harder cases (due to imbalance, label noise, or distribution shift), LiLAW implicitly increases their training signal. As a result, LiLAW can reduce error-rate disparities when subgroup disadvantage manifests as correctable high-loss or ambiguous examples. We evaluate the resulting changes in demographic parity, equalized odds, predictive parity, and accuracy parity empirically.

\begin{proposition}
Let $A$ be a sensitive attribute, $Y\in\{0,1\}$ be the true label, and $\widehat{Y}\in\{0,1\}$ be the prediction. Let $r$ be a reference group and let $g\neq r$ be a non-reference (disadvantaged) group. Suppose that LiLAW assigns larger effective weight to informative high-loss samples in group $g$, and that this extra training signal improves the error rates of group $g$ more than those of group $r$. Specifically, let superscripts $0$ and $1$ denote the metrics before and after LiLAW training, respectively, and assume:
\begin{equation} 
TPR_g^1-TPR_g^0
\ge
TPR_r^1-TPR_r^0,
\end{equation}
and:
\begin{equation} 
FPR_g^1-FPR_g^0
\le
FPR_r^1-FPR_r^0.
\end{equation}
If initially $TPR_g^0\le TPR_r^0$ and $FPR_g^0\ge FPR_r^0$, and the updates reduce disparity (rather than flip it), then the group-wise $TPR$ and $FPR$ disparities for group $g$ decrease. Consequently, LiLAW can reduce equalized odds disparities under these conditions. Changes in demographic parity, predictive parity, and accuracy parity require specific conditions and are evaluated empirically.
\label{prop}
\end{proposition}

\begin{proof}[Proof]
LiLAW optimizes the weighted objective:
\begin{equation} 
\min_{\theta}
\sum_{(x_i,\widetilde{y}_i)\in\mathcal{D}_t}
\mathcal{W}(s_i,\widetilde{y}_i)
\cdot
\ell(f_{\theta}(x_i),\widetilde{y}_i).
\end{equation}
Splitting the objective by groups gives:
\begin{equation} 
\min_{\theta}
\sum_{a}
\sum_{(x_i,\widetilde{y}_i)\in\mathcal{D}_a}
\mathcal{W}(s_i,\widetilde{y}_i)
\cdot
\ell(f_{\theta}(x_i),\widetilde{y}_i).
\end{equation}
Thus, if a group contains more informative high-loss samples and LiLAW assigns larger weights to those samples, that group contributes more to the weighted training objective. This does not guarantee fairness, but provides a way in which LiLAW can reduce group-specific errors.

Suppose the additional training signal improves group $g$'s TPR at least as much as group $r$'s:
\begin{equation} 
TPR_g^1-TPR_g^0
\ge
TPR_r^1-TPR_r^0.
\end{equation}
If $TPR_g^0\le TPR_r^0$ \& $TPR_g^1\le TPR_r^1$, i.e. group $g$ is disadvantaged before and after LiLAW, then:
\begin{align} 
|TPR_g^1-TPR_r^1|
&=
TPR_r^1-TPR_g^1 \\
&=
(TPR_r^0-TPR_g^0)
-
\Big[
(TPR_g^1-TPR_g^0)
-
(TPR_r^1-TPR_r^0)
\Big] \\
&\le
TPR_r^0-TPR_g^0 =
|TPR_g^0-TPR_r^0|.
\end{align}
Therefore, the $TPR$ disparity decreases.

Similarly, suppose group $g$'s FPR decreases at least as much as group $r$'s:
\begin{equation} 
FPR_g^1-FPR_g^0
\le
FPR_r^1-FPR_r^0.
\end{equation}
If $FPR_g^0\ge FPR_r^0$ \& $FPR_g^1\ge FPR_r^1$, i.e. group $g$ is disadvantaged before and after LiLAW, then: 
\begin{align}
|FPR_g^1-FPR_r^1|
&=
FPR_g^1-FPR_r^1 \\
&=
(FPR_g^0-FPR_r^0)
+
\Big[
(FPR_g^1-FPR_g^0)
-
(FPR_r^1-FPR_r^0)
\Big] \\
&\le
FPR_g^0-FPR_r^0 =
|FPR_g^0-FPR_r^0|.
\end{align}
Therefore, the $FPR$ disparity decreases. Together, these two results show that LiLAW can reduce equalized odds disparities when its additional weighting improves the disadvantaged group's error rates more than the reference group's.

The other fairness metrics do not follow automatically from reduced error rates. We have:
\begin{equation} 
ACC_a = \Pr(Y=\hat{Y}|A=a) = \Pr(Y=1 | A=a) TPR_a + \Pr(Y=0 | A=a)(1-FPR_a),
\end{equation}
\begin{equation} 
DP_a = \Pr(\hat{Y}=1|A=a) = \Pr(Y=1 | A=a) TPR_a + \Pr(Y=0 | A=a)FPR_a,
\end{equation}
\begin{equation} 
PPV_a = \Pr(Y=1\mid \hat{Y}=1,A=a) = \frac{\Pr(Y=1 | A=a) TPR_a} {\Pr(Y=1 | A=a) TPR_a+\Pr(Y=0 | A=a) FPR_a}.
\end{equation}

We see improvements in equalized odds due to lower $\Delta TPR_g$ and $\Delta FPR_g$. However, the three metrics above (demographic parity, predictive parity, and accuracy parity) depend on prior probabilities and on metric-specific combinations of $TPR$ and $FPR$ (positively or negatively correlated). Therefore, improvements in these metrics are reported above empirically.
\end{proof}

\subsection{Representative samples for each set of weights}
\label{supp:samples} 

\begin{figure}[ht!]
    \centering
    \includegraphics[scale=0.308]{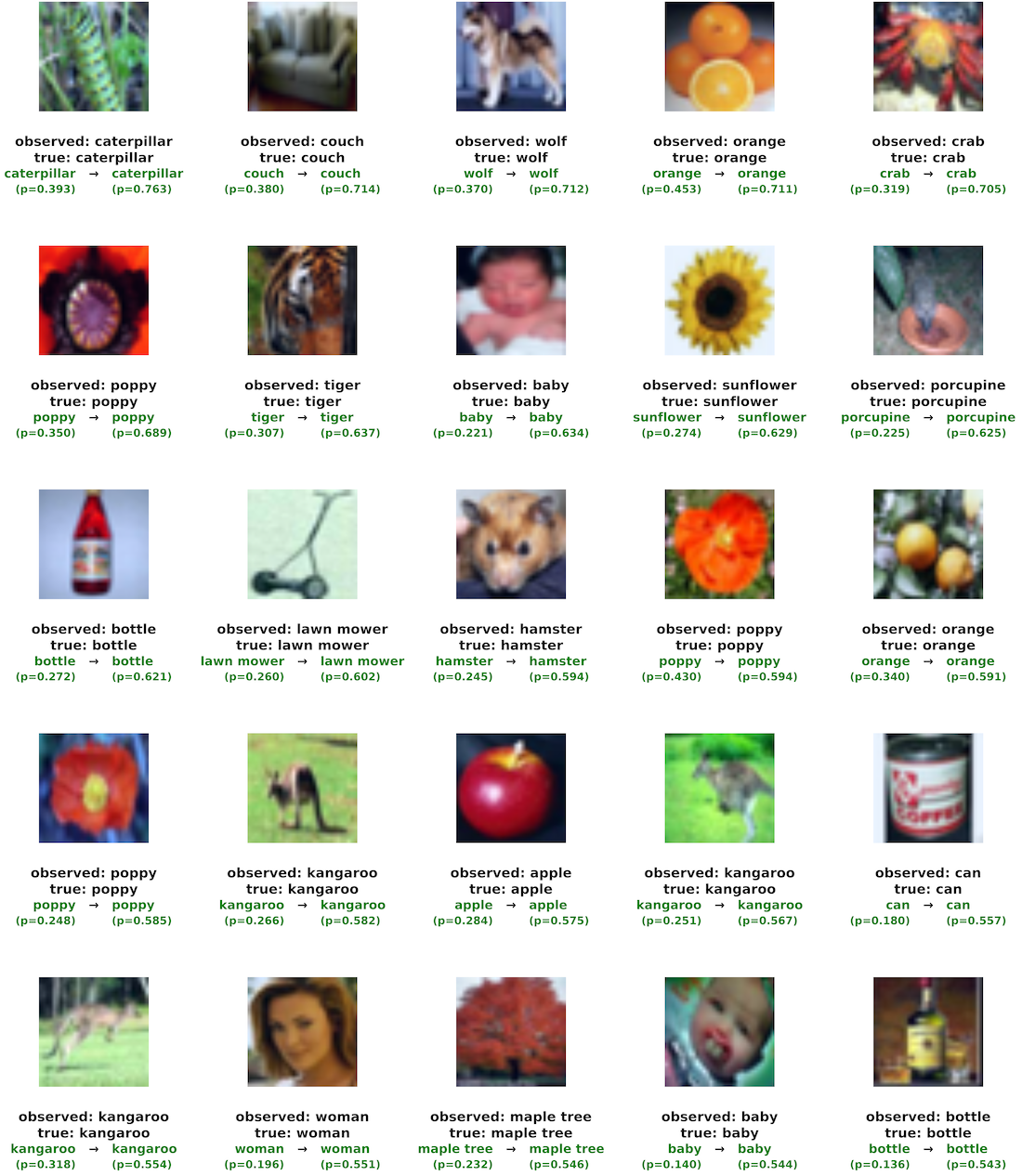}
    \caption{Top 25 training samples with the largest $\mathcal{W}_\alpha$ values. For each tile, the observed label and true label are shown above the prediction without LiLAW (left of the arrow) and prediction with LiLAW (right of the arrow); green text indicates a correct prediction and red text indicates an incorrect prediction with respect to the true label. In all 25 examples shown here, the observed label is the same as the true label, so these are clean samples, and the model prediction agrees with both the observed and true label for both predictions. The main change is confidence: the predicted-class probability rises, so the highest $\mathcal{W}_\alpha$ samples are mostly low-disagreement samples that move from moderate agreement into confident agreement. In the LiLAW geometry, these are the samples that become increasingly \emph{easy}: they already point in the correct direction, and LiLAW helps them move further into stable, high-confidence predictions.}
    \label{fig:alpha}
\end{figure}

\begin{figure}[ht!]
    \centering
    \includegraphics[scale=0.308]{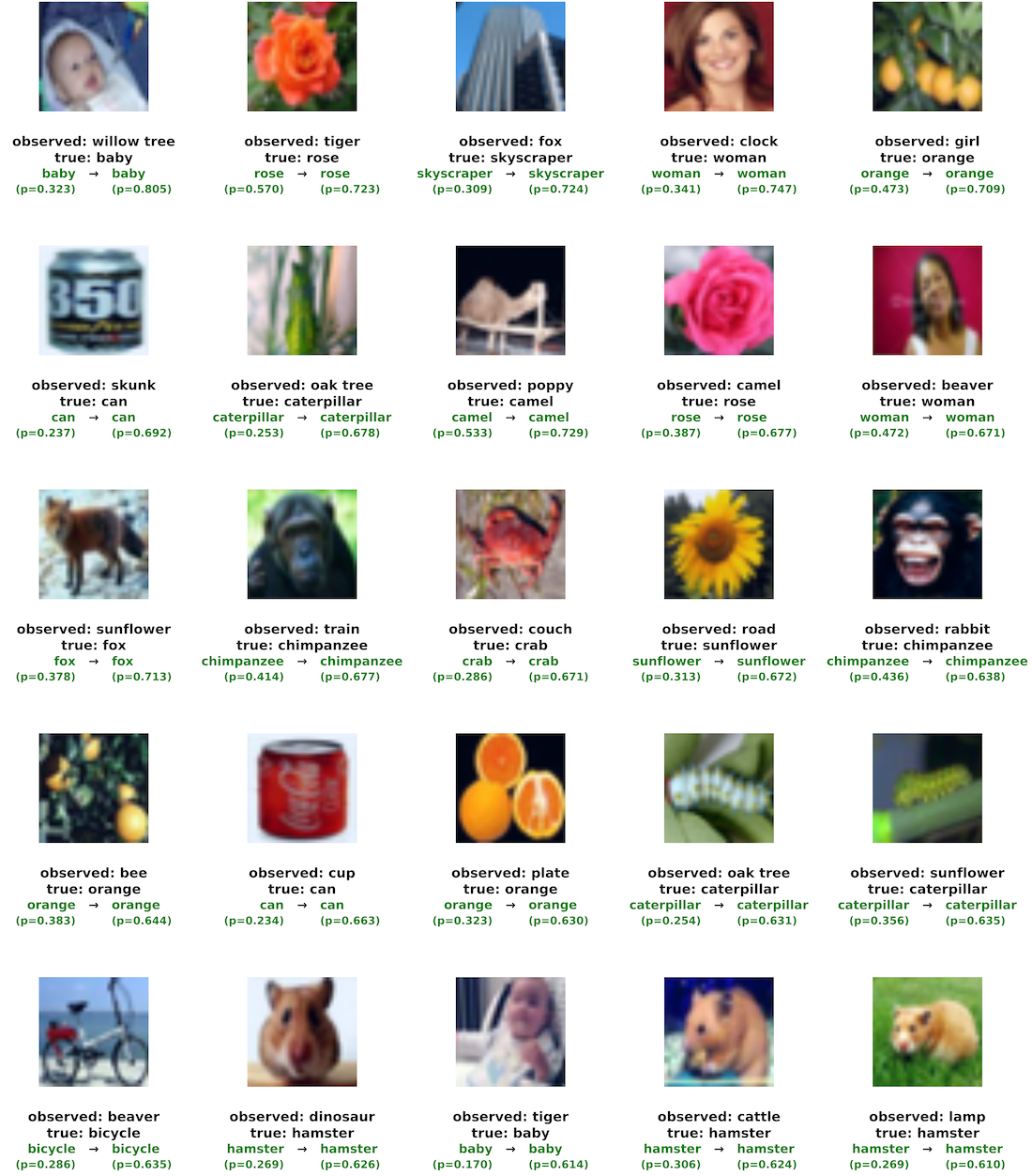}
    \caption{Top 25 training samples with the largest $\mathcal{W}_\beta$ values. For each tile, the observed label and true label are shown above the prediction without LiLAW (left of the arrow) and prediction with LiLAW (right of the arrow); green text indicates a correct prediction and red text indicates an incorrect prediction with respect to the true label. All 25 displayed samples have the observed label not equal to the true label, so they are noisy-label samples. Nevertheless, in every shown case, the model prediction agrees with the true label rather than the observed label, with and without LiLAW, and the predicted-class probability increases. These are \emph{hard} samples in the LiLAW geometry: the network is learning the visual semantics of the image correctly, but it increasingly and confidently disagrees with the observed annotation. Large $\mathcal{W}_\beta$ therefore identifies high-disagreement samples that are likely mislabeled and highlights the cases where the model is most certain that the observed label should not be trusted.}
    \label{fig:beta}
\end{figure}

\begin{figure}[ht!]
    \centering
    \includegraphics[scale=0.308]{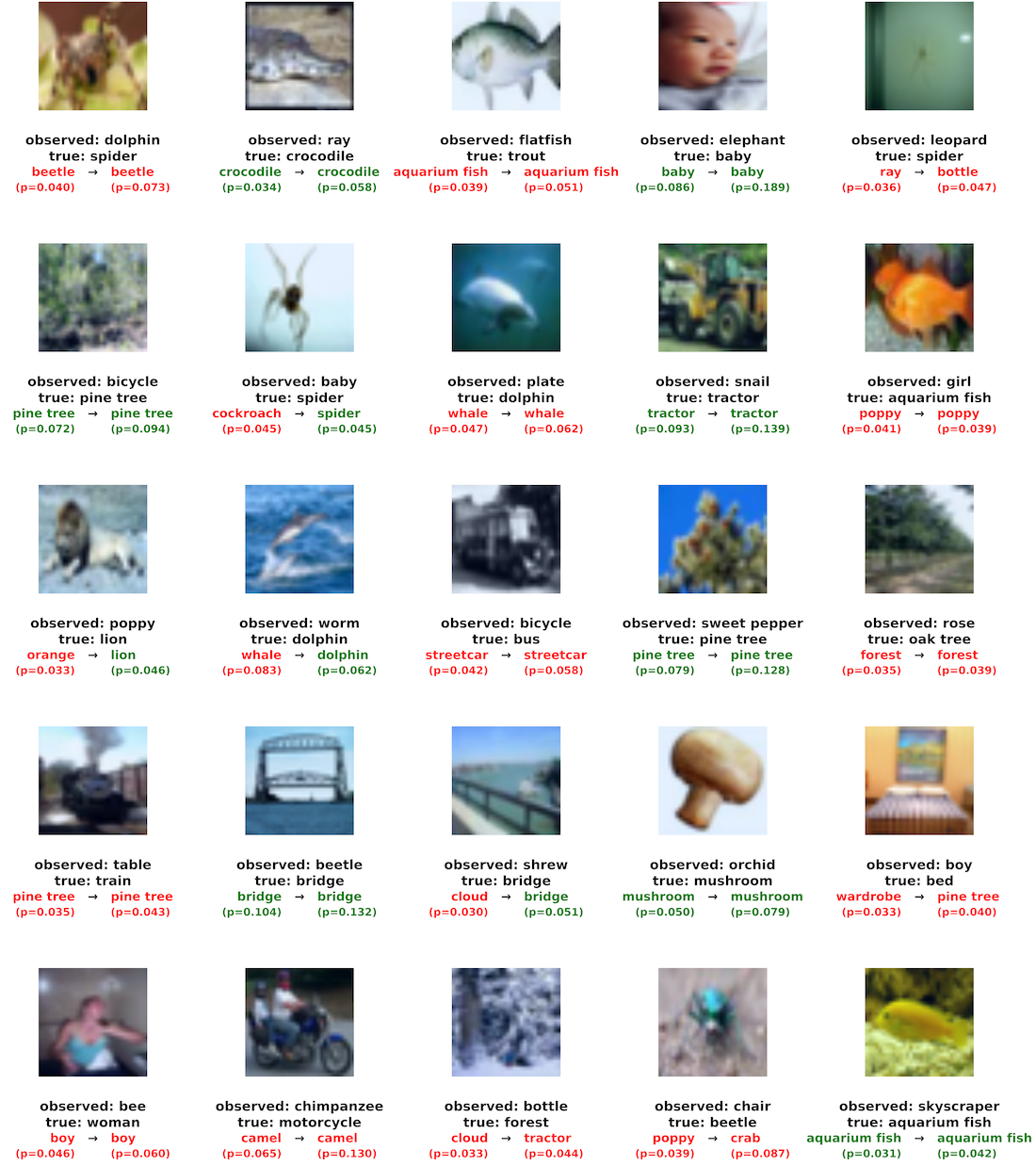}
    \caption{Top 25 training samples with the largest $\mathcal{W}_\delta$ values. For each tile, the observed label and true label are shown above the prediction without LiLAW (left of the arrow) and prediction with LiLAW (right of the arrow); green text indicates a correct prediction and red text indicates an incorrect prediction with respect to the true label. All displayed examples have the observed label not equal to the true label, so they are noisy-label samples, but their behavior is much less settled. The earlier and later predictions are mixed: some samples are correct with and without LiLAW, some switch from an incorrect class to the true class, and some remain incorrect even at the later stage. Their predicted-class probabilities stay low overall, increasing only slightly. This is the \emph{moderate} transition-band behavior that $\mathcal{W}_\delta$ is designed to capture: large $\mathcal{W}_\delta$ surfaces ambiguous mislabeled examples with some beginning to move toward the true class, some remaining borderline, and some having not yet escaped incorrect low-confidence predictions, bridging the easy region of $\mathcal{W}_\alpha$ and the hard region of $\mathcal{W}_\beta$.}
    \label{fig:delta}
\end{figure}

We display 25 samples with the top $\mathcal{W}_\alpha$ weights (see~\ref{supp:alpha}), top $\mathcal{W}_\beta$ weights (see~\ref{supp:beta}), and top $\mathcal{W}_\delta$ weights (see~\ref{supp:delta}). Figure~\ref{fig:alpha} contains clean samples where the observed and true labels agree, and the model stays correct while becoming more confident. Figure~\ref{fig:beta} contains mislabeled hard samples where the observed and true labels differ, but the model increasingly predicts the true class with higher confidence. Figure~\ref{fig:delta} contains mislabeled moderate samples whose predictions remain low-confidence, mixed, or only partially corrected over time, so they lie in the transition band between easy agreement and hard confident disagreement.

\subsubsection{Top 25 highest \texorpdfstring{$\mathcal{W}_\alpha$}{W alpha} samples from CIFAR-100-M}
\label{supp:alpha} 
\newpage
\subsubsection{Top 25 highest \texorpdfstring{$\mathcal{W}_\beta$}{W beta} samples from CIFAR-100-M}
\label{supp:beta} 
\newpage
\subsubsection{Top 25 highest \texorpdfstring{$\mathcal{W}_\delta$}{W delta} samples from CIFAR-100-M}
\label{supp:delta} 

%%%%%%%%%%%%%%%%%%%%%%%%%%%%%%%%%%%%%%%%%%%%%%%%%%%%%%%%%%%%

%\newpage
%\input{checklist.tex}

\end{document}